\documentclass[final,onefignum,onetabnum]{siamltexmm}
\usepackage{cite}
\usepackage{amssymb}
\usepackage{amsmath}
\usepackage{graphicx}
\usepackage{sidecap}
\usepackage{subcaption}
\usepackage[]{algorithm2e}

\usepackage{enumitem}
\usepackage{cleveref}
\usepackage{multirow}
\usepackage{tabularx}
\usepackage{epstopdf}
\usepackage{float}
\usepackage{IEEEtrantools}
\usepackage{placeins}

\usepackage{color,soul}
\soulregister\cite7
\soulregister\ref7

\makeatletter
\newcommand{\rmnum}[1]{\romannumeral #1}
\newcommand{\Rmnum}[1]{\expandafter\@slowromancap\romannumeral #1@}
\newcommand{\IEEElabel}[1]{\begingroup\addtocounter{equation}{-1}%
	\refstepcounter{equation}\label{#1}\endgroup}
\makeatother

\title{Patch-Ordering as a Regularization for Inverse Problems in Image Processing\thanks{{The research leading to these results has received funding from the	European Research Council under European Union’s Seventh Framework Program, ERC Grant agreement no. 320649.}}}

\author{Gregory~Vaksman\footnotemark[2]
\and Michael~Zibulevsky\footnotemark[3]
\and Michael~Elad\footnotemark[3]}

\begin{document}

%\sloppy
%\setlength{\emergencystretch}{40pt}

% make the title area
\maketitle
	
\newcommand{\slugmaster}{%
\slugger{siims}{xxxx}{xx}{x}{x--x}}%slugger should be set to juq, siads, sifin, or siims

\renewcommand{\thefootnote}{\fnsymbol{footnote}}
\footnotetext[2]{Department of Electrical Engineering, Technion – Israel Institute of Technology, Technion City, Haifa 32000, Israel (\email{grishav@tx.technion.ac.il})}
\footnotetext[3]{Department of Computer Science, Technion – Israel Institute of Technology, Technion City, Haifa 32000, Israel (\email{mzib@cs.technion.ac.il}, \email{elad@cs.technion.ac.il})}
\renewcommand{\thefootnote}{\arabic{footnote}}

\pagestyle{myheadings}
\thispagestyle{plain}
\markboth{GREGORY~VAKSMAN, MICHAEL~ZIBULEVSKY, AND MICHAEL~ELAD}{PATCH ORDERING REGULARIZATION FOR INVERSE PROBLEMS}

\graphicspath{{./Figures/}}

\begin{abstract}
	\par Recent work in image processing suggests that operating on (overlapping) patches in an image may lead to state-of-the-art results. This has been demonstrated for a variety of problems including denoising, inpainting, deblurring, and super-resolution. The work reported in~\cite{Ram_Patch_Ordering_2013,Ram_Patch_Ordering_Wavelet_Frame_2014} takes an extra step forward by showing that ordering these patches to form an approximate shortest path can be leveraged for better processing. The core idea is to apply a simple filter on the resulting 1D smoothed signal obtained after the patch-permutation. This idea has been also explored in combination with a wavelet pyramid, leading eventually to a sophisticated and highly effective regularizer for inverse problems in imaging. 	
	
	\par In this work we further study the patch-permutation concept, and harness it to propose a new simple yet effective regularization for image restoration problems. Our approach builds on the classic Maximum A'posteriori probability (MAP), with a penalty function consisting of a regular log-likelihood term and a novel permutation-based regularization term. Using a plain 1D Laplacian, the proposed regularization forces robust smoothness~($L1$) on the permuted pixels. Since the permutation originates from patch-ordering, we propose to accumulate the smoothness terms over all the patches' pixels. Furthermore, we take into account the found distances between adjacent patches in the ordering, by weighting the Laplacian outcome.
	
	\par We demonstrate the proposed scheme on a diverse set of problems: (\rmnum{1})~severe Poisson image denoising, (\rmnum{2})~Gaussian image denoising, (\rmnum{3})~image deblurring, and (\rmnum{4})~single image super-resolution. In all these cases, we use recent methods that handle these problems as initialization to our scheme. This is followed by an L-BFGS optimization of the above-described penalty function, leading to state-of-the-art results, and especially so for highly ill-posed cases.
\end{abstract}



\section{Introduction}
	\par In recent years we see an interesting trend, in which many image restoration algorithms choose to operate on local image patches rather than processing the image as a whole. These techniques impose statistical prior knowledge on the patches of the processed image. Surprisingly, in many cases these methods lead to state-of-the-art results. For example, in the Gaussian denoising case, the algorithm presented  in~\cite{Elad_Image_Denoising_KSVD_2006} performs denoising using a statistical model based on sparse representation of the image patches, training a dictionary using the K-SVD algorithm. The BM3D algorithm reported in~\cite{Egiazarian_BM3D_2007} exploits interrelations between patches by grouping them into 3D groups, and applying collaborative filtering on them that is based on sparse representation as well. The work reported in~\cite{Salmon_NLPCA_2012,Salmon_NLPCA_2014,Giryes_SPDA_2013} extends PCA and dictionary learning, both in the context of patches for handling severe Poisson image denoising. The scheme reported in~\cite{Yang_SR_2010} adopts the sparse representation model to handle the single image super-resolution problem. The papers~\cite{Yu_Sapiro_Mallat_MAP_EM_2012,Zoran_Weiss_EPLL_2011} both propose a GMM modeling of image patches, and demonstrate the effectiveness of this to variety of inverse problems. The NSCR method by Dong et. al.~\cite{Dong_SR_2013} uses sparse representation for solving both the super-resolution and the deblurring problems. The IDD-BM3D method in~\cite{Egiazarian_IDD_BM3D_2012} employs BM3D frames for image deblurring. All these papers and many others rely on operating on patches in order to complete the restoration task at hand. Many works use sophisticated priors when operating locally, and most often they resort to a simple averaging when combining the restored patches.

	\par The work reported in~\cite{Ram_Patch_Ordering_2013} takes another step toward exploiting interrelations between image patches in the image. The work reported in~\cite{Ram_Patch_Ordering_2013} proposes to construct a 1D smoothed signal by applying a permutation on the pixels of the corrupted image. The permutation is obtained by ordering image patches to form "the shortest possible path", approximating the solution of the traveling salesman problem (TSP). Given the sorted image, the clean image is recovered by applying a 1D filter on the ordered signal. This method is simple and yet it leads to high-quality results. On the down side, it is limited to the Gaussian denoising and inpainting problems. The work reported in~\cite{Ram_Patch_Ordering_Wavelet_Frame_2014} takes an extra step in exploring the patch-ordering concept. This work constructs a sophisticated and very powerful regularizer by combining the patch-permutation idea with a wavelet pyramid~\cite{Ram_Tree_Based_Wavelet_2011,Ram_Redundant_Wavelet_on_Graphs_2012}. The obtained regularizer is used for solving general inverse problems in imaging. This method leads to high-quality results in series of tests, however it is quite involved.

	\par In this paper we propose to compose the whole image from local patches using a prior that exploits interrelation between them. We harness the patch-permutation idea, merging it with the classical Maximum A'posteriori probability (MAP) estimator, by proposing a new, simple, yet powerful regularization for inverse imaging problems. We formulate the inverse problem as a weighted sum of two penalty terms: (\rmnum{1})~a regular negative log-likelihood, and (\rmnum{2})~a novel regularization expression that forces smoothness in a robust way, by basing it on reordered list of the image pixels, obtained according to the similarity of image patches. 

	\par For constructing the permutation-based regulatization, we follow the core idea presented in~\cite{Ram_Patch_Ordering_2013}. We rely on the assumption that in an ideal image, close similarity between patches indicates a proximity between their center pixels. We therefore build a 1D (piece-wise) smoothed signal by applying a patch-based permutation on the restored (unknown) image pixels. The permutation is obtained by extracting all possible patches with overlaps from the currently recovered image, and ordering them to form the (approximated) shortest possible path. The resulting ordering induces the permutation. 

	\par The proposed regularization forces smoothness on the obtained 1D signal via a Laplacian, penalized by the robust ${L_1}$ norm. Since the ordering is associated with all the pixels withing the patches, the smoothness term is accumulated over all these pixels. We also deploy weights that take into account the actual distances between the consecutive patches in the ordering. 

	\par The proposed scheme is demonstrated on several different problems: (\rmnum{1})~White additive Gaussian image denoising, (\rmnum{2})~severe Poisson image denoising, (\rmnum{3})~image deblurring, and (\rmnum{4})~single image super-resolution. We initialize our algorithm with an output of a recent method that handles each of these problems. The reconstructed image is then obtained by minimizing the penalty function described above using the L-BFGS method~\cite{Liu_LBFGS_1989,Schmidt_LBFGS_2005}. Our extensive experiments indicate that applying the proposed scheme leads to state-of-the-art results.

	\par We should note that the proposed scheme bares some similarity to recent work offering regularization of inverse problems by utilizing the similarity between patches formed as a graph~\cite{Milanfar_Tour_of_Modern_Image_Filtering_2013,Milanfar_General_Framework_for_Image_Restoration_2014,Kervrann_Denoising_Fluorescence_Microscopy_2010,Peyre_Non_local_regularization_of_inverse_problems_2011,Elmoataz_Nonlocal_Discrete_Regularization_on_Weighted_Graphs_2008,Bougleux_Local_and_nonlocal_discrete_regularization_on_weighted_graphs_2009,Liu_Progressive_Image_Denoising_Through_Hybrid_Graph_Laplacian_2014,Haque_Symmetric_Smoothing_Filters_From_Global_Consistency_2014,Romano_Boosting_of_Image_Denoising_Algorithms_2015}. These works propose various formats of using this graph's Laplacian as a sparsifying operator. Our approach could be considered as a special case of such a Laplacian regularization, which forces the graph to be a simple and continuous chained ordering of the image pixels. As such, the regularization we obtain is simpler and easier to manage (since we keep only one forward and one backward neighbors per each pixel). In addition, our approach also provides a stronger stabilizing effect for highly ill-posed problems since it ties all the pixels to each other.
	
	\par The paper is organized as follows. In Section~\ref{sec:permutations} we discuss the principles behind the permutation construction, and how it becomes useful as a regularizer. Section~\ref{sec:reconstruction_algorithm} describes the proposed algorithm, along with the numerical scheme used. Section~\ref{sec:results} presents experiment results and compares the new method with other leading schemes. Section~\ref{sec:conclusion} concludes this paper and raises directions for a future work. 

\section{Constructing the permutation}
	\label{sec:permutations}
	\par A common assumption in image processing is that clean images are usually (piece-wise) smooth, i.e. the difference between any two neighboring pixels tends to be small. Most image processing algorithms, be it for restoration, segmentation, compression, and more, rely on this model to some extent. The problem with this assumption, however, is that violation from this behavior is due to image edges and texture, and both are central in forming and defining the visual content of an image, and as such, they cannot be sacrificed as simple outliers.
	
	\par Adopting a totally different perspective towards handling of an image, a convenient and often used technique for developing an image processing algorithm is to convert the 2D image into a 1D array, treat the image as a vector, and then convert the resulting 1D array back to a 2D array. There are several popular scan methods that convert an image to a vector, the prominent of which are raster scan, zigzag scan, Hilbert Peano and other space-filling curves.
	
	\par The question we pose now is this: how can we combine the two approaches mentioned above? Namely, given a corrupted image ${\mathbf{y}\in\mathbb{R}^N}$, how should we construct a 2D-to-1D conversion that produces the smoothest possible vector when applied to a clean image, ${\mathbf{x}}$. Such a conversion would be extremely helpful for image restoration tasks, because in the 1D domain its processing is expected to be very simple and effective. We should emphasize that the sought conversion method must be robust, i.e. be able to produce a meaningful ordering even when operated on a corrupted data.
	
	\par A word on notations: The 2D-to-1D conversion we are seeking is denoted by ${\Omega}$, and represented by a permutation matrix ${P}$. I.e., applying ${P}$ to column-stacked image ${\mathbf{x}}$, ${P\mathbf{x}}$, produces a vector with the reordering ${\Omega}$.
	
	\par In this work we build on the ideas presented in~\cite{Ram_Patch_Ordering_2013}, while giving it a new and novel interpretation. In order to propose such a smoothing conversion, we shall assume that each pixel of the clean image ${\mathbf{x}}$ can be represented by the corresponding surrounding patch taken from the corrupted image ${\mathbf{y}}$. In other words, the assumption is that if two patches of the corrupted (or clean) image are close in terms of some distance function, then their corresponding central pixels in the clean image are expected to be similar as well. We will refer hereafter to this as the $A1$ {\em assumption}. Throughout this work we shall use the Euclidean distance for assessing proximity between patches. We extract all possible patches ${\{\mathbf{z_i}\}_1^N}$ of size ${\sqrt{n} \times \sqrt{n}}$ with overlaps from the image ${\mathbf{y}}$, where ${n \ll N}$.  Our method refers to the patches as points in ${\mathbb{R}^n}$, defining a graph where the patches are its vertices and distances between them are the edges.
	
	\par Recall that our aim is to find an ordering ${\Omega}$ (or ${P}$, its permutation matrix equivalent) such that the resulting permuted clean image is the smoothest. This can be defined as the following minimization task:
	\begin{equation}
	\min_P \|DP\mathbf{x}\|_1 = \min_\Omega \sum_{i=2}^{N} \left|x_{\Omega(i)} - x_{\Omega(i-1)}\right|\;,  
	\end{equation}
	where ${D}$ is a 1D difference operator (simple derivative). Interestingly, it is tempting to propose an ordering of the image patches in ${\mathbf{x}}$ based on a simple sort of their center pixels, as it will lead to the ideal (smoothest) permutation. However, since we assume that ${\mathbf{x}}$ is not available and instead we have ${\mathbf{y}}$, this option is impossible. Thus, the above objective is replaced by an alternative that relies on the assumption ${A1}$ made above:
	\begin{equation}
	\min_P \|DP\mathbf{x}\|_1 \approx \min_\Omega \sum_{i=2}^{N} \left\|\mathbf{z}_{\Omega(i)} - \mathbf{z}_{\Omega(i-1)}\right\|_2\;.
	\end{equation} 
	This means that the graph vertices are ordered to form the shortest possible path that visits each vertex exactly once. Namely, we formulate the quest of $\Omega$ as a classic Traveling Salesman Problem (TSP)~\cite{Cormen_Algorithms_2001}.
	
	\par
\begin{figure}
		\centering
		\begin{subfigure}{0.32\textwidth}
			\includegraphics[width=\textwidth]{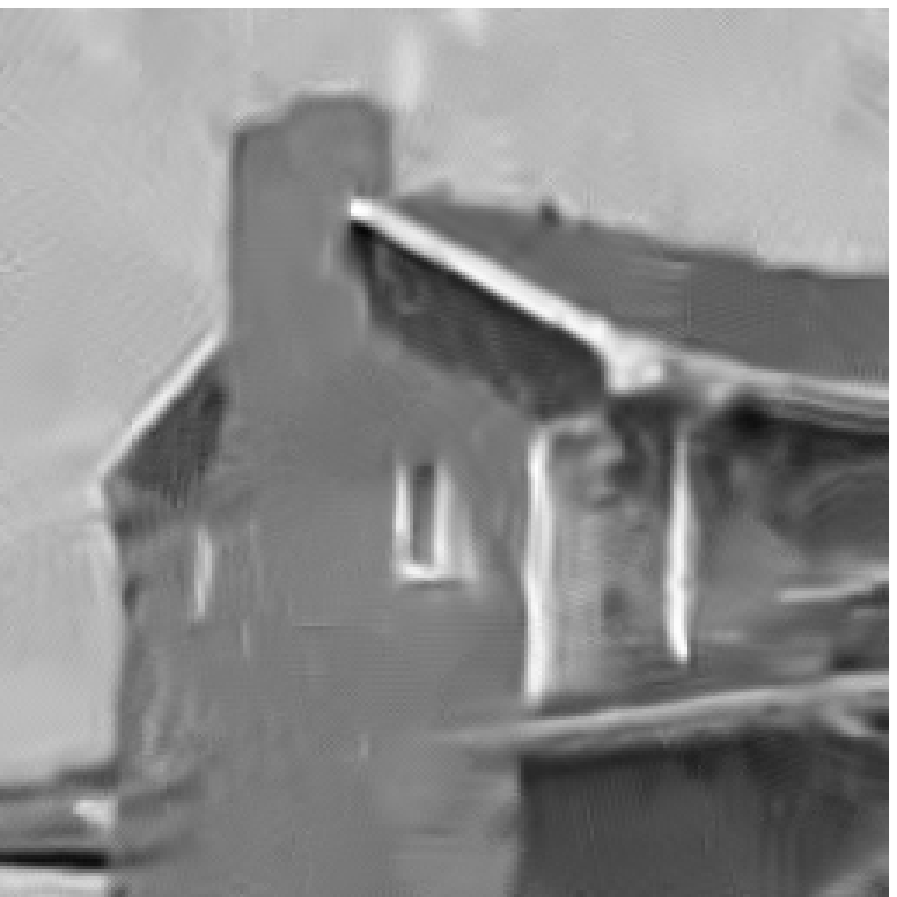}
			\caption{\centering The initial image (BM3D) \newline PSNR = 25.87dB}
			\label{fig:Lin-Kernighan:bm3d}
			\vspace*{6pt}
		\end{subfigure}
		\begin{subfigure}{0.32\textwidth}
			\includegraphics[width=\textwidth]{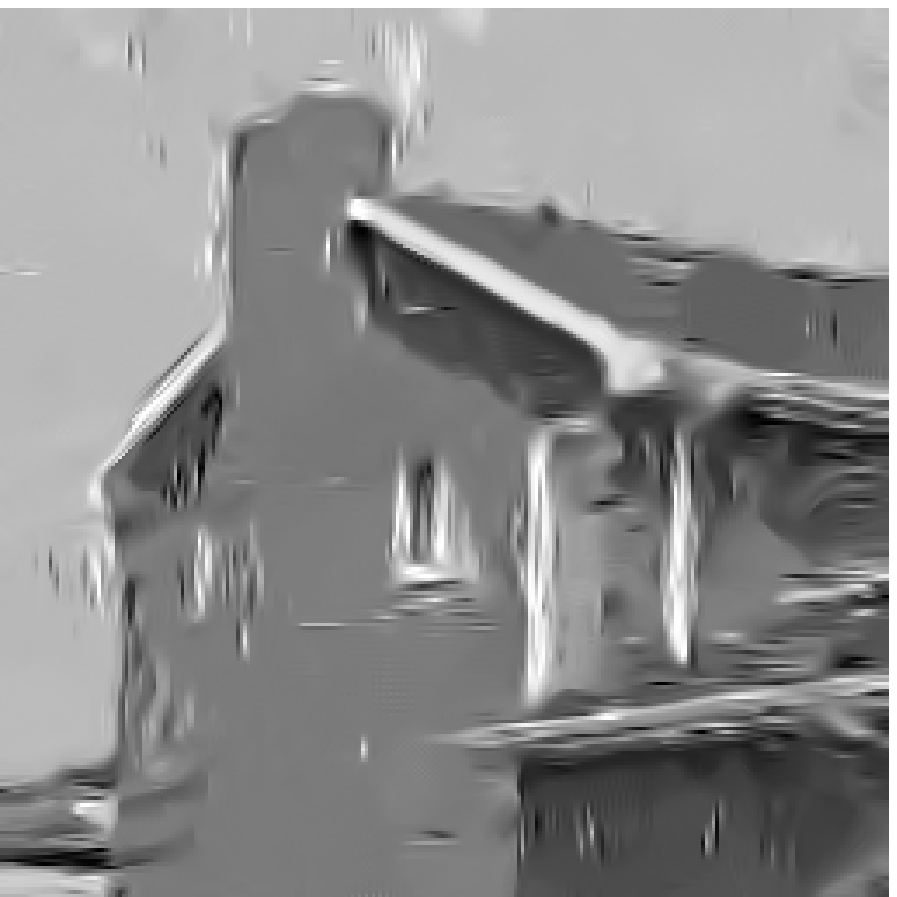}
			\caption{\centering Output with Lin-Kernighan \newline PSNR = 24.17dB}
			\label{fig:Lin-Kernighan:bm3d_out}
			\vspace*{6pt}
		\end{subfigure}
		\begin{subfigure}{0.32\textwidth}
			\includegraphics[width=\textwidth]{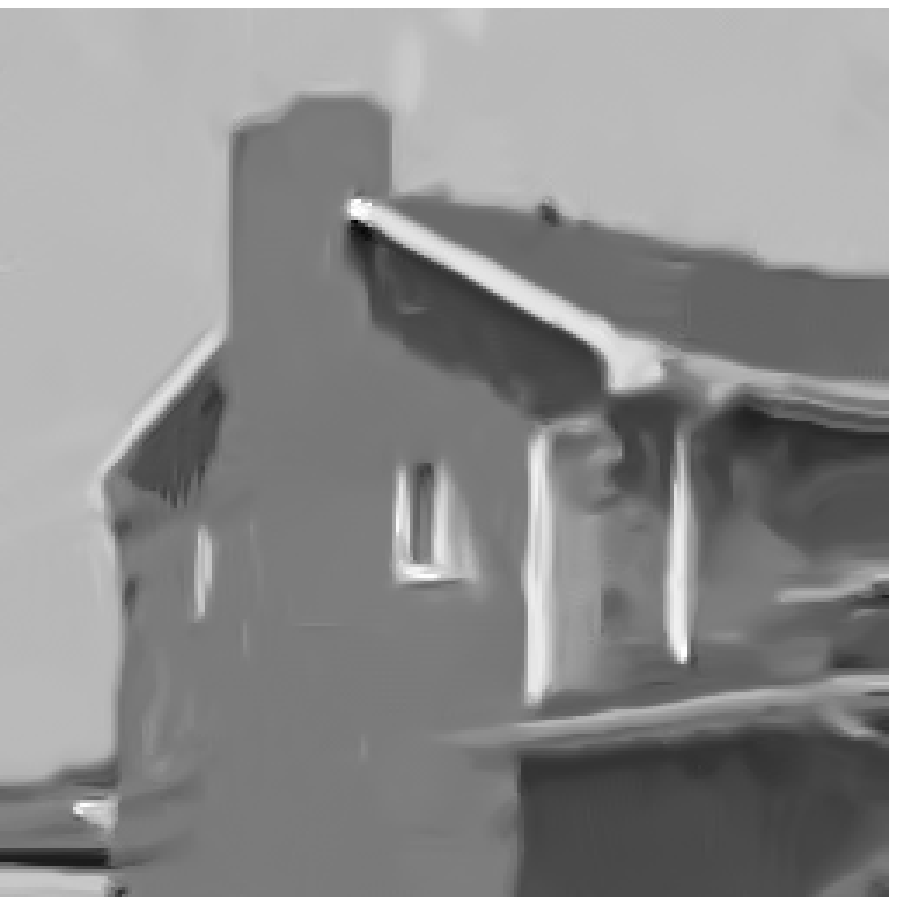}
			\caption{\centering Output with Algorithm~\ref{alg:reordering} \newline PSNR = 26.63dB}
			\label{fig:Lin-Kernighan:bm3d_NN_out}
			\vspace*{6pt}
		\end{subfigure}
		\caption{The output of our scheme with Lin-Kernighan heuristics or Algorithm~\ref{alg:reordering} for solving the TSP problem. The scheme is applied for the Gaussian denoising task with ${\sigma = 100}$, and initialized with the BM3D result.}
		\label{fig:Lin-Kernighan}
	\end{figure}

 A natural way to proceed is to use a known approximation algorithm that is known to perform well for the TSP. The problem with this approach is that the permutation induced from a too-good TSP solver is in fact creating and magnifying artifacts. An example for this behavior and the artifacts induced are shown in Figure~\ref{fig:Lin-Kernighan}, where we use the Lin-Kernighan algorithm for approximating the TSP solution~\cite{Lin_Kernighan_1973,concorde_2006}, and then apply our recovery algorithm. We should note that the two orderings (our approach, as described in Algorithm \ref{alg:reordering} and Lin-Kernighan) lead to an average total variation measure, $\sum_{k = 1}^{N - 1}|\mathbf{x}_{k+1} - \mathbf{x}_k|$, of $2.17 \times 10^{-2}$ and $1.79 \times 10^{-2}$, respectively, when assessed on the original (true) image. This implies that the better ordering leads to $17.5$\% improvement in terms of the smoothness obtained, but this does not translate to better outcome in our algorithm.
	
	\par The problem with a too-good TSP solver is that it adjusts the ordering to artifacts in the initial image, thereby creating correlation between the regularizer and the artifacts, which magnifies these artifacts.
	
	\par
\begin{figure}
		\centering
		\begin{subfigure}{0.32\textwidth}
			\includegraphics[width=\textwidth]{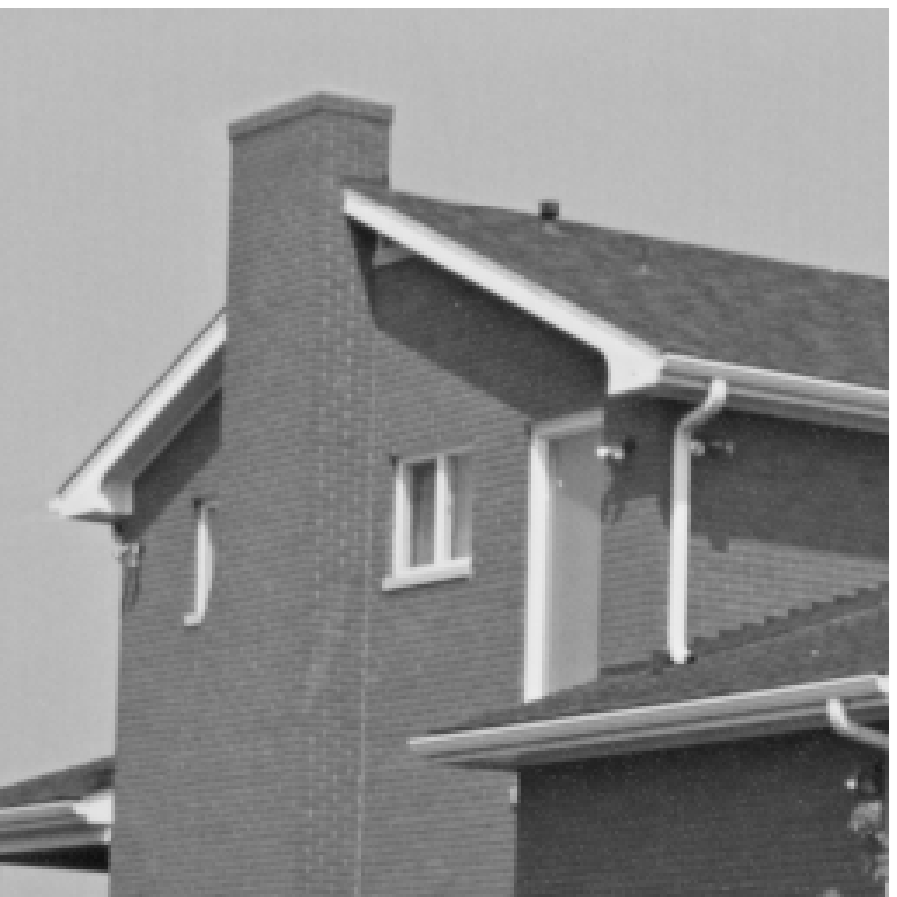}
			\caption{Original image}
			\label{fig:reordering:original}
			\vspace*{6pt}
		\end{subfigure}
		\begin{subfigure}{0.32\textwidth}
			\includegraphics[width=\textwidth]{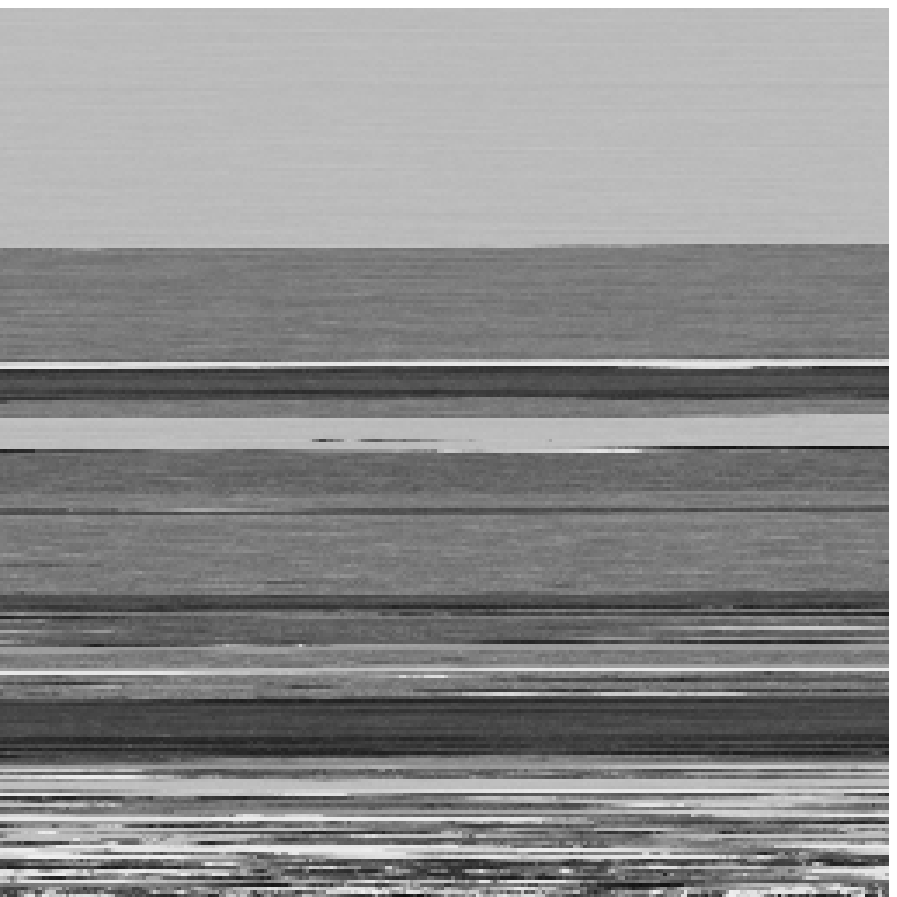}
			\caption{Reordered image}
			\label{fig:reordering:reordered}
			\vspace*{6pt}
		\end{subfigure}
		\caption{The original and the reordered \emph{House} image. Notice that the last portion of the reordered image is not smooth, due to the greedy nature of the NN heuristics in Algorithm~\ref{alg:reordering}.}
		\label{fig:reordering}
	\end{figure}

 In order to overcome this artifact magnification problem we use a randomized approach for the TSP solution. More specifically, we use the randomized version of NN (Nearest Neighbor) heuristics presented in \cite{Ram_Patch_Ordering_2013} as a TSP solver. This algorithm starts from an arbitrary patch ${\mathbf{z}_{i_0}}$ and continues by finding the closest two neighbors to the currently held vertex with the restriction that they have not been assigned yet, choosing one of them at random. At the $k$-th stage of the algorithm we have accumulated already $k$ vertices. Given the last of them, ${\mathbf{z}_{\Omega(k)}}$, we choose either ${\mathbf{z}_{i1}}$ or ${\mathbf{z}_{i2}}$, its two closest neighbors, with probabilities ${p_1 = \alpha \cdot  \exp{\left(-\|\mathbf{z}_{\Omega(k)} - \mathbf{z}_{i_1}\|_2^2/\delta\right)}}$ and ${p_2 = \alpha \cdot  \exp{\left(-\|\mathbf{z}_{\Omega(k)} - \mathbf{z}_{i_2}\|_2^2/\delta\right)}}$, where ${\alpha}$ is chosen such as ${p_1 + p_2 = 1}$. The nearest neighbor search is performed from within the set of unvisited patches, and it is limited to a window of size ${B \times B}$ around $\mathbf{z}_{\Omega{k}}$. If there is only one unvisited patch in this region, the heuristics chooses it. When no unvisited patches remain, the first and second nearest neighbor search is performed among all unvisited image patches.  The ${B \times B}$ square restriction is designed to reduce computation complexity; however it is important also for assuring that relevant patches are matched. The randomized NN heuristics is summarized in Algorithm~\ref{alg:reordering}. An example of a reordered image is presented in Figure \ref{fig:reordering}.
	
	\par A drawback of the described reordering is it's greediness nature. Indeed, as can be seen in Figure~\ref{fig:reordering}, the last part of the reordered clean image is not smooth. This is due to the fact that in the last stages of Algorithm~\ref{alg:reordering} very few unvisited patches remain for the nearest neighbor search. While this may seem troubling, in section~\ref{sec:discussion} we shall explain why this phenomenon has little effect on the final restored results. 
	
	\SetAlCapSkip{1em}
	\IncMargin{1em}
	\RestyleAlgo{boxed}
	\SetAlgoInsideSkip{medskip}
	
	\begin{algorithm}
		\SetKwInput{parameters}{Parameters}
		\SetKwInput{initialization}{Initialization}
		\flushleft
		\TitleOfAlgo{Randomized NN heuristics.}
		\parameters{We are given the image patches ${\{\mathbf{z}_i\}_{i=1}^N}$ and probability parameter~${\delta}$.}
		\initialization{Choose an arbitrary index ${i}$ and set ${\Omega(1)=\{i\}}$.} 
		\ForEach{${k = 1, \ldots, N - 1}$} {
			\begin{minipage}{0.95\hsize}
				\begin{itemize}[leftmargin=*] 
					\item[-] Set ${A_k}$ to be set of indices of ${B\times{}B}$ patches around ${\mathbf{z}_{\Omega(k)}}$. 
				\end{itemize} 
			\end{minipage}
			\flushleft
			\uIf{${|A_k \backslash \Omega| = 1}$} {
				\begin{minipage}{0.95\hsize}
					\begin{itemize}[leftmargin=*]
						\item[-] Set ${\Omega(k + 1)}$ to be ${A_k \backslash \Omega}$.
					\end{itemize}
				\end{minipage}
			} 
			\Else{
				\begin{minipage}{0.95\hsize}
					\flushleft
					\uIf{${|A_k \backslash \Omega| \geq 2}$} {
						\begin{itemize}[leftmargin=*]
							\item[-] Find ${\mathbf{z}_{i_1}}$ -- the nearest neighbor to ${\mathbf{z}_{\Omega(k)}}$, such that ${i_1 \in A_k}$ and ${i_1 \notin \Omega}$.
							\item[-] Find ${\mathbf{z}_{i_2}}$ -- the second nearest neighbor to ${\mathbf{z}_{\Omega(k)}}$, such that ${i_2 \in A_k}$ and ${i_2 \notin \Omega}$.
						\end{itemize}
					}
					\Else(\tcp*[h]{${|A_k \backslash \Omega| = 0}$}) {
						\begin{itemize}[leftmargin=*]
							\item[-] Find ${\mathbf{z}_{i_1}}$ -- the nearest neighbor to ${\mathbf{z}_{\Omega(k)}}$, such that ${i_1 \notin \Omega}$.
							\item[-] Find ${\mathbf{z}_{i_2}}$ -- the second nearest neighbor to ${\mathbf{z}_{\Omega(k)}}$, such that ${i_2 \notin \Omega}$.
						\end{itemize}
					}
					\begin{itemize}[leftmargin=*]
						\item[-] Set ${\Omega(k + 1)}$ to be:
						\begin{itemize}[leftmargin=*]
							\item[${\circ}$] ${\{i_1\}}$ with probability ${p_1 = \alpha \cdot  \exp{\left(-\frac{\|\mathbf{z}_{\Omega(k)} - \mathbf{z}_{i_1}\|_2^2}{\delta}\right)}}$,
							\item[${\circ}$] ${\{i_2\}}$ with probability  ${p_2 = \alpha \cdot \exp{\left(-\frac{\|\mathbf{z}_{\Omega(k)} - \mathbf{z}_{i_2}\|_2^2}{\delta}\right)}}$, 
							${\alpha}$ is chosen such as ${p_1 + p_2 = 1}$.
						\end{itemize}
					\end{itemize}
				\end{minipage}
			}
		}
		\KwOut{The set ${\Omega}$ holds the proposed ordering.}
		\caption{Randomized NN heuristics}
		\label{alg:reordering}
	\end{algorithm}
	\DecMargin{1em}

\section{The reconstruction algorithm}
	\label{sec:reconstruction_algorithm}
	\subsection{Constructing the Regularization Term}
		\par Our regularizer is constructed as a sum of several smoothness terms. A smoothness term is obtained by applying the matrices $P$, $L$ and $M$ to the column-stacked image $\mathbf{x}$, ${MLP\mathbf{x}}$, and penalizing the result by the robust $L_1$ norm:
		\begin{equation}
		\label{eq:mlpx_norm}
		r(\mathbf{x}) = \|MLP\mathbf{x}\|_1 \;.
		\end{equation}
		In this expression ${P}$ is the permutation matrix that represents the ordering $\Omega$ obtained by the TSP solver. ${L}$ is a simple 1D Laplacian, and ${M}$ is a weighting diagonal matrix of the form ${M = diag(m_k)}$, where
		\begin{equation}
		m_k = \min\{\frac{\gamma_k}{\beta_k}, m_{max}\} \;.
		\label{eq:mk}
		\end{equation}
		The weights $\{m_k\}_1^N$ take into account the distances between the consecutive patches in the ordering. Intuitively we expect that the centers of the closer patches will be more similar. Therefore the weights are chosen to be inversely proportional to the ${L_2}$ norm of the 1D-Laplacian of the corresponding patches. In other words, the ${\beta_k}$ coefficients are calculated using the following formula:
		\begin{equation}
		\beta_k = \frac{1}{2}\left\|2\mathbf{z}_k - \mathbf{z}_{k - 1} - \mathbf{z}_{k + 1}\right\|_2 \;,
		\label{eq:bk}
		\end{equation}
		where ${1\leq{k}\leq{N}}$, and $\{\mathbf{z}_k\}_1^N$ are the ordered patches. The boundary cases ($k = 1$, and $k = N$) are handled by setting ${\mathbf{z}_0 = \mathbf{z}_1}$, and ${\mathbf{z}_{N+1} = \mathbf{z}_N}$. We note that if $\beta_k$ would have been defined as in Equation~(\ref{eq:bk}), but assuming patches of size $1 \times 1$ pixels (i.e., only the center pixel is the actual patch), then our regularization simplifies to become the $L_0$-norm. The reason is that in this case $\beta_k$ hold simply the absolute values of the 1D Laplacian over the ordered pixels. The penalty term itself computes the very same Laplacian and divides by these $\beta_k$ values. Thus, all those ratios are '1'-es for non-zero values and '0' elsewhere, thus obtaining an $L_0$-norm. Thus, the term $r(\mathbf{x})$ in Equation~(\ref{eq:mlpx_norm}) seeks to sparsify the second derivative of the permuted image in a robust way.
		
		\par The problem with the $\{1/\beta_k\}_1^N$ coefficients is that the distances between the patches which contain edges or texture are usually relatively high, and therefore these weights are relatively low. The coefficients ${\{\gamma_k\}_1^N}$ that appear in Equation~(\ref{eq:mk}) are designed to overcome this problem, by magnifying the weights for edge/texture patches,
		\begin{equation}
		\gamma_k = 
		\begin{cases}
		\gamma_{edge} & \text{patch } \mathbf{z}_k \text{ contains edges,} \\
		1 & \mathbf{z}_k \text{ is a flat patch,}
		\end{cases}
		\end{equation}
		where $\gamma_{edge} \ge 1$ is a parameter to be set. In order to identify patches that contain edges or texture, our algorithm calculates the magnitude of image gradient using the central difference method. Namely, ${g_{i,j}}$, the magnitude of the gradient at location ($i$, $j$), is calculated using:
		\begin{equation*}
		g_{i,j} = \sqrt{\left(g_{i,j}^x\right)^2 + \left(g_{i,j}^y\right)^2} \;,
		\end{equation*}
		where ${g_{i,j}^x}$ and ${g_{i,j}^y}$ are the horizontal and vertical gradients at location ($i$, $j$). If we denote by $x_{i,j}$ value of the pixel at location ($i$, $j$), then the ${g_{i,j}^x}$ and ${g_{i,j}^y}$ are given by:
		\begin{equation*}
		g_{i,j}^x = \frac{1}{2}\left(x_{i+1,j} - x_{i-1,j}\right) \;, \quad g_{i,j}^y = \frac{1}{2}\left(x_{i,j+1} - x_{i,j-1}\right) \; .
		\end{equation*}
		A patch is identified as \emph{active} if the sum of its gradient magnitudes ${g_{i,j}}$ is above a threshold ${g_{thr}}$, i.e:
		\begin{equation}
		\gamma_k =
		\begin{cases}
		\gamma_{edge} & \sum\limits_{i,j \in \mathbf{z}_k} g_{i,j} > g_{thr}, \\
		1 & else
		\end{cases}
		\end{equation}
		Finally, the values of ${\{m_k\}_1^N}$ are clipped by ${m_{max}}$ in order to reduce the condition number of the operator MLP, thereby increasing the rate of convergence of the optimization problem.
	
	\subsection{Subimage Accumulation}
		\par
\begin{figure}
			\centering
			\begin{subfigure}{0.32\textwidth}
				\includegraphics[width=\textwidth]{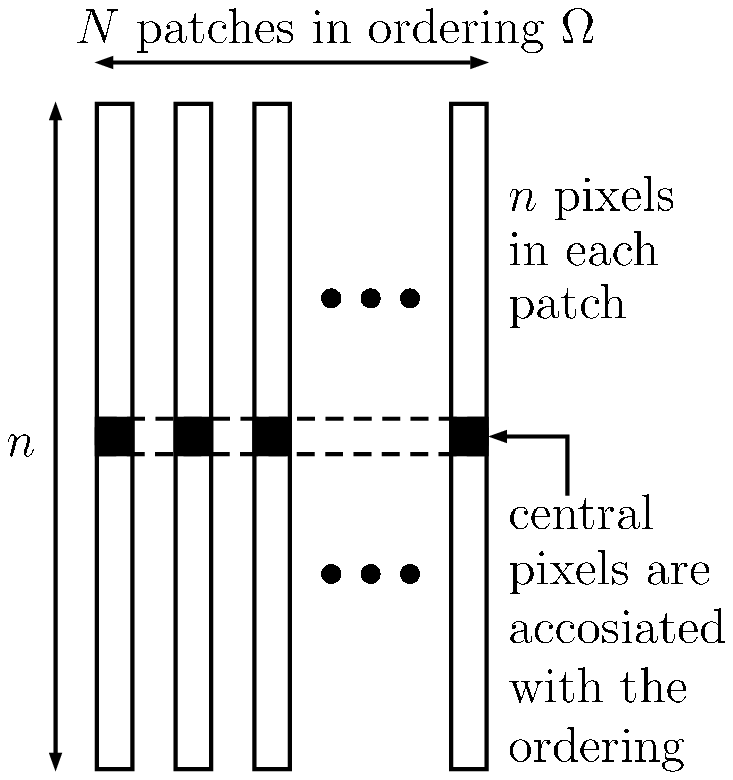}
				\caption{\tabular[t]{@{}l@{}} Associating the central \\ pixels with the ordering \endtabular}
				\label{fig:association:central}
				\vspace*{6pt}
			\end{subfigure}
			\begin{subfigure}{0.32\textwidth}
				\includegraphics[width=\textwidth]{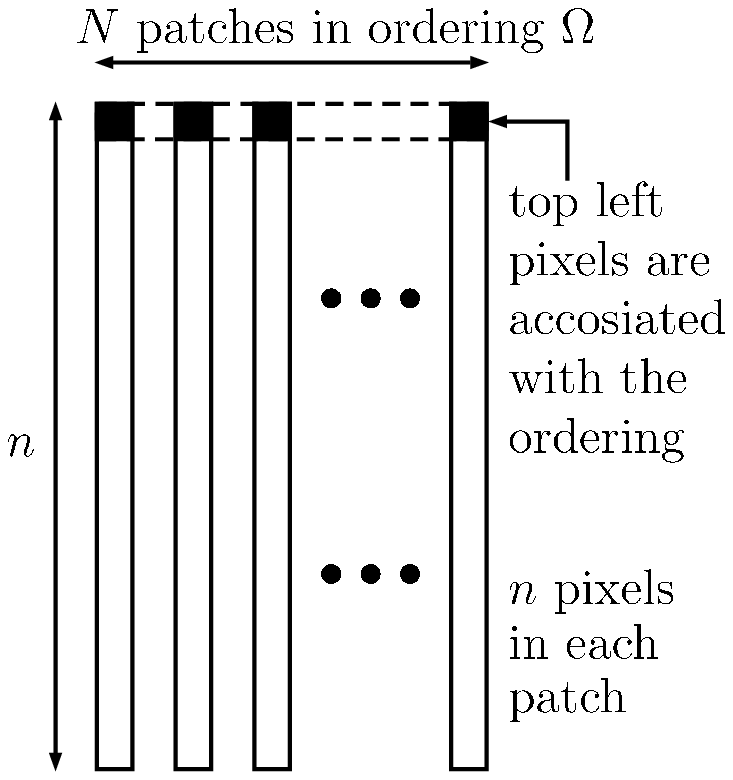}
				\caption{\tabular[t]{@{}l@{}} Associating the top-left \\ pixels with the ordering \endtabular}
				%			\caption{Associating the top-left pixels with the ordering}
				\label{fig:association:top_left}
				\vspace*{6pt}
			\end{subfigure}
			\caption{Associating pixels with the ordering -- patches are represented as n-dimensional column vectors.}
			\label{fig:association}
		\end{figure}

\begin{figure}
			\centering
			\begin{subfigure}{0.32\textwidth}
				\includegraphics[width=\textwidth]{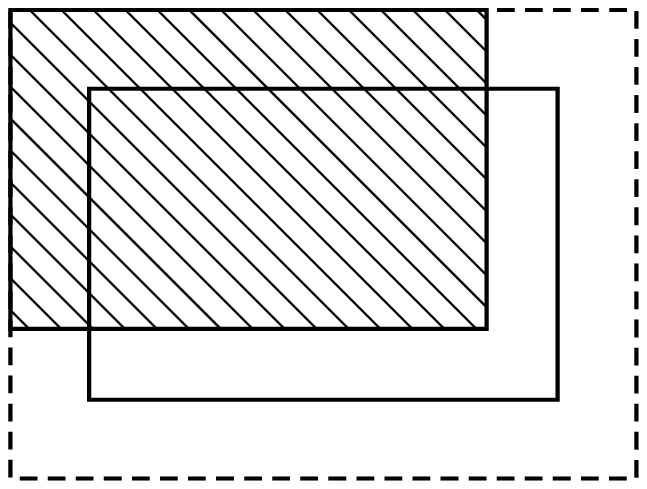}
				\caption{${i = j = 1}$}
				\label{fig:S_i_j:ij_1}
				\vspace*{6pt}
			\end{subfigure}
			\begin{subfigure}{0.32\textwidth}
				\includegraphics[width=\textwidth]{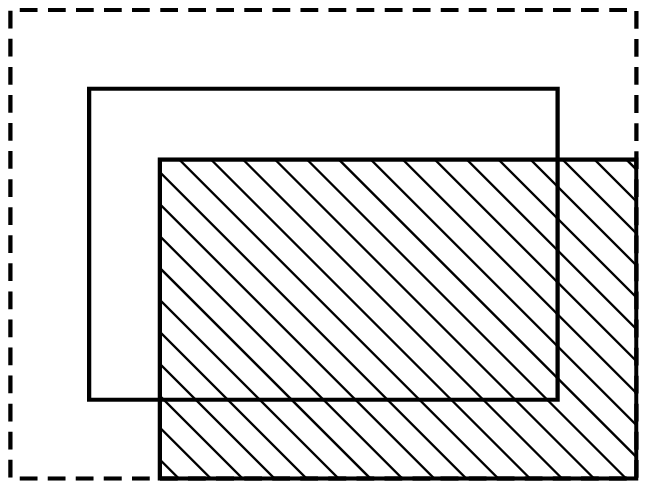}
				\caption{${i = j = \sqrt{n}}$}
				\label{fig:S_i_j:ij_n}
				\vspace*{6pt}
			\end{subfigure}
			\caption{${S_{[i,j]}}$ operator applied on the image. White solid rectangle is the original image, the dashed part corresponds to the mirror padding, and the hatched rectangle is the resulting shifted image.}
			\label{fig:S_i_j}
		\end{figure}

 In the smoothness term in Equation~(\ref{eq:mlpx_norm}) the ordering $\Omega$ is associated only with the central pixels of the patches. However, since the permutation originates from full patch ordering, it makes sense to associate the ordering with all the pixels withing the patches, as shown in Figure~\ref{fig:association}. Therefore, we propose to construct the regularizer as a sum of the smoothness terms for all pixels within the patches:
		\begin{equation}
		\label{eq:mlpsx_norm}
		r(\mathbf{x}) = \sum_{i = 1}^{\sqrt{n}} \sum_{j = 1}^{\sqrt{n}} \|MLPS_{i,j}\mathbf{x}\|_1 \;.
		\end{equation}
		The ${S_{i,j}}$ operator associates an (i,j)-shifted sub-image of $\mathbf{x}$ with the ordering ${\Omega}$. The notion of sub-images and their role here is depicted in Figure~\ref{fig:S_i_j}. First, the image is padded using mirror reflection of itself with ${\lfloor\sqrt{n}/2\rfloor}$ pixels on all sides. Then ${S_{i,j}}$ extracts an $N \times N$ sub-image starting from the \mbox{(i,j)-th} location in the padded image. For example: in Figure~\ref{fig:association:central} we are referring to the location ${i = j = \lfloor \sqrt{n}/2 \rfloor + 1}$, while in Figure~\ref{fig:association:top_left} it is ${i = j = 1}$.
		\par
\begin{figure}
			\centering
			\begin{subfigure}{0.32\textwidth}
				\includegraphics[width=\textwidth]{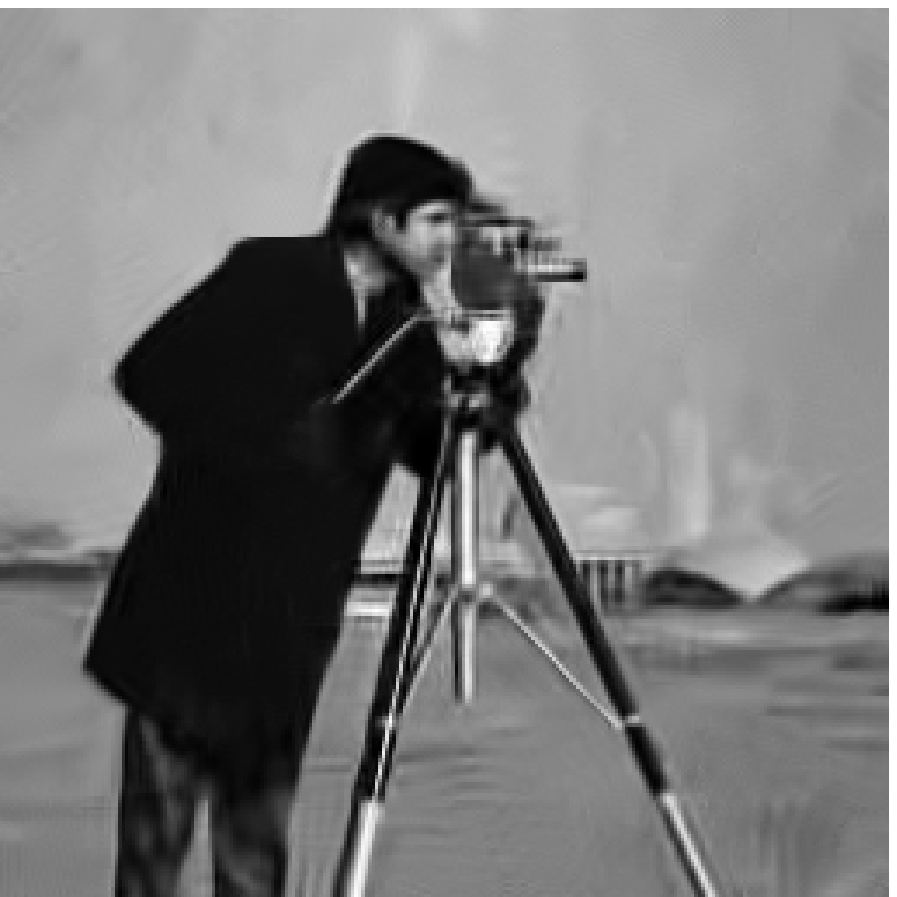}
				\caption{\centering The initial image (BM3D) \newline PSNR = 26.12dB \newline}
				\vspace*{6pt}
			\end{subfigure}
			\begin{subfigure}{0.32\textwidth}
				\includegraphics[width=\textwidth]{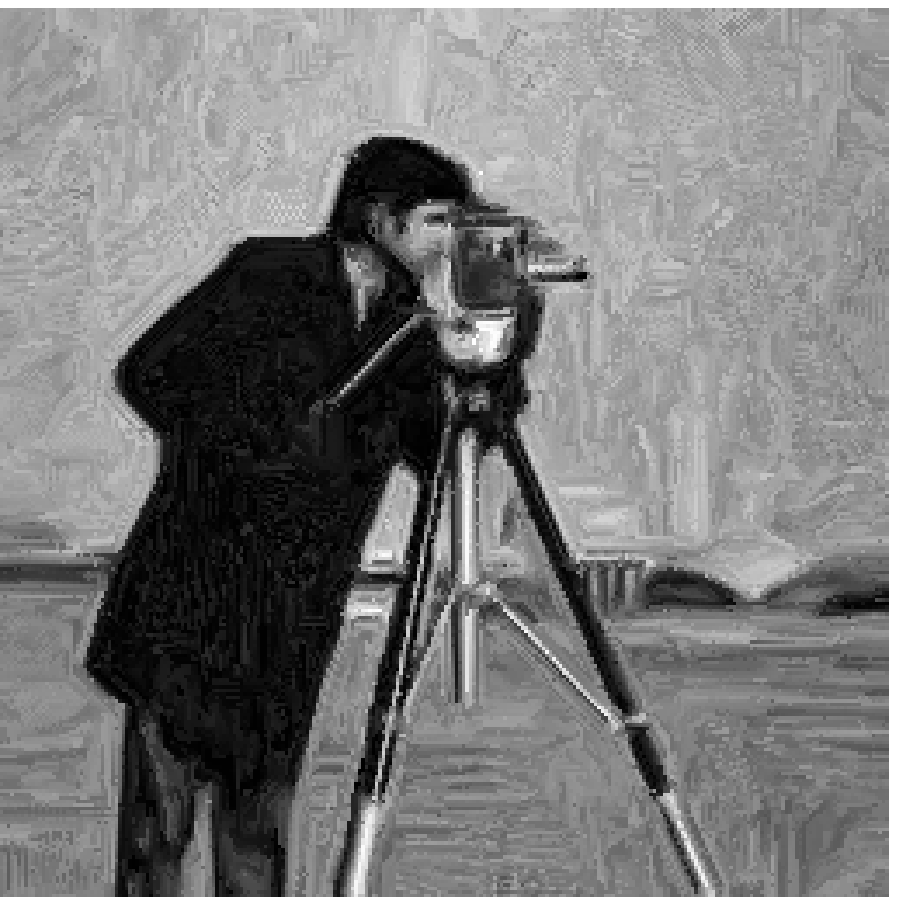}
				\caption{\tabular[t]{@{}c@{}} Output without \\ subimage accumulation \\ PSNR = 23.37dB \endtabular}
				\vspace*{6pt}
			\end{subfigure}
			\begin{subfigure}{0.32\textwidth}
				\includegraphics[width=\textwidth]{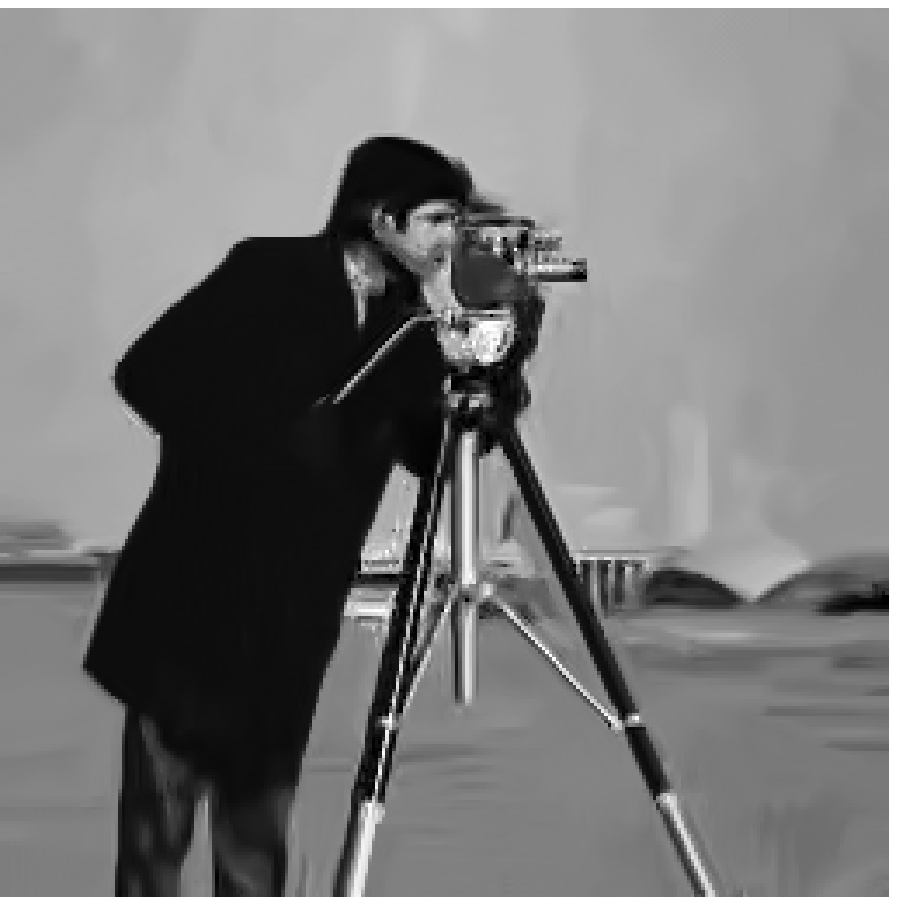}
				\caption{\tabular[t]{@{}c@{}} Output with \\ subimage accumulation \\ PSNR = 26.69dB \endtabular}
				\vspace*{6pt}
			\end{subfigure}
			\caption{The output of our restoration scheme with and without subimage accumulation. The scheme is applied to the  Gaussian denoising task with ${\sigma = 50}$, and initialized with BM3D.}
			\label{fig:no_subimages}
		\end{figure}

 Actually, accumulating the smoothness terms over patch pixels increases the number of orderings from 1 to ${n}$, and introduces an implicit spatial prior. This way the similarity is forced between the whole patches rather than only between their central pixels. An example of the output of our reconstruction scheme without subimage accumulation is shown in Figure~\ref{fig:no_subimages}. The resulting image seems very rugged due to the lack of a spatial smoothing prior.
		\par The work reported in~\cite{Ram_Patch_Ordering_2013} suggests to further increase the number of the orderings by applying the TSP solver 10 times (each time starting it from a random patch $\mathbf{z}_0$), this way creating 10 different permutation matrices. We explored this option by accumulating the regularization term in Equation~(\ref{eq:mlpsx_norm}) over a group of such permutations. For the regularization we employ here, our finding suggests that this approach has a negligible effect on the results.
		
	\subsection{Building the Objective Function}
		\par In order to build the objective function, we denote by ${\mathbf{y} \in \mathbb{R}^N}$ the column stacked version of the corrupted image, by ${\mathbf{x} \in \mathbb{R}^N}$ the desired image, and by ${r(\mathbf{x})}$ the regularization term in Equation~(\ref{eq:mlpsx_norm}). Then, the inverse problem is formulated as a weighted sum of the negative log-likelihood term ${f_L(\mathbf{x, y})}$, and the regularizer ${r(\mathbf{x})}$. In order to keep ${\mathbf{x}}$ values within the pixel boundaries ${[x_{min}, x_{max}]}$ we add to the objective function two penalties: ${p(x_{min} \cdot \mathbf{1}, \mathbf{x})}$ and ${p(\mathbf{x}, x_{max} \cdot \mathbf{1})}$. Therefore, the reconstruction algorithm is generally formulated as the following minimization problem:
		\begin{equation}
		\min \limits_{\mathbf{x}} \quad f_L(\mathbf{x, y}) + \mu \cdot r(\mathbf{x}) + p^{}(x_{min} \cdot \mathbf{1}, \mathbf{x}) + p(\mathbf{x}, x_{max} \cdot \mathbf{1}),
		\label{eq:inv_prob}
		\end{equation}
		where ${p(\mathbf{u}, \mathbf{w})}$ is defined as:
		\begin{equation}
		p(\mathbf{u}, \mathbf{w}) = c \sum_{k = 1}^{N} \left(|u_k - w_k| + u_k - w_k\right) \;.
		\end{equation}
		Each component of ${p(\mathbf{u}, \mathbf{w})}$ is greater than zero when ${u_k > w_k}$, and grows with the difference ${u_k - w_k}$. The parameter $c$ controls the strength with which we enforce this penalty.
		
		\par In order to use a derivative based optimization method for solving the obtained minimization problem, the energy function in Equation~(\ref{eq:inv_prob}) is smoothed by replacing the absolute values ${|\cdot|}$ with ${\rho(\cdot, \epsilon)}$, when
		\begin{equation}
		\rho(w, \epsilon) = \frac{w^2}{|w| + \epsilon} \;. 
		\end{equation}
		\sloppy
		In other words, ${r(\mathbf{x})}$ is replaced with its smooth version ${r_{sm}(\mathbf{x}, \epsilon)}$, and ${p(\mathbf{u}, \mathbf{w})}$ with ${p_{sm}(\mathbf{u}, \mathbf{w}, \epsilon)}$, where:
		\begin{equation}
		r_{sm}(\mathbf{x}, \epsilon) = \sum_{i = 1}^{\sqrt{n}} \sum_{j = 1}^{\sqrt{n}} \sum_{k = 1}^{N} \rho\left(\left[MLPS_{i,j}\mathbf{x}\right]_k, \epsilon\right) \;,
		\end{equation}
		\fussy
		and
		\begin{equation}
		p_{sm}(\mathbf{u}, \mathbf{w}, \epsilon) = c \sum_{k = 1}^{N} \left[\rho(u_k - w_k, \epsilon) + u_k - w_k\right] \;.
		\end{equation}
		The function ${\rho(w, \epsilon)}$ is smooth and convex (in ${w}$), since 
		\begin{equation*}
		\frac{d\rho(w)}{dw} = \frac{w |w| + 2w\epsilon}{(|w| + \epsilon)^2} \quad \text{and} \quad \frac{d^2\rho(w)}{dw} = \frac{2\epsilon^2}{(|w| + \epsilon)^3} \;.
		\end{equation*}
		In all our simulations we solve the unconstrained smoothed optimization problem using L-BFGS method~\cite{Schmidt_LBFGS_2005} implemented in minFunc~\cite{Liu_LBFGS_1989}. The minimization task runs approximately 200-300 iterations. For $256 \times 256$ images it takes around 1-2 minutes, and for $512 \times 512$ images 5-10 minutes. Our simulations ran on Intel i7 core with 16GB RAM. The code of Algorithm 1 and parts of the code of the minFunc are implemented in C. The rest of the code is implemented in Matlab.

\section{Experimental Results}
	\label{sec:results}
	\subsection{Gaussian Denoising}
		\par For the Gaussian denoising task, the smooth unconstrained optimization problem is formulated as:
		\begin{equation}
		\min \limits_{\mathbf{x}} \quad \frac{1}{2} \left\| \mathbf{x} - \mathbf{y} \right\|_2^2 + \mu \cdot r_{sm}(\mathbf{x}, \epsilon_r) + p_{sm}(\mathbf{0}, \mathbf{x}, \epsilon_p) + p_{sm}(\mathbf{x}, \mathbf{1}, \epsilon_p) \;.
		\end{equation}
		\par
\begin{figure}
			\centering
			\begin{subfigure}{0.32\textwidth}
				\includegraphics[width=\textwidth]{house_clean}
				\caption{\centering Original \emph{House} \newline }
				\vspace*{6pt}
			\end{subfigure}
			\begin{subfigure}{0.3\textwidth}
				\includegraphics[width=\textwidth]{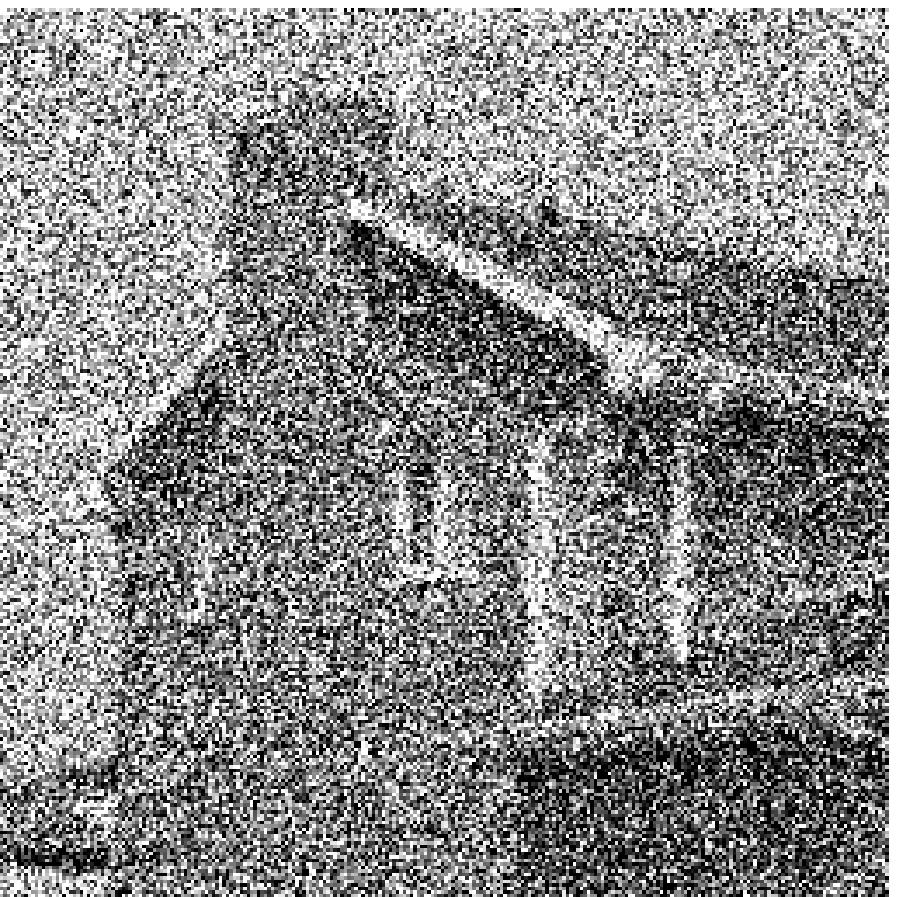}
				\caption{\centering Noisy with ${\sigma = 100}$ \newline PSNR = 8.1dB}
				\vspace*{6pt}
			\end{subfigure}
			\begin{subfigure}{0.3\textwidth}
				\includegraphics[width=\textwidth]{house100_bm3d}
				\caption{\centering The initial image (BM3D) \newline PSNR = 25.87dB}
				\vspace*{6pt}
			\end{subfigure}
			\begin{subfigure}{0.32\textwidth}
				\includegraphics[width=\textwidth]{house100_s49_NN}
				\caption{\centering Our output \newline PSNR = 26.63dB}
				\vspace*{6pt}
			\end{subfigure}
			\begin{subfigure}{0.32\textwidth}
				\includegraphics[width=\textwidth]{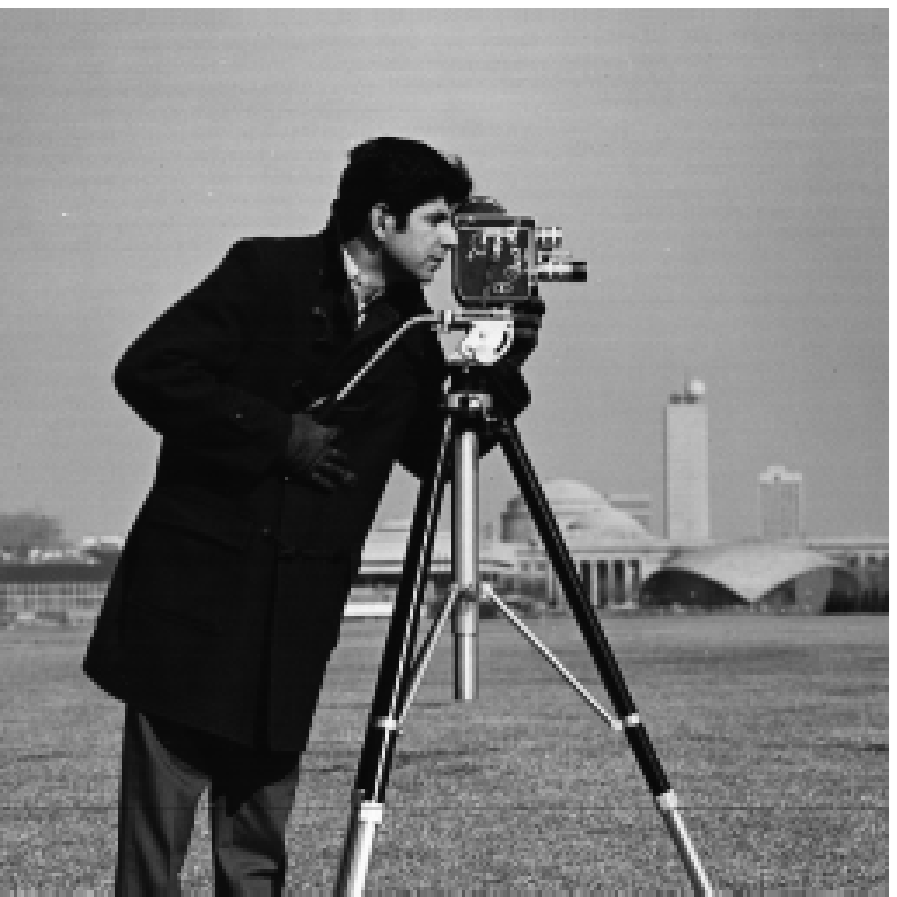}
				\caption{\centering Original \emph{C.man} \newline }
				\vspace*{6pt}
			\end{subfigure} \\
			\begin{subfigure}{0.32\textwidth}
				\includegraphics[width=\textwidth]{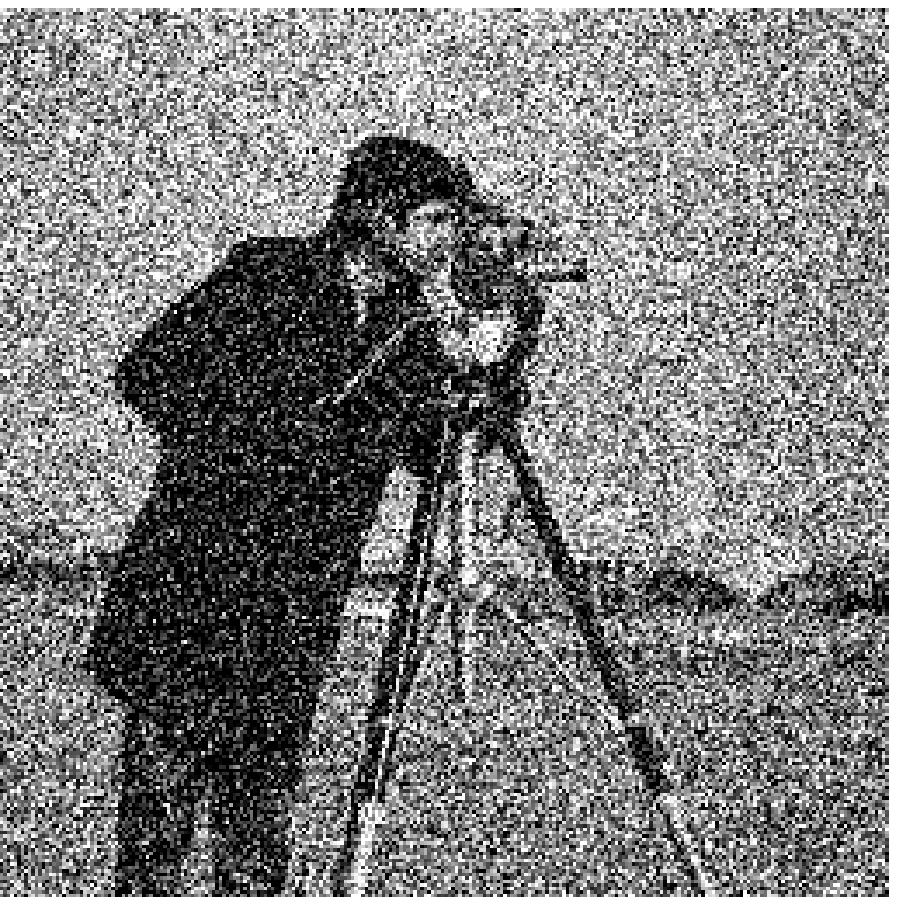}
				\caption{\centering Noisy with ${\sigma = 75}$ \newline PSNR = 10.64dB}
				\vspace*{6pt}
			\end{subfigure}
			\begin{subfigure}{0.32\textwidth}
				\includegraphics[width=\textwidth]{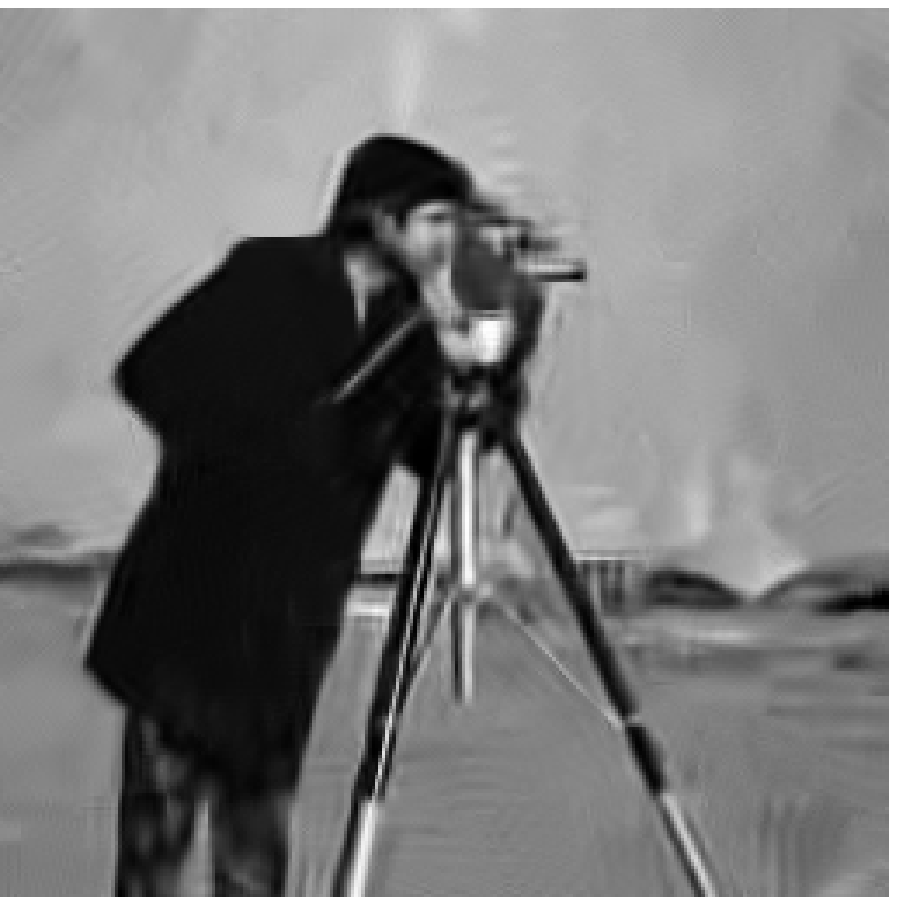}
				\caption{\centering The initial image (BM3D) \newline PSNR = 24.27dB}
				\vspace*{6pt}
			\end{subfigure}
			\begin{subfigure}{0.32\textwidth}
				\includegraphics[width=\textwidth]{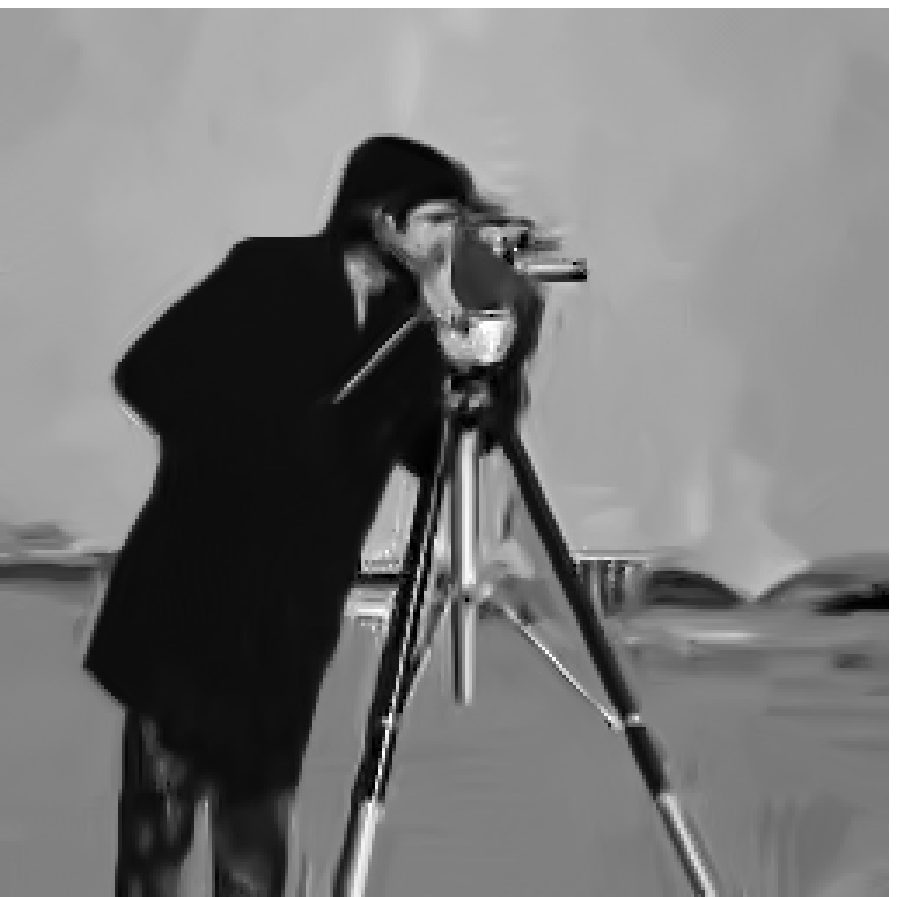}
				\caption{\centering Our output \newline PSNR = 24.95dB}
				\vspace*{6pt}
			\end{subfigure}
			\caption{Example of Gaussian denoising results for the images \emph{House} and \emph{Cameraman} with ${\sigma = 100 \text{ and } 75}$ respectively.}
			\label{fig:gauss_res}
		\end{figure}

 We run denoising experiments for noise levels ${\sigma = 25, 50, 75, \text{ and } 100}$. Our scheme is initialized with the output of the BM3D algorithm~\cite{Egiazarian_BM3D_2007}, i.e. the ordering ${\Omega}$ is calculated using the BM3D output. The simulation parameters are summarized in~\cref{tab:gauss_params_common,tab:gauss_params_test}. In~\cref{tab:gauss_res} we bring quantitative results of these experiments. For each test we compare the PSNR achieved by our scheme with the one referring to the initial images and show the improvement. Note that we do not bring SSIM measure of quality in this table, simply because the conclusions these values lead to are the same as the ones drawn from the PSNR. Examples of qualitative results are shown in Figure~\ref{fig:gauss_res}. For ${\sigma = 50}$ and higher, we get an improvement in almost all experiments. For medium noise level, for example ${\sigma = 25}$, our scheme does not succeed to improve the PSNR in most of the experiments, because the global patch ordering forces self-similarity on the image patches, and therefore it tends to eliminate distinctive details. In addition, our algorithm does not perform well on image areas that contain sensitive texture or high amount of edges. Examples of such areas are: striped pants in the Barbara image and friction ridges in the fingerprint image.
		
		\par Along with the experiments described above we also performed the Gaussian denoising experiment (with $\sigma = 75$) while replacing the $L_2$ distance by the SSIM index. We calculated the SSIM measure of small patches using the algorithm presented in~\cite{Zhou_SSIM_2004} with two differences: (\rmnum{1}) we used ${\sqrt{n} \times \sqrt{n}}$ Gaussian weighting function with $\sigma = 1.5$ for computing the local statistics (mean, variance and covariance), instead of ${11 \times 11}$ Gaussian used in~\cite{Zhou_SSIM_2004}; (\rmnum{2}) the local statistics of any pixel in the patch was calculated using only the pixels that belong to the patch. For example, for computing the local statistics of the top-left pixel of the patch, $x_{1,1}$, we used $1/4$ of the Gaussian, i.e. pixels $x_{i,j}$, where ${1 \le i,j \le \sqrt{n}/2}$. In fact, full ${\sqrt{n} \times \sqrt{n}}$ Gaussian window was used only for calculating the statistics of the central pixel of the patch. Unfortunately, this scheme led to performance deterioration when compared to the $L_2$ option. We believe that the reason for such behavior is this: SSIM ordering minimizes the distance between patches in terms of their mean, variance and covariance, while our regularizer penalizes for distance between pixels. Thus, in this case, the ordering and the regularizer are not consistent with each other. Clearly, this opens up an opportunity to redefine the regularizer to work in terms of the SSIM measure. We leave this as a future extension of our work.
		
		\begin{table}
			\footnotesize
			\renewcommand{\arraystretch}{1.3}
			\caption{Common parameters for Gaussian denoising, debluring, and super-resolution tests.}
			\label{tab:gauss_params_common}
			\centering
			\begin{tabular}{||c|c|c|c|c|c|c|c|c||}
				\hline \hline
				${\delta}$ & ${\gamma_{edge}}$ & ${m_{max}}$ & ${g_{thr}}$ 
				& ${\epsilon_r}$ & ${\epsilon_p}$ & c & ${\sqrt{n}}$ & B
				\\ \hline
				${10^6}$ & 1.5 & 20 & 3.5 & ${10^{-1}}$ & ${10^{-3}}$ & 1 & 7 & 121
				\\ \hline \hline 
			\end{tabular}
		\end{table}
		\begin{table}
			\footnotesize
			\renewcommand{\arraystretch}{1.3}
			\caption{Gaussian denoising parameters per ${\sigma}$.}
			\label{tab:gauss_params_test}
			\centering
			\begin{tabular}{||c||c|c|c|c||}
				\hline \hline
				${\sigma}$ & 25 & 50 & 75 & 100 
				\\ \hline
				${\mu \times 10^2}$ & ${2.5 / n}$ & ${5 / n}$ & ${8 / n}$ & ${12 / n}$
				\\ \hline \hline 
			\end{tabular}
		\end{table}
		
		\begin{table*}
			\footnotesize
			\renewcommand{\arraystretch}{1.3}
			\caption{Gaussian denoising results. Results are averaged over five experiments.}
			\label{tab:gauss_res}
			\centering
			\begin{tabular}{||c||c|c|c||c|c|c||}
				\hline \hline
				$\sigma$ / PSNR & 
				\multicolumn{3}{c||}{25 / 20.17} & 
				\multicolumn{3}{c||}{50 / 14.15} 
				\\ \hline
				& BM3D & Our & Improvement & BM3D & Our & Improvement 
				\\ \hline
				Cameraman 
				& 29.43 & \textbf{29.44} & 0.01 & 26.15 & \textbf{26.65} & 0.50 
				\\ \hline
				House 
				& 32.88 & \textbf{33.05} & 0.17 & 29.64 & \textbf{30.21} & 0.56 
				\\ \hline
				Peppers 
				& 30.23 & \textbf{30.38} & 0.15 & 26.70 & \textbf{27.09} & 0.39 
				\\ \hline
				Montage 
				& 32.32 & \textbf{32.59} & 0.28 & 27.79 & \textbf{28.57} & 0.78 
				\\ \hline
				Lena 
				& \textbf{32.06} & 31.96 & -0.10 & 29.01 & \textbf{29.13} & 0.12 
				\\ \hline
				Barbara 
				& \textbf{30.68} & 30.39 & -0.30 & \textbf{27.23} & 27.15 & -0.08 
				\\ \hline
				Boats 
				& \textbf{29.87} & 29.66 & -0.20 & 26.70 & \textbf{26.81} & 0.11 
				\\ \hline
				Fprint 
				& \textbf{27.71} & 27.15 & -0.56 & \textbf{24.53} & 24.22 & -0.32 
				\\ \hline
				Man 
				& \textbf{29.59} & 29.52 & -0.07 & 26.79 & \textbf{26.89} & 0.11 
				\\ \hline
				Couple 
				& \textbf{29.69} & 29.68 & -0.01 & 26.45 & \textbf{26.61} & 0.16 
				\\ \hline
				Hill 
				& \textbf{29.83} & 29.70 & -0.13 & 27.15 & \textbf{27.22} & 0.07 
				\\ \hline
				\textbf{Average}
				& \textbf{30.39} & 30.32 & -0.07 & 27.10 & \textbf{27.32} & 0.22 
				\\ \hline \hline
				$\sigma$ / PSNR & 
				\multicolumn{3}{c||}{75 / 10.63} & 
				\multicolumn{3}{c||}{100 / 8.13} 
				\\ \hline
				& BM3D & Our & Improvement & BM3D & Our & Improvement 
				\\ \hline
				Cameraman 
				& 24.36 & \textbf{25.01} & 0.65 & 23.10 & \textbf{23.73} & 0.63 
				\\ \hline
				House 
				& 27.48 & \textbf{28.18} & 0.69 & 25.90 & \textbf{26.66} & 0.77 
				\\ \hline
				Peppers 
				& 24.71 & \textbf{25.13} & 0.42 & 23.27 & \textbf{23.68} & 0.41 
				\\ \hline
				Montage 
				& 25.39 & \textbf{26.36} & 0.97 & 23.75 & \textbf{24.71} & 0.97 
				\\ \hline
				Lena 
				& 27.19 & \textbf{27.41} & 0.21 & 25.85 & \textbf{26.14} & 0.29 
				\\ \hline
				Barbara 
				& 25.15 & \textbf{25.20} & 0.05 & 23.65 & \textbf{23.74} & 0.09 
				\\ \hline
				Boats 
				& 25.01 & \textbf{25.17} & 0.16 & 23.85 & \textbf{24.03} & 0.18 
				\\ \hline
				Fprint 
				& \textbf{22.82} & 22.64 & -0.19 & \textbf{21.59} & 21.50 & -0.10 
				\\ \hline
				Man 
				& 25.29 & \textbf{25.44} & 0.15 & 24.21 & \textbf{24.38} & 0.17 
				\\ \hline
				Couple 
				& 24.71 & \textbf{24.86} & 0.14 & 23.54 & \textbf{23.65} & 0.11 
				\\ \hline
				Hill 
				& 25.63 & \textbf{25.76} & 0.13 & 24.57 & \textbf{24.71} & 0.14 
				\\ \hline
				\textbf{Average}
				& 25.25 & \textbf{25.56} & 0.31 & 23.93 & \textbf{24.27} & 0.33 
				\\ \hline \hline
			\end{tabular}
		\end{table*}
		
	\subsection{Debluring}
		\par For the debluring task the smooth unconstrained optimization problem is the following:
		\begin{equation}
			\min \limits_{\mathbf{x}}. \quad \frac{1}{2} \left\| H\mathbf{x} - \mathbf{y} \right\|_2^2 + \mu \cdot r_{sm}(\mathbf{x}, \epsilon_r) + p_{sm}(\mathbf{0}, \mathbf{x}, \epsilon_p) + p_{sm}(\mathbf{x}, \mathbf{1}, \epsilon_p),
		\end{equation}
		where ${H}$ is a blur matrix.  
		\par
\begin{figure}
			\centering
			\begin{subfigure}{0.32\textwidth}
				\includegraphics[width=\textwidth]{cameraman}
				\caption{\centering Original \emph{C.man} \newline \newline }
				\vspace*{6pt}
			\end{subfigure}
			\begin{subfigure}{0.32\textwidth}
				\includegraphics[width=\textwidth]{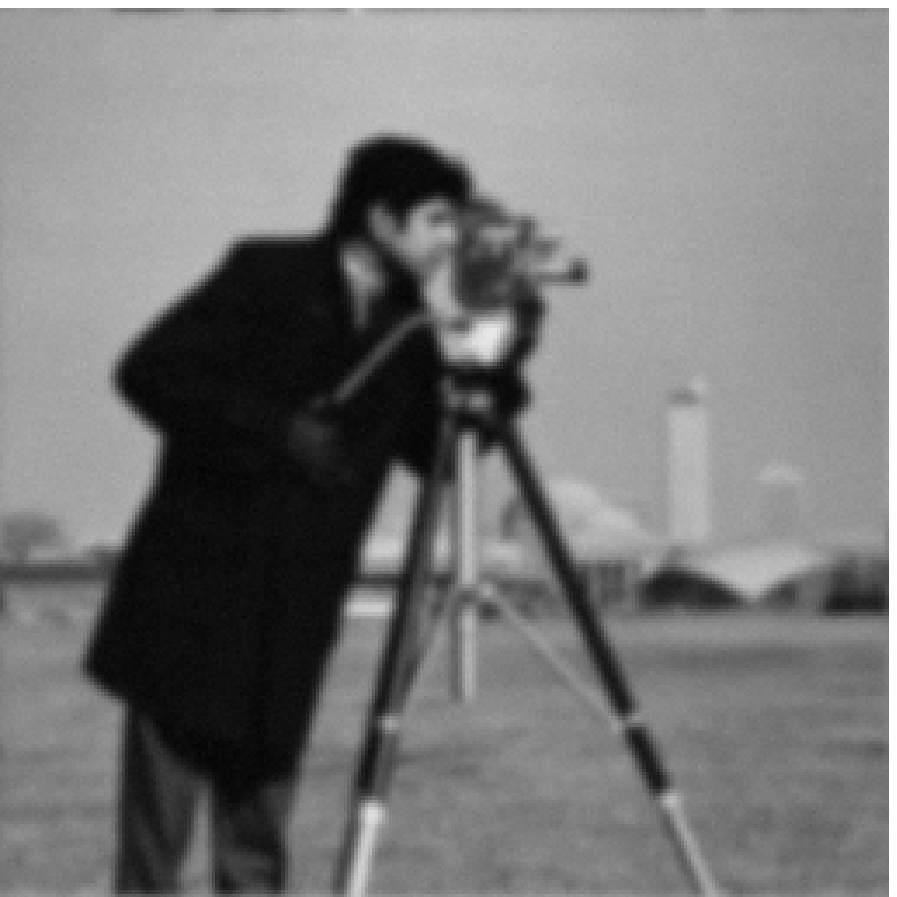}
				\caption{\centering Blurred (test 5) \newline PSNR = 23.36 \newline }
				\vspace*{6pt}
			\end{subfigure}
			\begin{subfigure}{0.32\textwidth}
				\includegraphics[width=\textwidth]{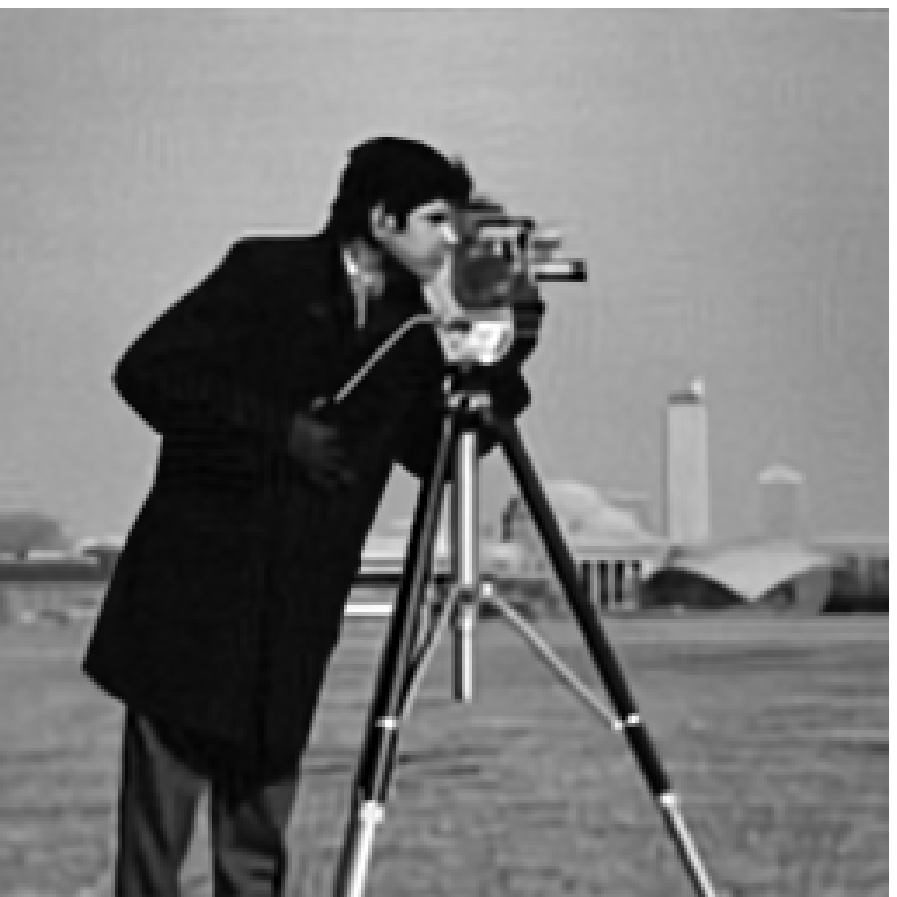}
				\caption{\centering The initial image \newline (IDD-BM3D) \newline PSNR = 27.69dB}
				\vspace*{6pt}
			\end{subfigure}
			\begin{subfigure}{0.32\textwidth}
				\includegraphics[width=\textwidth]{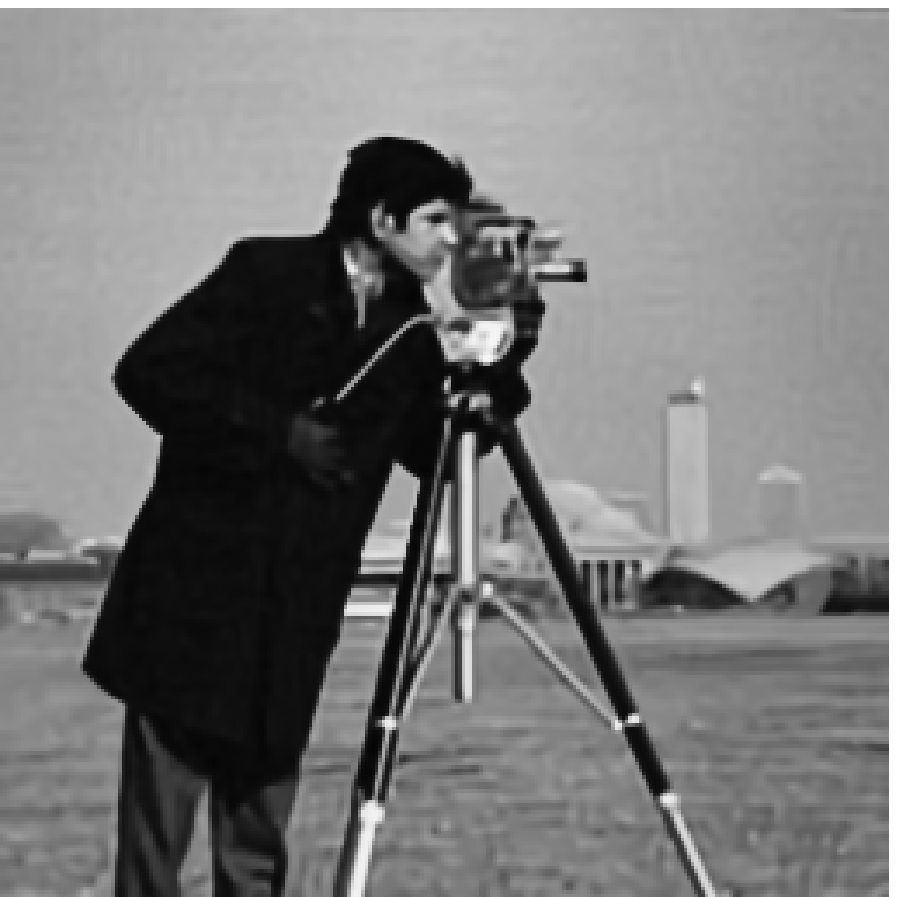}
				\caption{\centering Our output \newline PSNR = 28.50dB}
				\vspace*{6pt}
			\end{subfigure}
			\begin{subfigure}{0.32\textwidth}
				\includegraphics[width=\textwidth]{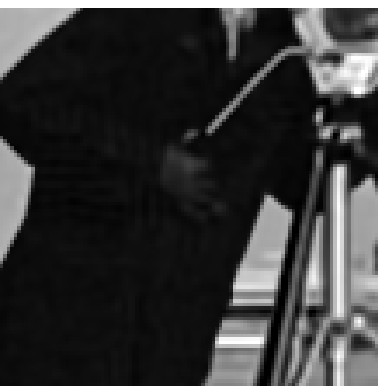}
				\caption{\centering Zoom of the initial image (IDD-BM3D)}
				\vspace*{6pt}
			\end{subfigure}
			\begin{subfigure}{0.32\textwidth}
				\includegraphics[width=\textwidth]{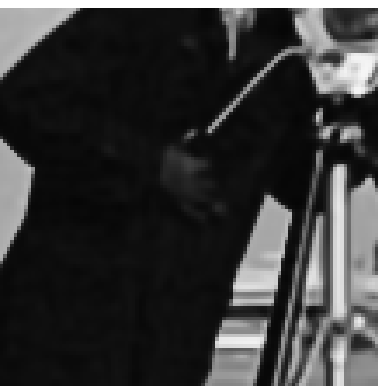}
				\caption{\centering Zoom of our output \newline}
				\vspace*{6pt}
			\end{subfigure}
			\caption{Example of debluring results of test 5 for the \emph{Cameraman} image.}
			%		\vspace*{12pt}
			\label{fig:db_res}
		\end{figure}

 We repeat the experiments reported in~\cite{Egiazarian_IDD_BM3D_2012}. The blur parameters for each experiment, PSF (point spread function) ${h(x_1, x_2)}$ and ${\sigma^2}$ of the noise, are summarized in~\cref{tab:blur_params}. The PSFs are normalized so that ${\sum h(i,j) = 1}$. Our scheme is initialized with the output of the IDD-BM3D algorithm~\cite{Egiazarian_IDD_BM3D_2012}. The simulation parameters are summarized in~\cref{tab:gauss_params_common,tab:db_params_test}. In~\cref{tab:db_res} we bring quantitative results of the deblurring experiments. For each we compare the PSNR achieved by our scheme with one of the initial images and show the improvement. Examples of qualitative results are shown in Figure~\ref{fig:db_res}. Our algorithm improved the image quality in most experiments.
		\begin{table}
			\footnotesize
			\renewcommand{\arraystretch}{1.3}
			\caption{Blur and noise variance used in each scenario.}
			\label{tab:blur_params}
			\centering
			\begin{tabular}{||c|c|c||}
				\hline \hline
				Scenario & PSF & ${\sigma^2}$
				\\ \hline
				1 & ${1/(1 + x_1^2 + x_2^2), x_1, x_2 = -7, \ldots, 7}$ & 2
				\\ \hline
				2 & ${1/(1 + x_1^2 + x_2^2), x_1, x_2 = -7, \ldots, 7}$ & 8
				\\ \hline
				3 & ${9 \times 9}$ uniform & ${\approx 0.3}$
				\\ \hline
				4 & ${[1\ 4\ 6\ 4\ 1]^T[1\ 4\ 6\ 4\ 1]/256}$ & 49
				\\ \hline
				5 & Gaussian with ${std = 1.6}$ & 4
				\\ \hline
				6 & Gaussian with ${std = 0.4}$ & 64
				\\ \hline \hline 
			\end{tabular}
		\end{table}
		\begin{table}
			\footnotesize
			\renewcommand{\arraystretch}{1.3}
			\caption{Debluring parameters per scenario}
			\label{tab:db_params_test}
			\centering
			\begin{tabular}{||c||c|c|c|c|c|c||}
				\hline \hline
				Scenario & 1 & 2 & 3 & 4 & 5 & 6   
				\\ \hline
				${\mu \times 10^5}$ & ${9 / n}$  & ${24 / n}$ & ${1.6 / n}$ & ${140 / n}$ & ${8 / n}$ & ${500 / n}$
				\\ \hline \hline 
			\end{tabular}
		\end{table}

		\begin{table*}
			\footnotesize
			\renewcommand{\arraystretch}{1.3}
			\caption{Debluring results. Results are averaged over five experiments.}
			\label{tab:db_res}
			\centering
			\begin{tabular}{||c||c||c||c|c||c|c||c|c||c|c||}
				\hline \hline
				\multirow{2}{*}{Test} & \multirow{2}{*}{Image} 
				& \multirow{2}{*}{BSNR} 
				& \multicolumn{2}{c||}{Input} 
				& \multicolumn{2}{c||}{IDD-BM3D} 
				& \multicolumn{2}{c||}{Our} 
				& \multicolumn{2}{c||}{Improvement} \\ \cline{4-11}
				& & & PSNR & SSIM & PSNR & SSIM & PSNR & SSIM & PSNR & SSIM
				\\ \hline \hline
				\multirow{5}{*}{1}
				& C.man & 31.87 & 22.23 & 0.709 & 31.11 & 0.891 
				& \textbf{31.49} & \textbf{0.904} & 0.38 & 0.013
				\\ \cline{2-11}
				& House & 29.16 & 25.61 & 0.767 & 35.54 & 0.889 
				& \textbf{36.02} & \textbf{0.896} & 0.48 & 0.007
				\\ \cline{2-11}
				& Lena & 29.89 & 27.25 & 0.882 & 35.20 & 0.972 
				& \textbf{35.63} & \textbf{0.978} & 0.43 & 0.006
				\\ \cline{2-11}
				& Barbara & 30.81 & 23.34 & 0.795 & 30.97 & 0.970
				& \textbf{31.09} & \textbf{0.974} & 0.12 & 0.003
				\\ \cline{2-11}
				& \textbf{Average} & 30.44 & 24.61 & 0.788 & 33.20 & 0.931
				& \textbf{33.56} & \textbf{0.938} & 0.35 & 0.007
				\\ \hline \hline
				\multirow{5}{*}{2}
				& C.man & 25.85 & 22.16 & 0.668 & 29.31 & 0.862
				& \textbf{29.75} & \textbf{0.873} & 0.44 & 0.011
				\\ \cline{2-11}
				& House & 23.14 & 25.46 & 0.724 & 33.99 & 0.869
				& \textbf{34.56} & \textbf{0.879} & 0.57 & 0.009
				\\ \cline{2-11}
				& Lena & 23.87 & 27.04 & 0.872 & 33.64 & 0.957
				& \textbf{34.12} & \textbf{0.966} & 0.49 & 0.008
				\\ \cline{2-11}
				& Barbara & 24.79 & 23.25 & 0.789 & 27.20 & 0.926
				& \textbf{27.29} & \textbf{0.930} & 0.08 & 0.004
				\\ \cline{2-11}
				& \textbf{Average} & 24.43 & 24.47 & 0.763 & 31.04 & 0.904  
				& \textbf{31.43} & \textbf{0.912} & 0.39 & 0.008
				\\ \hline \hline
				\multirow{5}{*}{3}
				& C.man & 40.00 & 20.77 & 0.624 & \textbf{31.24} & 0.899
				& 31.17 & \textbf{0.910} & -0.08 & 0.011
				\\ \cline{2-11}
				& House & 40.00 & 24.11 & 0.697 & 36.98 & 0.918 
				& \textbf{37.57} & \textbf{0.928} & 0.60 & 0.009
				\\ \cline{2-11}
				& Lena & 40.00 & 25.84 & 0.829 & 34.74 & 0.968
				& \textbf{35.14} & \textbf{0.973} & 0.40 & 0.004
				\\ \cline{2-11}
				& Barbara & 40.00 & 22.49 & 0.737 & \textbf{28.53} & \textbf{0.933}
				& 28.31 & 0.930 & -0.23 & -0.003
				\\ \cline{2-11}
				& \textbf{Average} & 40.00 & 23.30 & 0.722 & 32.87 & 0.930 
				& \textbf{33.05} & \textbf{0.935} & 0.18 & 0.006
				\\ \hline \hline
				\multirow{5}{*}{4}
				& C.man & 18.53 & 24.63 & 0.609 & 28.63 & 0.858
				& \textbf{29.21} & \textbf{0.873} & 0.58 & 0.014
				\\ \cline{2-11}
				& House & 15.99 & 28.08 & 0.631 & 33.85 & 0.868
				& \textbf{34.41} & \textbf{0.879} & 0.57 & 0.011
				\\ \cline{2-11}
				& Lena & 16.47 & 28.81 & 0.903 & 33.76 & 0.957 
				& \textbf{34.29} & \textbf{0.967} & 0.52 & 0.010
				\\ \cline{2-11}
				& Barbara & 17.35 & 24.22 & 0.849 & 26.09 & 0.908
				& \textbf{26.17} & \textbf{0.913} & 0.07 & 0.005
				\\ \cline{2-11}
				& \textbf{Average} & 17.17 & 26.44 & 0.748 & 30.58 & 0.898
				& \textbf{31.02} & \textbf{0.908} & 0.44 & 0.010
				\\ \hline \hline	
				\multirow{5}{*}{5}
				& C.man & 29.19 & 23.36 & 0.734 & 27.69 & 0.858
				& \textbf{28.58} & \textbf{0.873} & 0.88 & 0.015
				\\ \cline{2-11}
				& House & 26.61 & 27.82 & 0.794 & 33.55 & 0.874
				& \textbf{34.17} & \textbf{0.883} & 0.62 & 0.009
				\\ \cline{2-11}
				& Lena & 27.18 & 29.16 & 0.928 & 34.00 & 0.968
				& \textbf{34.38} & \textbf{0.973} & 0.38 & 0.005
				\\ \cline{2-11}
				& Barbara & 28.07 & 23.77 & 0.831 & 24.93 & 0.883
				& \textbf{25.01} & \textbf{0.886} & 0.07 & 0.003
				\\ \cline{2-11}
				& \textbf{Average} & 27.77 & 26.03 & 0.822 & 30.04 & 0.896
				& \textbf{30.53} & \textbf{0.904} & 0.49 & 0.008
				\\ \hline \hline
				\multirow{5}{*}{6}
				& C.man & 17.76 & 29.83 & 0.703 & 34.69 & 0.932
				& \textbf{34.89} & \textbf{0.939} & 0.21 & 0.007
				\\ \cline{2-11}
				& House & 15.15 & 30.00 & 0.682 & \textbf{37.08} & \textbf{0.920}
				& 36.74 & 0.911 & -0.34 & -0.010
				\\ \cline{2-11}
				& Lena & 15.52 & 30.02 & 0.911 & \textbf{36.34} & \textbf{0.972}
				& 36.32 & \textbf{0.972} & -0.02 & 0.000
				\\ \cline{2-11}
				& Barbara & 16.59 & 29.78 & 0.939 & \textbf{35.22} & 0.979
				& 35.21 & \textbf{0.980} & -0.01 & 0.000
				\\ \cline{2-11}
				& \textbf{Average} & 16.36 & 29.91 & 0.809 & \textbf{35.83} & \textbf{0.951}
				& 35.79 & 0.951 & -0.04 & -0.001
				\\ \hline \hline
			\end{tabular}
		\end{table*}

	\subsection{Super-Resolution}
		\par For the single image Super-Resolution (SR) task, our problem is:
		\begin{equation}
			\min \limits_{\mathbf{x}} \quad \frac{1}{2} \left\| RH\mathbf{x} - \mathbf{y} \right\|_2^2 + \mu \cdot r_{sm}(\mathbf{x}, \epsilon_r) + p_{sm}(\mathbf{0}, \mathbf{x}, \epsilon_p) + p_{sm}(\mathbf{x}, \mathbf{1}, \epsilon_p),
		\end{equation}
		where ${H}$ blurs the high resolution image with a ${7 \times 7}$ Gaussian kernel with standard deviation 1.6, and ${R}$ downsamples the image by a scaling factor 3 in each pixel. In addition, Gaussian noise with ${\sigma = 5}$ is added to the low resolution image.
		\par
\begin{figure}
			\centering
			\begin{subfigure}{0.32\textwidth}
				\includegraphics[width=\textwidth]{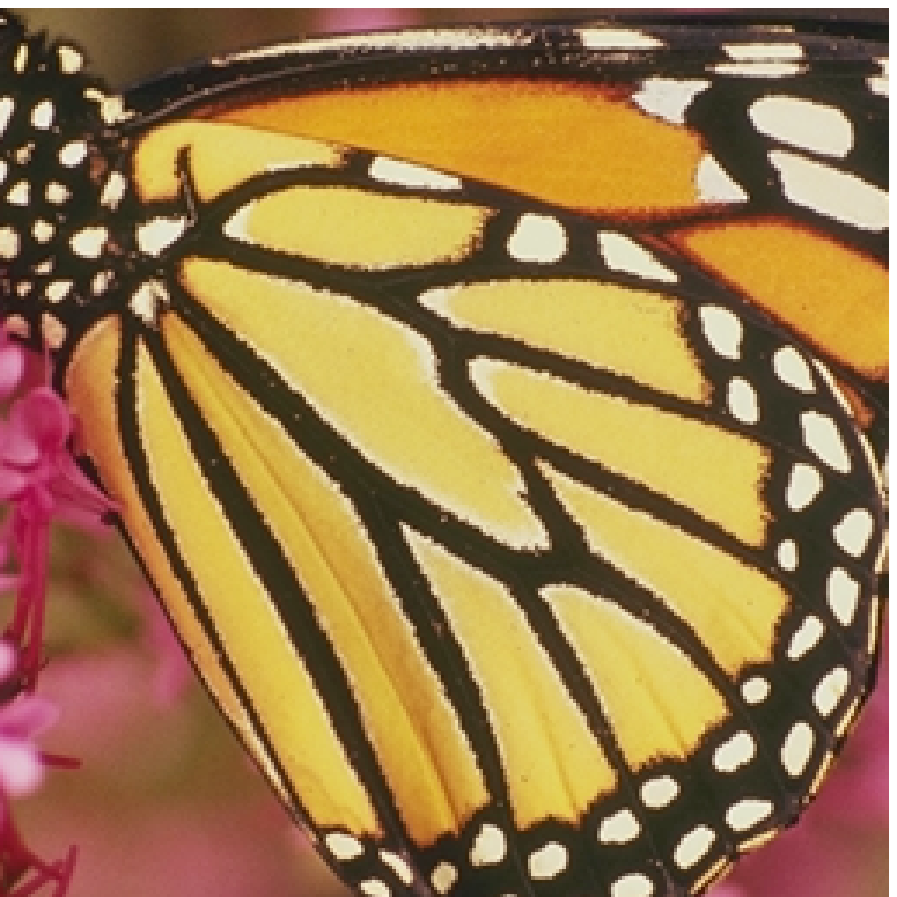}
				\caption{\centering Original \emph{Butterfly} \newline }
				\vspace*{6pt}
			\end{subfigure}
			\begin{subfigure}{0.32\textwidth}
				\includegraphics[width=\textwidth]{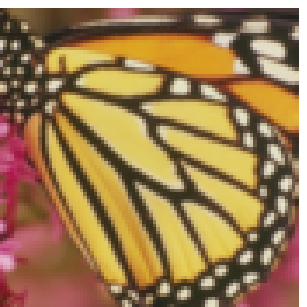}
				\caption{\centering Low resolution \newline }
				\vspace*{6pt}
			\end{subfigure}
			\begin{subfigure}{0.32\textwidth}
				\includegraphics[width=\textwidth]{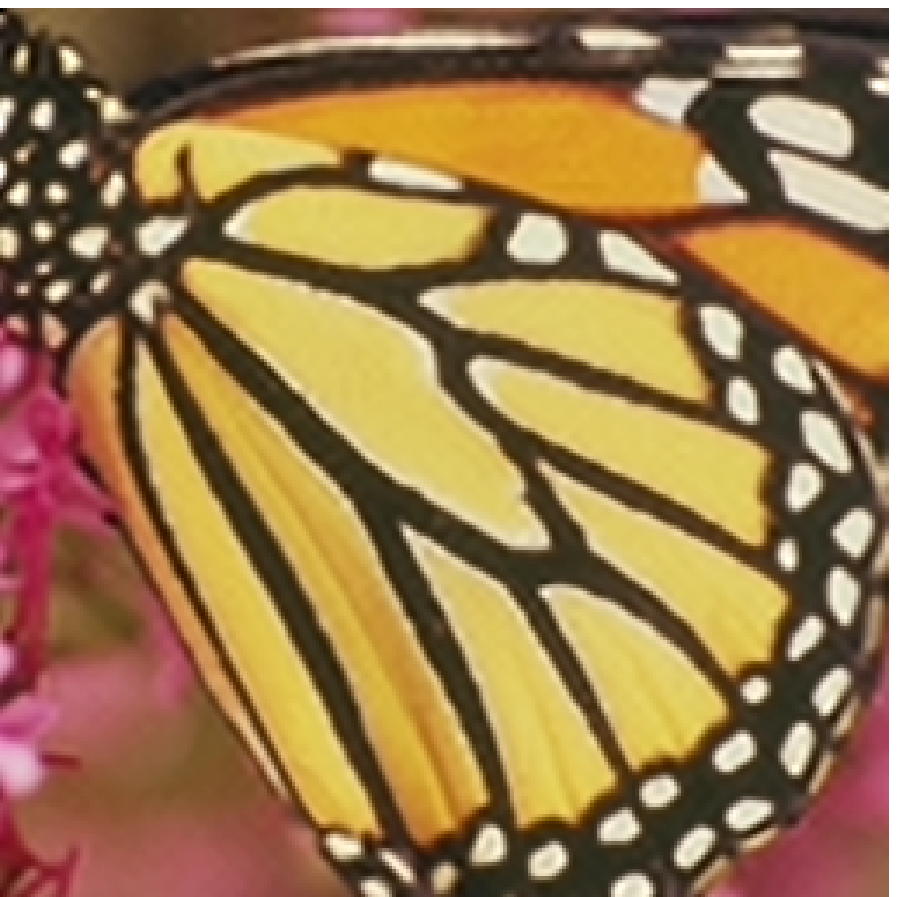}
				\caption{\centering The initial image (NSCR) \newline PSNR = 28.11dB}
				\vspace*{6pt}
			\end{subfigure}
			\begin{subfigure}{0.32\textwidth}
				\includegraphics[width=\textwidth]{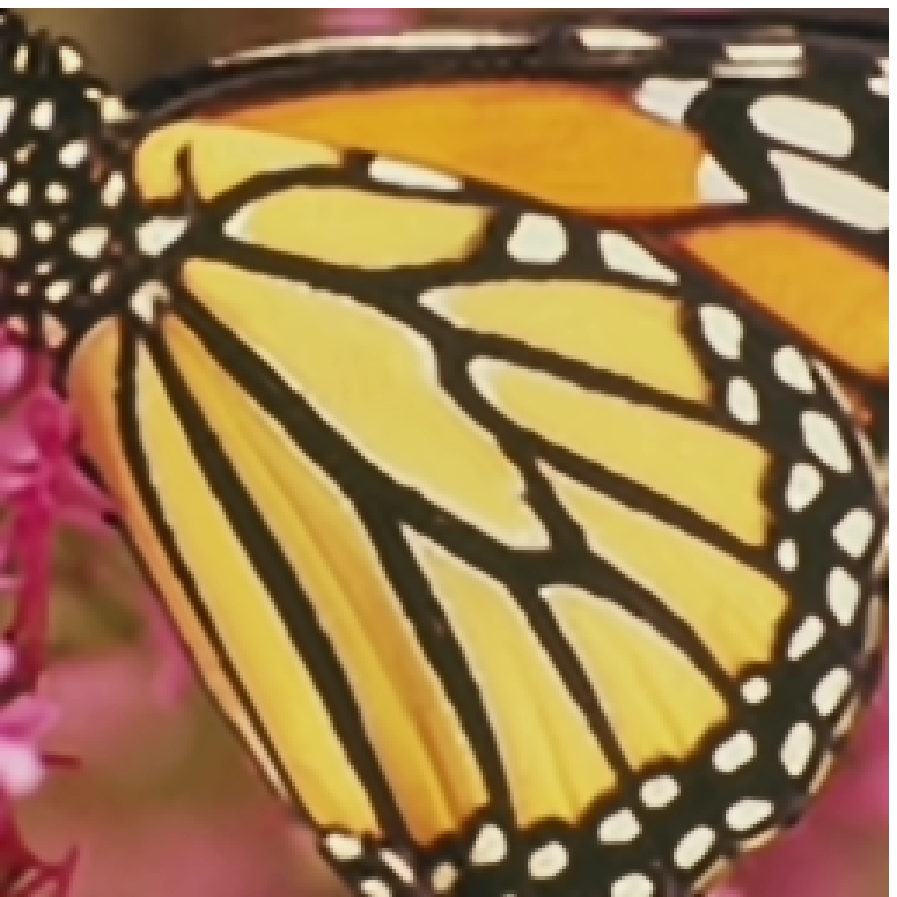}
				\caption{\centering Our output \newline PSNR = 29.41dB}
				\vspace*{6pt}
			\end{subfigure}
			\begin{subfigure}{0.32\textwidth}
				\includegraphics[width=\textwidth]{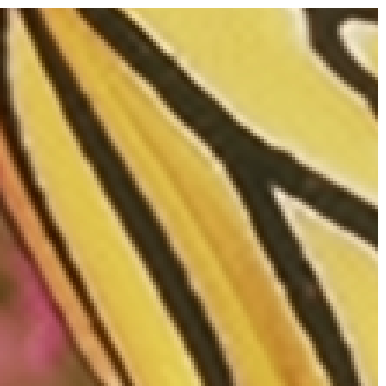}
				\caption{\centering Zoom of the initial image (NSCR)}
				\vspace*{6pt}
			\end{subfigure}
			\begin{subfigure}{0.32\textwidth}
				\includegraphics[width=\textwidth]{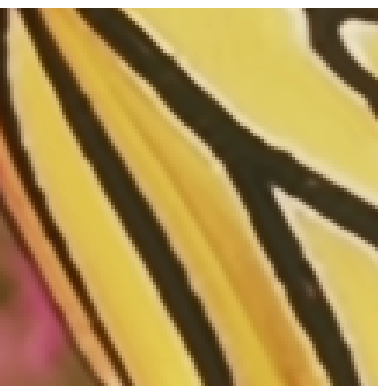}
				\caption{\centering Zoom of our output \newline}
				\vspace*{6pt}
			\end{subfigure}
			\caption{Example of super-resolution (without noise) results for the \emph{Butterfly} image.}
			\label{fig:sr_res}
		\end{figure}

 We repeat the experiments reported in ~\cite{Dong_SR_2013}, including both noiseless and noisy cases. Our scheme is initialized with output of NCSR algorithm~\cite{Dong_SR_2013}. The simulation parameters are summarized in~\cref{tab:gauss_params_common,tab:sr_params_test}. In~\cref{tab:sr_res} are shown the quantitative results of the SR experiments. Qualitative results are shown in Figure~\ref{fig:sr_res}. Our algorithm improves the image quality almost in all experiments.
		\begin{table}
			\footnotesize
			\renewcommand{\arraystretch}{1.3}
			\caption{Super-resolution parameters per test}
			\label{tab:sr_params_test}
			\centering
			\begin{tabular}{||c||c|c||}
				\hline \hline
				Test & Noiseless & Noisy
				\\ \hline
				${\mu \times 10^5}$ & ${1 / n}$ & ${9 / n}$
				\\ \hline \hline 
			\end{tabular}
		\end{table}
		
		\begin{table*}
			\footnotesize
			\renewcommand{\arraystretch}{1.3}
			\caption{Super-Resolution results. Results are averaged over five noise realizations.}
			\label{tab:sr_res}
			\centering
			\begin{tabular}{||c||c|c||c|c||c|c||}
				\hline \hline
				\multicolumn{7}{||c||}{Noiseless}
				\\ \hline \hline
				\multirow{2}{*}{Image}
				& \multicolumn{2}{c||}{NCSR} 
				& \multicolumn{2}{c||}{Our} 
				& \multicolumn{2}{c||}{Improvement} \\ \cline{2-7}
				& PSNR & SSIM & PSNR & SSIM & PSNR & SSIM
 				\\ \hline
				Butterfly & 28.11 & 0.916 & \textbf{29.43} & \textbf{0.932} & 1.32 & 0.017 
				\\ \hline
				Flower & 29.51 & 0.856 & \textbf{29.82} & \textbf{0.865} & 0.31 & 0.009
				\\ \hline
				Girl & \textbf{33.36} & \textbf{0.827} & \textbf{33.36} & 0.824 & 0.00 & -0.003
				\\ \hline
				Parthenon & 27.18 & 0.751 & \textbf{25.39} & \textbf{0.757} & 0.21 & 0.006
				\\ \hline
				Parrot & 30.52 & 0.914 & \textbf{31.02} & \textbf{0.921} & 0.50 & 0.006 
				\\ \hline
				Raccoon & 29.28 & \textbf{0.771} & \textbf{29.41} & 0.766 & 0.13 & -0.004 
				\\ \hline
				Bike & 24.75 & 0.803 & \textbf{25.15} & \textbf{0.815} & 0.40 & 0.011 
				\\ \hline
				Hat & 31.25 & 0.870 & \textbf{31.58} & \textbf{0.877} & 0.33 & 0.007
				\\ \hline
				Plants & 34.00 & 0.918 & \textbf{34.60} & \textbf{0.924} & 0.61 & 0.006 
				\\ \hline
				\textbf{Average} & 29.81 & 0.847 & \textbf{30.23} & \textbf{0.853} & 0.42 & 0.006
				\\ \hline \hline 
				
				\multicolumn{7}{||c||}{Noisy}
				\\ \hline \hline
				\multirow{2}{*}{Image}
				& \multicolumn{2}{c||}{NCSR} 
				& \multicolumn{2}{c||}{Our} 
				& \multicolumn{2}{c||}{Improvement} \\ \cline{2-7}
				& PSNR & SSIM & PSNR & SSIM & PSNR & SSIM
				\\ \hline
				Butterfly & 26.87 & 0.888 & \textbf{28.00} & \textbf{0.903} & 1.13 & 0.15
				\\ \hline
				Flower & 28.08 & 0.793 & \textbf{28.44} & \textbf{0.807} & 0.36 & 0.14
				\\ \hline
				Girl & 32.02 & 0.764 & \textbf{32.10} & \textbf{0.767} & 0.07 & 0.004
				\\ \hline
				Parthenon & 26.38 & 0.699 & \textbf{26.63} & \textbf{0.710} & 0.25 & 0.010
				\\ \hline
				Parrot & 29.51 & 0.877 & \textbf{29.86} & \textbf{0.879} & 0.35 & 0.003
				\\ \hline
				Raccoon & 28.02 & 0.681 & \textbf{28.14} & \textbf{0.689} & 0.12 & 0.008
				\\ \hline
				Bike & 23.79 & 0.737 & \textbf{24.28} & \textbf{0.760} & 0.49 & 0.023
				\\ \hline
				Hat & 29.96 & 0.824 & \textbf{30.37} & \textbf{0.830} & 0.41 & 0.006
				\\ \hline
				Plants & 31.74 & 0.859 & \textbf{32.22} & \textbf{0.866} & 0.49 & 0.006
				\\ \hline
				\textbf{Average} & 28.49 & 0.791 & \textbf{28.89} & \textbf{0.801} & 0.41 & 0.010
				\\ \hline \hline 
			\end{tabular}
		\end{table*}

	\subsection{Poisson denoising}
		The Poisson negative log-likelihood is given by the following formula:
		\begin{equation}
		f_{poiss}(\mathbf{x}, \mathbf{y}) = \sum\limits_{k=1}^{N}f_k(x_k) \;,
		\end{equation}
		where 
		\begin{equation*}
		f_k(x) = -y_k \cdot log(x) + x \;.
		\end{equation*}
		Since ${f_k(x)}$ are not defined for negative ${x}$ values, we extrapolate for ${x < \epsilon}$ using the second order Taylor series:
		\begin{equation*}
		\tilde{f}_k(x, \epsilon) = 
		\begin{cases}
		f_k(x) & x \ge \epsilon \;, \\
		f_k(\epsilon) + f_k'(\epsilon)(x - \epsilon) + \frac{1}{2}f_k''(\epsilon)(x - \epsilon)^2 & x < \epsilon \;.
		\end{cases}
		\end{equation*}		
		Then we construct the extrapolated likelihood term, ${\tilde{f}_{poiss}(\mathbf{x}, \mathbf{y}, \epsilon)}$, as a sum of the $\tilde{f}_k(x_k, \epsilon)$ functions:
		\begin{equation}
		\tilde{f}_{poiss}(\mathbf{x}, \mathbf{y}, \epsilon) = \sum\limits_{k=1}^{N}\tilde{f}_k(x_k, \epsilon).
		\end{equation}
		Since the ${\tilde{f}_k(x, \epsilon)}$ functions penalize the negative $x$ values when ${y_k > 0}$, the corresponding $k$-th components of the ${p_{sm}(\mathbf{0}, \mathbf{x}, \epsilon_p)}$ function can be omitted. Therefore, the ${p_{sm}(\mathbf{0}, \mathbf{x}, \epsilon_p)}$ function is reduced to the ${p_{sm}^{y = 0}(\mathbf{u}, \mathbf{w}, \epsilon)}$ function:
		\begin{equation}
		p_{sm}^{y = 0}(\mathbf{u}, \mathbf{w}, \epsilon) = c \sum_{k: y_k = 0} \left[\rho(u_k - w_k, \epsilon) + u_k - w_k\right] \;.
		\end{equation}
		The difference between the ${p_{sm}(\mathbf{0}, \mathbf{x}, \epsilon_p)}$  and ${p_{sm}^{y = 0}(\mathbf{u}, \mathbf{w}, \epsilon)}$ is that the ${p_{sm}^{y = 0}(\mathbf{u}, \mathbf{w}, \epsilon)}$ summarizes only over the components for which ${y_k = 0}$. Thus, the unconstrained optimization problem for the Poisson denoising is formulated by:
		\begin{equation}
		\min \limits_{\mathbf{x}} \quad \tilde{f}_{poiss}(\mathbf{x}, \mathbf{y}, \epsilon_f) + \mu \cdot r_{sm}(\mathbf{x}, \epsilon_r) + p_{sm}^{y = 0}(\mathbf{0}, \mathbf{x}, \epsilon_p) + p_{sm}(\mathbf{x}, x_{max} \cdot \mathbf{1}, \epsilon_p) \;,
		\end{equation}
		where ${x_{max}}$ is calculated using the \emph{peak} value: ${x_{max} = peak / max\_pix}$, and $max\_pix$ is the maximum pixel value of the clean image.
		\par
\begin{figure*}
			\centering
			\begin{subfigure}{0.32\textwidth}
				\includegraphics[width=\textwidth]{house_clean}
				\caption{\centering Original \emph{House} \newline }
				\vspace*{6pt}
			\end{subfigure}
			\begin{subfigure}{0.32\textwidth}
				\includegraphics[width=\textwidth]{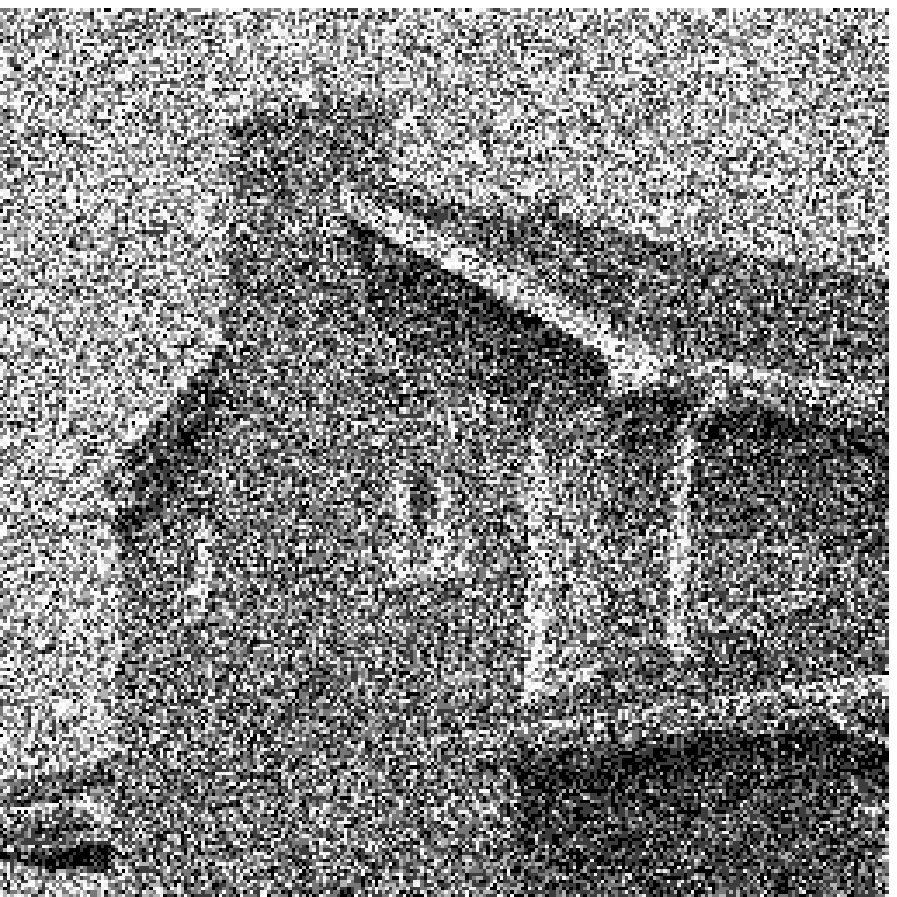}
				\caption{\centering Noisy with peak = 4 \newline PSNR = 8.40dB}
				\vspace*{6pt}
			\end{subfigure}
			\begin{subfigure}{0.32\textwidth}
				\includegraphics[width=\textwidth]{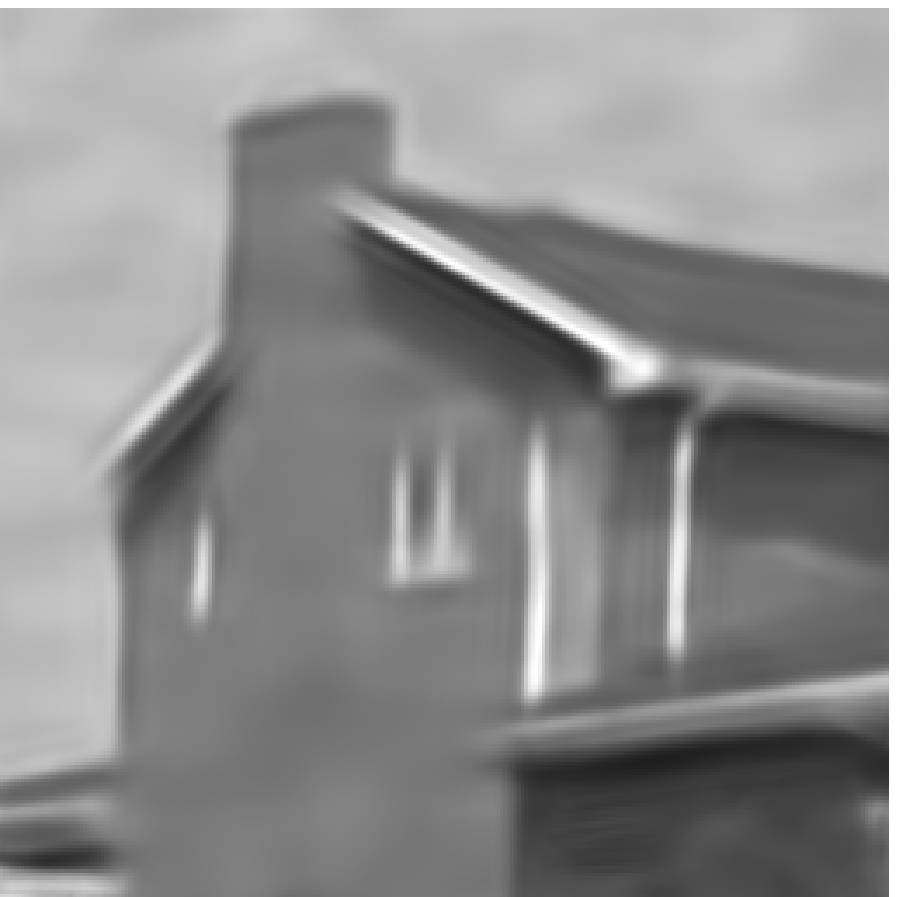}
				\caption{\centering Initialization (SPDA) \newline PSNR = 25.96dB}
				\vspace*{6pt}
			\end{subfigure}
			\begin{subfigure}{0.32\textwidth}
				\includegraphics[width=\textwidth]{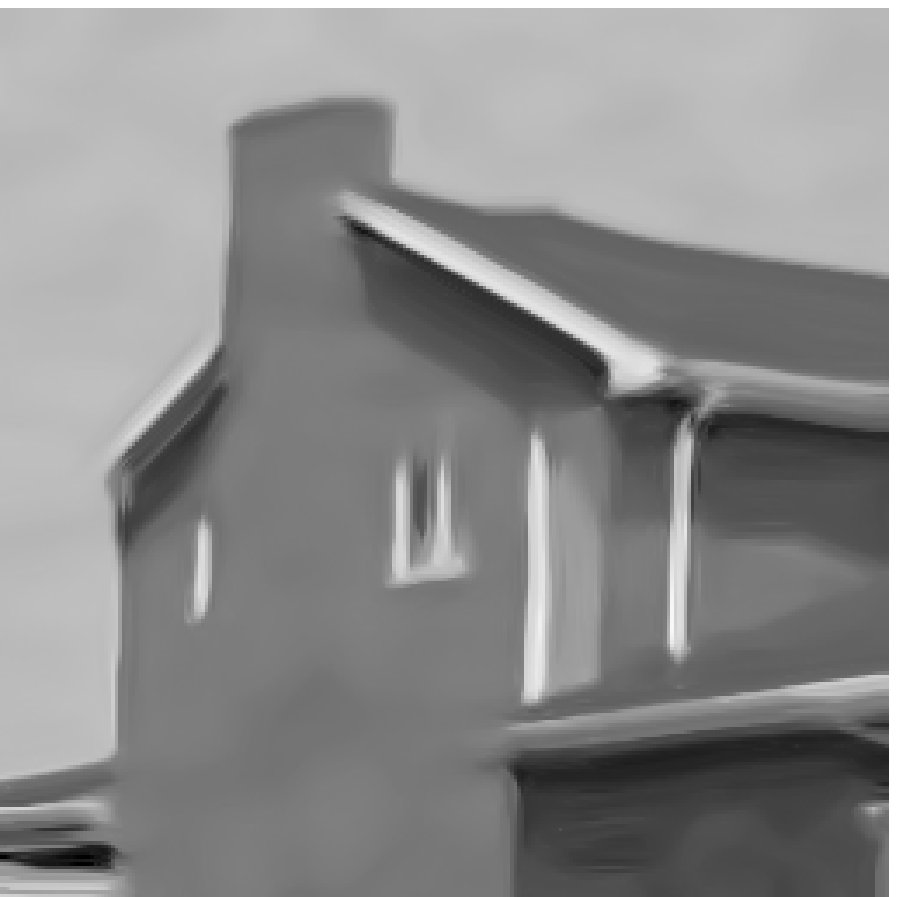}
				\caption{\centering Our output \newline PSNR = 27.00dB}
				\vspace*{6pt}
			\end{subfigure}
			\begin{subfigure}{0.32\textwidth}
				\includegraphics[width=\textwidth]{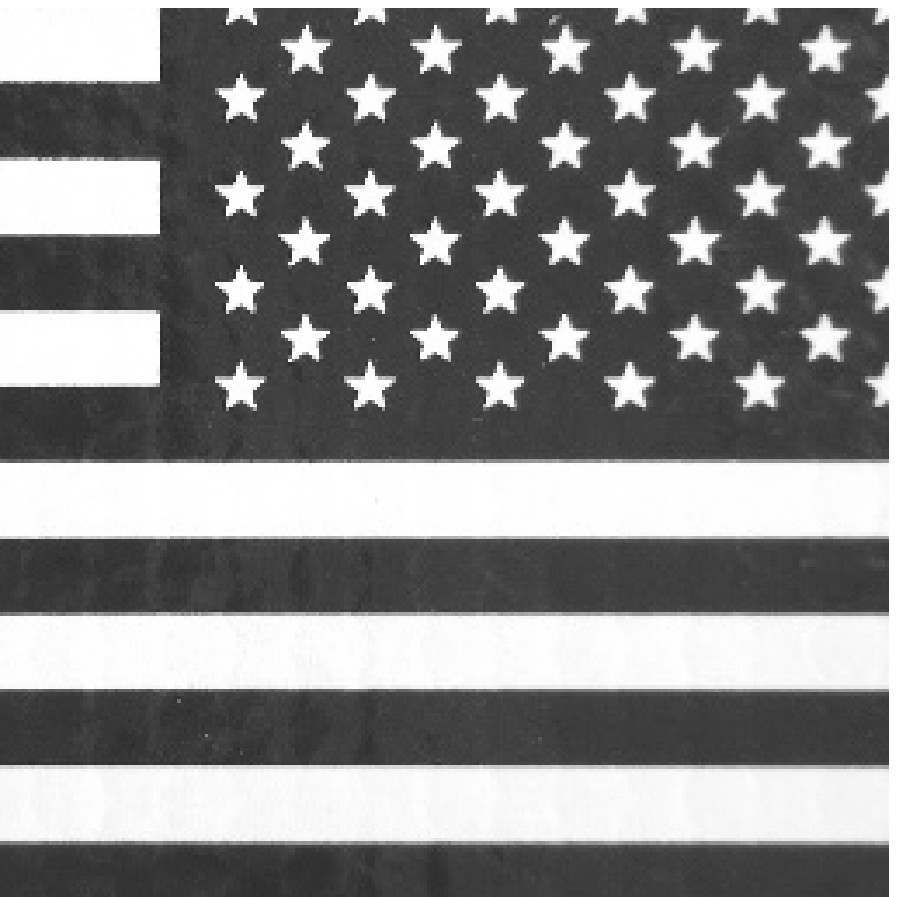}
				\caption{\centering Original \emph{Flag} \newline }
				\vspace*{6pt}
			\end{subfigure} \\
			\begin{subfigure}{0.32\textwidth}
				\includegraphics[width=\textwidth]{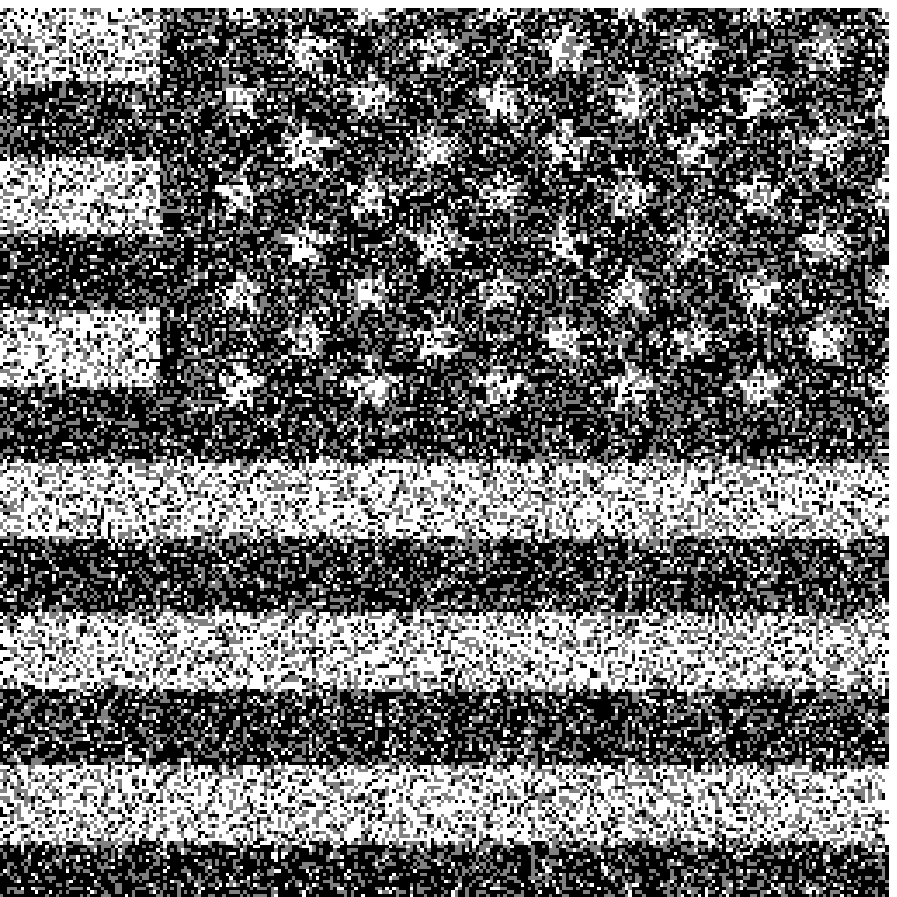}
				\caption{\centering Noisy with peak = 2 \newline PSNR = 5.90dB}
				\vspace*{6pt}
			\end{subfigure}
			\begin{subfigure}{0.32\textwidth}
				\includegraphics[width=\textwidth]{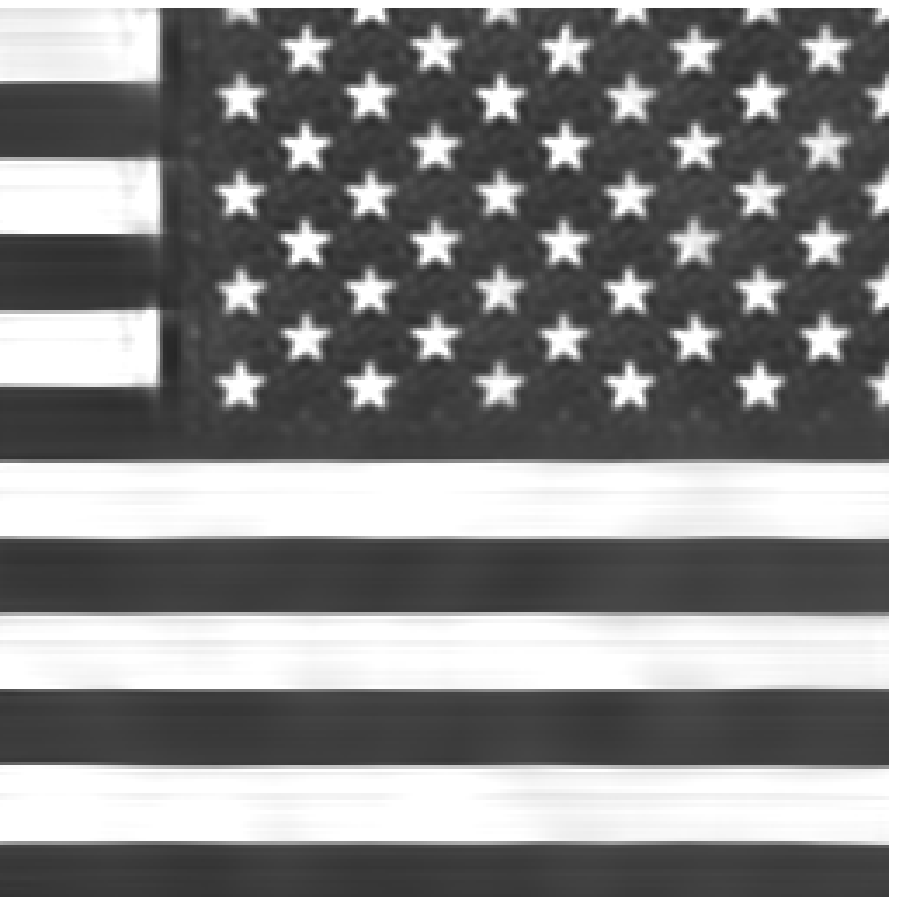}
				\caption{\centering Initialization (SPDA) \newline PSNR = 25.68dB}
				\vspace*{6pt}
			\end{subfigure}
			\begin{subfigure}{0.32\textwidth}
				\includegraphics[width=\textwidth]{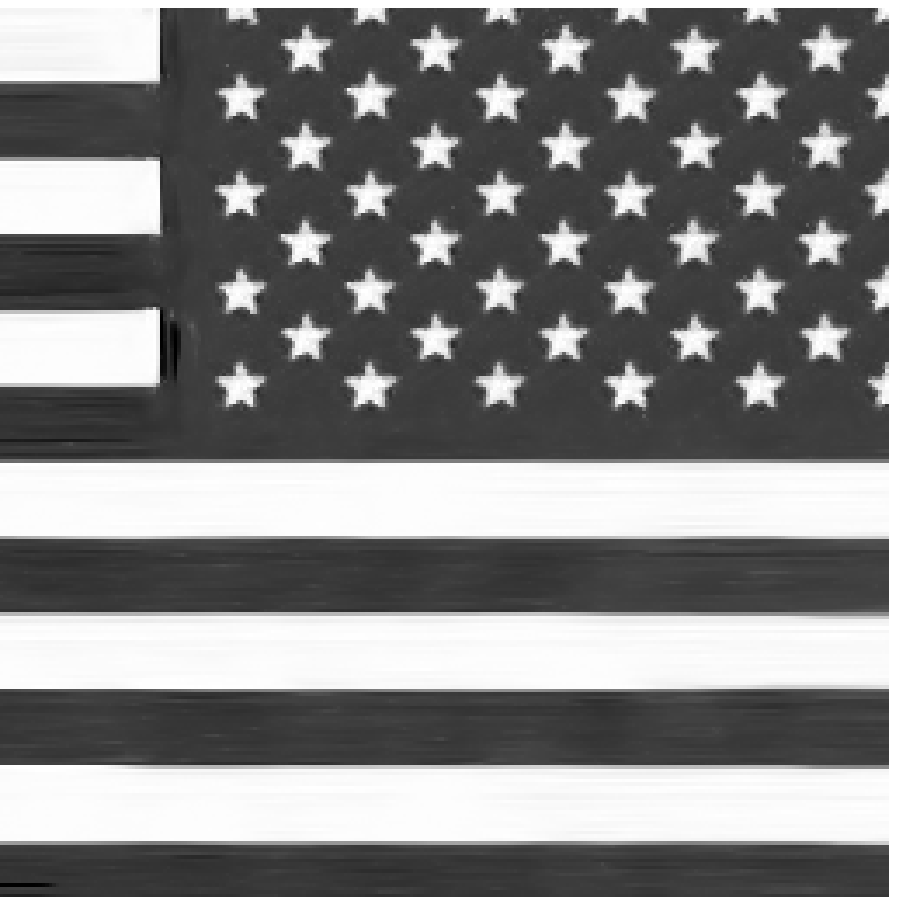}
				\caption{\centering Our output \newline PSNR = 26.47dB}
				\vspace*{6pt}
			\end{subfigure}
			\caption{Example of Poisson denoising results for the images \emph{House} and \emph{Flag} with \emph{peak} = 4 and 2 respectively.}
			\label{fig:poiss_res}
		\end{figure*}

 We run experiments for several noise levels -- \emph{peak} = 4, 2, 1, 0.5, 0.2 and 0.1. Our scheme is initialized with the output of the SPDA algorithm~\cite{Giryes_SPDA_2013}. Since SPDA  achieves better results for low peaks using binning, we also use binning in the simulations with peaks: 0.5, 0.2 and 0.1. The simulation parameters are summarized in~\cref{tab:poisson_params_common,tab:poisson_params_test}. In~\cref{tab:poiss_res} we show quantitative results of the experiments. Examples of qualitative results are shown in Figure~\ref{fig:poiss_res}. For the high peak values (peaks 4 and 2 without binning, and peak 0.5 with binning) our algorithm improves the image quality almost in all experiments.
		
		\begin{table}
			\footnotesize
			\renewcommand{\arraystretch}{1.3}
			\caption{Common Poisson denoising parameters.}
			\label{tab:poisson_params_common}
			\centering
			\begin{tabular}{||c||c|c|c|c|c|c|c|c||}
				\hline \hline
				Bin & ${\delta}$ & ${m_{max}}$ & ${\epsilon_r}$ & ${\epsilon_p}$ & ${\epsilon_f}$ & c & ${\sqrt{n}}$ & B
				\\ \hline
				no & ${10^6}$ & 5 & ${10^{-1}}$ & ${10^{-3}}$ & ${10^{-3}}$ & 1 & 9 & 201
				\\ \hline
				${3 \times 3}$ & ${10^6}$ & 5 & ${10^{-1}}$ & ${10^{-3}}$ & ${10^{-3}}$ & 1 & 7 & 101
				\\ \hline \hline 
			\end{tabular}
		\end{table}
		\begin{table}
			\footnotesize
			\renewcommand{\arraystretch}{1.3}
			\caption{Poisson denoising parameters per \emph{peak}}
			\label{tab:poisson_params_test}
			\centering
			\begin{tabular}{||c||c|c|c|c|c|c||}
				\hline \hline
				Peak & 4 & 2 & 1 & 0.5 & 0.2 & 0.1   
				\\ \hline
				${g_{thr}}$ & 20 & N/A & N/A & 10 & N/A & N/A
				\\ \hline
				${\gamma_{edge}}$ & 2.5 & 1 & 1 & 2.5 & 1 & 1
				\\ \hline
				${\mu}$ & ${0.6 / n}$ & ${0.9 / n}$ & ${1.35 / n}$ & ${0.55 / n}$ & ${0.95 / n}$ & ${1.15 / n}$
				\\ \hline \hline 
			\end{tabular}
		\end{table}
		
		\begin{table*}
			\footnotesize
			\renewcommand{\arraystretch}{1.3}
			\caption{Poisson denoising results. Results are averaged over five experiments. For each method, the first row presents the reconstruction PSNR, and the second row shows the SSIM measure.}
			\label{tab:poiss_res}
			\centering
			\begin{tabular}{||c||c||c|c|c|c|c|c|c||c||}
				\hline \hline
				%			& \multicolumn{8}{c||}{Peak = 4} 
				%			\\ \hline
				Peak & Method  
				& Bridge & Camera & Flag & House 
				& Peppers & Saturn & Swoosh & \textbf{Average} 
				\\ \hline \hline
				\multirow{6}{*}{4} 
				& \multirow{2}{*}{SPDA} 
				& 20.55 & 21.87 & 26.75 & 26.02
				& 22.02 & 31.03 & 32.67 & 25.85 
				\\ \cline{3-10}
				& & 0.367 & \textbf{0.673} & 0.883 & 0.753 
				& 0.699 & 0.853 & 0.960 & 0.741 
				\\ \cline{2-10}
				& \multirow{2}{*}{Our}
				& \textbf{20.92} & \textbf{22.84} & \textbf{28.35} & \textbf{27.15}
				& \textbf{23.68} & \textbf{31.43} & \textbf{33.71} & \textbf{26.87} 
				\\ \cline{3-10}
				& & \textbf{0.398} & 0.664 & \textbf{0.905} & \textbf{0.783} 
				& \textbf{0.726} & \textbf{0.905} & \textbf{0.969} & \textbf{0.764}
				\\ \cline{2-10}
				& \multirow{2}{*}{Impr.}
				& 0.37 & 0.97 & 1.60 & 1.13
				& 1.66 & 0.40 & 1.03 & 1.02 
				\\ \cline{3-10}
				& & 0.031 & -0.009 & 0.023 & 0.031 
				& 0.027 & 0.052 & 0.009 & 0.023 
				\\ \hline \hline
				
				\multirow{6}{*}{2} 
				& \multirow{2}{*}{SPDA} 
				& 21.14 & 21.49 & 25.41 & 25.07
				& 21.17 & 29.36 & 29.24 & 24.55 
				\\ \cline{3-10}
				& & 0.352 & \textbf{0.651} & 0.864 & 0.716 
				& 0.662 & 0.820 & 0.893 & 0.708 
				\\ \cline{2-10}
				& \multirow{2}{*}{Our}
				& \textbf{20.22} & \textbf{21.93} & \textbf{26.30} & \textbf{25.63}
				& \textbf{21.77} & \textbf{29.41} & \textbf{30.15} & \textbf{25.06} 
				\\ \cline{3-10}
				& & \textbf{0.362} & 0.605 & \textbf{0.894} & \textbf{0.749} 
				& \textbf{0.684} & \textbf{0.871} & \textbf{0.930} & \textbf{0.728} 
				\\ \cline{2-10}
				& \multirow{2}{*}{Impr.}
				& 0.08 & 0.43 & 0.89 & 0.56
				& 0.60 & 0.05 & 0.91 & 0.50 
				\\ \cline{3-10}
				& & 0.010 & -0.045 & 0.029 & 0.033 
				& 0.022 & 0.050 & 0.038 & 0.020 
				\\ \hline \hline
				
				\multirow{6}{*}{1} 
				& \multirow{2}{*}{SPDA} 
				& 19.21 & 20.17 & \textbf{22.69} & 22.62
				& 19.94 & \textbf{27.02} & 26.41 & 22.58 
				\\ \cline{3-10}
				& & 0.304 & \textbf{0.587} & \textbf{0.821} & 0.632 
				& 0.609 & 0.778 & 0.823 & 0.650 
				\\ \cline{2-10}
				& \multirow{2}{*}{Our}
				& \textbf{19.24} & \textbf{20.46} & 22.36 & \textbf{23.20}
				& \textbf{19.99} & 26.93 & \textbf{27.37} & \textbf{22.79} 
				\\ \cline{3-10}
				& & \textbf{0.315} & 0.537 & 0.819 & \textbf{0.677} 
				& \textbf{0.621} & \textbf{0.826} & \textbf{0.880} & \textbf{0.668} 
				\\ \cline{2-10}
				& \multirow{2}{*}{Impr.}
				& 0.03 & 0.28 & -0.33 & 0.57
				& 0.04 & -0.09 & 0.96 & 0.21 
				\\ \cline{3-10}
				& & 0.011 & -0.049 & -0.001 & 0.045 
				& 0.013 & 0.048 & 0.056 & 0.018 
				\\ \hline \hline
				
				\multirow{6}{*}{0.5} 
				& \multirow{2}{*}{SPDAbin} 
				& 18.58 & 18.94 & 19.39 & 21.23
				& 18.60 & 25.89 & 26.56 & 21.31 
				\\ \cline{3-10}
				& & 0.273 & 0.548 & 0.691 & 0.627
				& 0.543 & 0.756 & 0.910 & 0.621 
				\\ \cline{2-10}
				& \multirow{2}{*}{Our + bin}
				& \textbf{18.69} & \textbf{19.68} & \textbf{19.63} & \textbf{21.72} 
				& \textbf{18.92} & \textbf{26.37} & \textbf{27.55} & \textbf{21.79} 
				\\ \cline{3-10}
				& & \textbf{0.285} & \textbf{0.558} & \textbf{0.703} & \textbf{0.656} 
				& \textbf{0.571} & \textbf{0.835} & \textbf{0.933} & \textbf{0.649} 
				\\ \cline{2-10}
				& \multirow{2}{*}{Impr.}
				& 0.10 & 0.74 & 0.24 & 0.49
				& 0.32 & 0.48 & 0.98 & 0.48 
				\\ \cline{3-10}
				& & 0.012 & 0.010 & 0.012 & 0.030 
				& 0.028 & 0.079 & 0.023 & 0.028 
				\\ \hline \hline
				
				\multirow{6}{*}{0.2} 
				& \multirow{2}{*}{SPDAbin} 
				& \textbf{17.88} & 17.97 & \textbf{18.60} & 19.59
				& \textbf{17.57} & 24.03 & 23.70 & 19.91 
				\\ \cline{3-10}
				& & \textbf{0.253} & 0.478 & 0.670 & 0.521
				& 0.492 & 0.691 & 0.787 & 0.556 
				\\ \cline{2-10}
				& \multirow{2}{*}{Our + bin}
				& 17.83 & \textbf{18.29} & 18.58 & \textbf{20.16} 
				& 17.56 & \textbf{24.09} & \textbf{24.88} & \textbf{20.20} 
				\\ \cline{3-10}
				& & 0.250 & \textbf{0.500} & \textbf{0.673} & \textbf{0.607} 
				& \textbf{0.523} & \textbf{0.813} & \textbf{0.893} & \textbf{0.609} 
				\\ \cline{2-10}
				& \multirow{2}{*}{Impr.}
				& -0.05 & 0.31 & -0.02 & 0.58
				& -0.01 & 0.05 & 1.17 & 0.29 
				\\ \cline{3-10}
				& & -0.003 & 0.023 & 0.003 & 0.086 
				& 0.031 & 0.122 & 0.106 & 0.053 
				\\ \hline \hline
				
				\multirow{6}{*}{0.1} 
				& \multirow{2}{*}{SPDAbin} 
				& 17.04 & 16.80 & \textbf{16.22} & 18.78
				& \textbf{16.29} & \textbf{21.94} & 21.99 & 18.44 
				\\ \cline{3-10}
				& & 0.222 & \textbf{0.493} & 0.578 & 0.567 
				& 0.477 & 0.656 & 0.827 & 0.546 
				\\ \cline{2-10}
				& \multirow{2}{*}{Our + bin}
				& \textbf{17.10} & \textbf{17.16} & 16.03 & \textbf{18.91}
				& 16.28 & 21.72 & \textbf{22.38} & \textbf{18.51} 
				\\ \cline{3-10}
				& & \textbf{0.227} & 0.462 & \textbf{0.592} & \textbf{0.590} 
				& \textbf{0.491} & \textbf{0.787} & \textbf{0.874} & \textbf{0.575} 
				\\ \cline{2-10}
				& \multirow{2}{*}{Impr.}
				& 0.06 & 0.36 & -0.19 & 0.13
				& -0.01 & -0.11 & 0.38 & 0.07 
				\\ \cline{3-10}
				& & 0.005 & -0.030 & 0.014 & 0.024 
				& 0.014 & 0.131 & 0.047 & 0.029 
				\\ \hline \hline
				
			\end{tabular}
		\end{table*}

\pagebreak
\section{Discussion}
	\label{sec:discussion}
	\subsection{Relation to the work reported in~\cite{Ram_Patch_Ordering_2013,Ram_Patch_Ordering_Wavelet_Frame_2014}}
		\par There is an interesting connection between the approach presented in this paper and the work reported in~\cite{Ram_Patch_Ordering_2013,Ram_Patch_Ordering_Wavelet_Frame_2014}, which introduced the idea of patch-ordering. This connection is easiest to explain in the context of Gaussian denoising. If in our objective function in Equation~(\ref{eq:inv_prob}) we replace the robust statistics ($L_1$) regularization with a Euclidean norm ($L_2$), remove the penalties $p(x_{min} \cdot \mathbf{1}, \mathbf{x})$ and $p(\mathbf{x}, x_{max} \cdot \mathbf{1})$, remove the weighting matrix, and do not employ subimage accumulation, we get the following minimization problem
		\begin{equation}
		\min_{\mathbf{x}} \quad \frac{1}{2}\|\mathbf{x} - \mathbf{y}\|_2^2 + \frac{\mu}{2}\|LP\mathbf{x}\|^2_2 \;,
		\label{eq:Gauss_LPx_L2}
		\end{equation}
		which can be rewritten as
		\begin{equation}
		\min_{\mathbf{x}} \quad \frac{1}{2}\|P\mathbf{x} - P\mathbf{y}\|_2^2 + \frac{\mu}{2}\|LP\mathbf{x}\|^2_2 \;,
		\end{equation}
		since P is unitary. The solution of this problem is
		\begin{equation}
		\hat{\mathbf{x}} = P^{-1}\left(I + \mu L^TL\right)^{-1}P\mathbf{y} \;.
		\end{equation}
		In other words,
		\begin{equation}
		\hat{x} = P^{-1}HP\mathbf{y} \;,
		\end{equation}
		where $H = \left(I + \mu L^TL\right)^{-1}$ is a circulant matrix that represents a convolution filter. Therefore, the image denoising task obtained by the objective function in~(\ref{eq:Gauss_LPx_L2}) reduces to applying permutation $P$ on the noisy image $\mathbf{y}$, smoothing the permuted signal $P\mathbf{y}$ with a convolution filter $H$, and obtaining the result by applying the inverse permutation $P^{-1}$. This solution is similar to the basic idea of the scheme in~\cite{Ram_Patch_Ordering_2013}, as formulated in Equation~(2) in part~\Rmnum{2}.
		
		\par Now, lets add subimage accumulation in order to get closer to the actual scheme used in both our work and~\cite{Ram_Patch_Ordering_2013}. The minimization problem in~(\ref{eq:Gauss_LPx_L2}) transforms into
		\begin{equation}
		\min_{\mathbf{x}} \quad \frac{1}{2}\|\mathbf{x} - \mathbf{y}\|^2_2 + \frac{\mu}{2n} \sum_{i = 1}^{\sqrt{n}} \sum_{j = 1}^{\sqrt{n}} \left\| LPS_{i,j}\mathbf{x} \right\|^2_2 \;.
		\label{eq:Gauss_LPSx_L2}
		\end{equation}
		If we pad the image by a circular repetition, the $S_{i,j}$ matrices are unitary and circulant. In this case the objective function in Equation~(\ref{eq:Gauss_LPSx_L2}) can be rewritten as the sum of contributions of all subimages,
		\begin{equation}
		\min_{\mathbf{x}} \quad \frac{1}{n} \sum_{i = 1}^{\sqrt{n}} \sum_{j = 1}^{\sqrt{n}} \left( \frac{1}{2} \left\| PS_{i, j}(\mathbf{x} - \mathbf{y}) \right\|^2_2 + \frac{\mu}{2} \left\| LPS_{i, j}\mathbf{x}\right\|^2_2 \right) \;,
		\end{equation}
		where the contribution of the subimage $S_{i,j}$ is 
		\begin{equation}
		\frac{1}{2} \left\| PS_{i, j}(\mathbf{x} - \mathbf{y}) \right\|^2_2 + \frac{\mu}{2} \left\| LPS_{i, j}\mathbf{x}\right\|^2_2 \;.
		\end{equation}
		The approach taken in this work is to optimize this penalty directly (if we would have chosen to use the $L_2$ as a regularizer). In contrast, one could apply a sub-optimal minimization strategy that minimizes the contribution of each subimage independently and averages the obtained values of $\hat{\mathbf{x}}_{i,j}$. In fact, this is closely related to the approach taken in~\cite{Ram_Patch_Ordering_2013}. The result of this sub-optimal approach will be
		\begin{equation}
		\hat{\mathbf{x}}_{sub\_optimal} = \frac{1}{n} \sum_{i = 1}^{\sqrt{n}} \sum_{j = 1}^{\sqrt{n}} \hat{\mathbf{x}}_{i,j} =  \frac{1}{n} \sum_{i = 1}^{\sqrt{n}} \sum_{j = 1}^{\sqrt{n}} S^{-1}_{i,j}P^{-1}\left(I + \mu L^TL \right)^{-1}PS_{i,j}\mathbf{y} \;,
		\label{x_sub_opt_full}
		\end{equation}
		where $\hat{\mathbf{x}}_{i,j}$ minimizes contribution of the subimage $S_{i,j}$
		\begin{equation}
		\hat{\mathbf{x}}_{i,j} = S^{-1}_{i,j}P^{-1}\left(I + \mu L^TL\right)^{-1}PS_{i,j}\mathbf{y} \;.
		\end{equation}
		Equation~(\ref{x_sub_opt_full}) can be rewritten as
		\begin{equation}
		\hat{\mathbf{x}}_{sub\_optimal} = \frac{1}{n} \sum_{i = 1}^{\sqrt{n}} \sum_{j = 1}^{\sqrt{n}} S^{-1}_{i,j}P^{-1}HPS_{i,j}\mathbf{y} \;,
		\label{x_sub_opt_h}
		\end{equation}
		where $H = \left(I + \mu L^TL\right)^{-1}$ is the convolution filter we have seen before. Therefore, the formula in~(\ref{x_sub_opt_h}) reduces to the following operations: (\rmnum{1}) extract all possible subimages using the $S_{i,j}$ operators and apply permutation $P$ on each subimage, (\rmnum{2}) smooth the permuted signals with convolutional filter $H$, (\rmnum{3}) apply inverse permutation $P^{-1}$ on the results, (\rmnum{4}) plug each subimage into its original place and average the results. To conclude, the formula in Equation~(\ref{x_sub_opt_h}) is reminiscent of the denoising strategy presented in Equation~(7) of part~\Rmnum{2} in~\cite{Ram_Patch_Ordering_2013}. We should stress, however, that the work in~\cite{Ram_Patch_Ordering_2013} is not a special case of the method presented in this paper, due to various additional enhancements used in~\cite{Ram_Patch_Ordering_2013}, which are not exploited in this work. These include learning the linear filter, cycle-spinning over the choice of the permutation, and more.
		
		\par The work reported in~\cite{Ram_Patch_Ordering_Wavelet_Frame_2014}, as well as the presented work here, constructs an objective function with a permutation-based regularization. However, the scheme in~\cite{Ram_Patch_Ordering_Wavelet_Frame_2014} uses multiscale decomposition within the regularization, which makes the algorithm quite involved and computationally expensive. As a consequence, our method is simpler than the one proposed in~\cite{Ram_Patch_Ordering_Wavelet_Frame_2014}, while remaining very effective. In Table~\ref{tab:gauss_comparison_1_2_our} we compare the performance of our scheme with the algorithms presented in~\cite{Ram_Patch_Ordering_2013,Ram_Patch_Ordering_Wavelet_Frame_2014} for the Gaussian denoising problem. In Table~\ref{tab:deblur_comparison_2_our} we bring a comparison with~\cite{Ram_Patch_Ordering_Wavelet_Frame_2014} for image debluring. Tables~\ref{tab:gauss_comparison_1_2_our} and~\ref{tab:deblur_comparison_2_our} show that our method outperforms the scheme in~\cite{Ram_Patch_Ordering_2013}, and achieves results that are comparable to the ones reported in~\cite{Ram_Patch_Ordering_Wavelet_Frame_2014}.
		
		\begin{table*}
			\footnotesize
			\renewcommand{\arraystretch}{1.3}
			\caption{A comparison between the results of the proposed scheme for Gaussian denoising and the ones reported in~\cite{Ram_Patch_Ordering_2013} and~\cite{Ram_Patch_Ordering_Wavelet_Frame_2014}. The results of the new method are averaged over five experiments.}
			\label{tab:gauss_comparison_1_2_our}
			\centering
			\begin{tabular}{||c||c|c|c||c|c|c||}
				\hline \hline
				$\sigma$ / PSNR & 
				\multicolumn{3}{c||}{25 / 20.17} & 
				\multicolumn{3}{c||}{50 / 14.15} 
				\\ \hline
				& \cite{Ram_Patch_Ordering_2013} & \cite{Ram_Patch_Ordering_Wavelet_Frame_2014} & Ours & \cite{Ram_Patch_Ordering_2013} & \cite{Ram_Patch_Ordering_Wavelet_Frame_2014} & Ours 
				\\ \hline
				Lena 
				& 31.80 & \textbf{32.26} & 31.96 & 28.96 & \textbf{29.30} & 29.13 
				\\ \hline
				Barbara 
				& 30.47 & \textbf{30.90} & 30.39 & 27.35 & \textbf{27.78} & 27.15 
				\\ \hline
				Boats 
				& 29.70 & \textbf{29.88} & 29.66 & 26.69 & \textbf{26.91} & 26.81 
				\\ \hline
				Fprint 
				& \textbf{27.34} & 27.32 & 27.15 & 24.13 & 24.06 & \textbf{24.22} 
				\\ \hline
				House 
				& 32.54 & 32.37 & \textbf{33.05} & 29.64 & 29.56 & \textbf{30.21} 
				\\ \hline
				Peppers 
				& 30.01 & 30.33 & \textbf{30.38} & 26.75 & 26.93 & \textbf{27.09} 
				\\ \hline
				\textbf{Average}
				& 30.31 & \textbf{30.51} & 30.43 & 27.25 & 27.42 & \textbf{27.44} 
				\\ \hline \hline
				$\sigma$ / PSNR & 
				\multicolumn{3}{c||}{75 / 10.63} & 
				\multicolumn{3}{c||}{100 / 8.13} 
				\\ \hline
				& \cite{Ram_Patch_Ordering_2013} & \cite{Ram_Patch_Ordering_Wavelet_Frame_2014} & Ours & \cite{Ram_Patch_Ordering_2013} & \cite{Ram_Patch_Ordering_Wavelet_Frame_2014} & Ours 
				\\ \hline
				Lena 
				& 27.22 & \textbf{27.50} & 27.41 & 26.01 & \textbf{26.36} & 26.14 
				\\ \hline
				Barbara 
				& 25.42 & \textbf{25.82} & 25.20 & 24.07 & \textbf{24.46} & 23.74 
				\\ \hline
				Boats 
				& 24.99 & 25.15 & \textbf{25.17} & 23.90 & \textbf{24.04} & 24.03 
				\\ \hline
				Fprint 
				& 22.47 & 22.47 & \textbf{22.64} & 21.44 & \textbf{21.53} & 21.50 
				\\ \hline
				House 
				& 27.79 & 27.37 & \textbf{28.18} & 26.30 & 25.98 & \textbf{26.66} 
				\\ \hline
				Peppers 
				& 24.72 & 24.98 & \textbf{25.13} & 23.21 & 23.56 & \textbf{23.68} 
				\\ \hline
				\textbf{Average}
				& 25.44 & 25.55 & \textbf{25.62} & 24.16 & \textbf{24.32} & 24.29 
				\\ \hline \hline
			\end{tabular}
		\end{table*}
		
		\begin{table*}
			\footnotesize
			\renewcommand{\arraystretch}{1.3}
			\caption{A comparison between the results of the proposed scheme for image debluring and the ones reported in~\cite{Ram_Patch_Ordering_Wavelet_Frame_2014}. Results of our scheme are averaged over five experiments.}
			\label{tab:deblur_comparison_2_our}
			\centering
			\begin{tabular}{||c|c|c|c|c|c|c|c||}
				\hline \hline
				Image & Method & Test 1 & Test 2 & Test 3 & Test 4 & Tests 5 & Test 6
				\\ \hline \hline
				\multirow{2}{*}{Lena} 
				& \cite{Ram_Patch_Ordering_Wavelet_Frame_2014}
				& \textbf{8.56} & 6.92 & 8.86 & \textbf{5.52} & 4.95 & \textbf{6.91}
				\\ \cline{2-8}
				& ours 
				& 8.38 & \textbf{7.08} & \textbf{9.30} & 5.48 & \textbf{5.22} & 6.30
				\\ \hline
				\multirow{2}{*}{Barbara} 
				& \cite{Ram_Patch_Ordering_Wavelet_Frame_2014}
				& \textbf{8.06} & \textbf{4.57} & \textbf{6.01} & \textbf{2.20} & \textbf{1.41} & \textbf{6.06}
				\\ \cline{2-8}
				& ours 
				& 7.75 & 4.04 & 5.82 & 1.95 & 1.24 & 5.43
				\\ \hline
				\multirow{2}{*}{House} 
				& \cite{Ram_Patch_Ordering_Wavelet_Frame_2014}
				& \textbf{10.44} & 8.79 & 13.11 & \textbf{6.38} & 5.95 & \textbf{7.56}
				\\ \cline{2-8}
				& ours
				& 10.41 & \textbf{9.10} & \textbf{13.46} & 6.33 & \textbf{6.35} & 6.74
				\\ \hline
				\multirow{2}{*}{C.man} 
				& \cite{Ram_Patch_Ordering_Wavelet_Frame_2014}
				& 9.24 & 7.38 & 10.21 & 4.34 & 4.68 & \textbf{5.26}
				\\ \cline{2-8}
				& ours 
				& \textbf{9.26} & \textbf{7.59} & \textbf{10.40} & \textbf{4.58} & \textbf{5.22} & 5.06
				\\ \hline
				\multirow{2}{*}{\textbf{Average}} 
				& \cite{Ram_Patch_Ordering_Wavelet_Frame_2014}
				& \textbf{9.08} & 6.92 & 9.55 & \textbf{4.61} & 4.25 & \textbf{6.45}
				\\ \cline{2-8}
				& ours 
				& 8.95 & \textbf{6.96} & \textbf{9.75} & 4.58 & \textbf{4.50} & 5.88
				\\ \hline \hline	
			\end{tabular}
		\end{table*}

		\subsection{Choosing the TSP solver}
		\par
\begin{figure}
			\centering
			\begin{subfigure}{0.32\textwidth}
				\includegraphics[width=\textwidth]{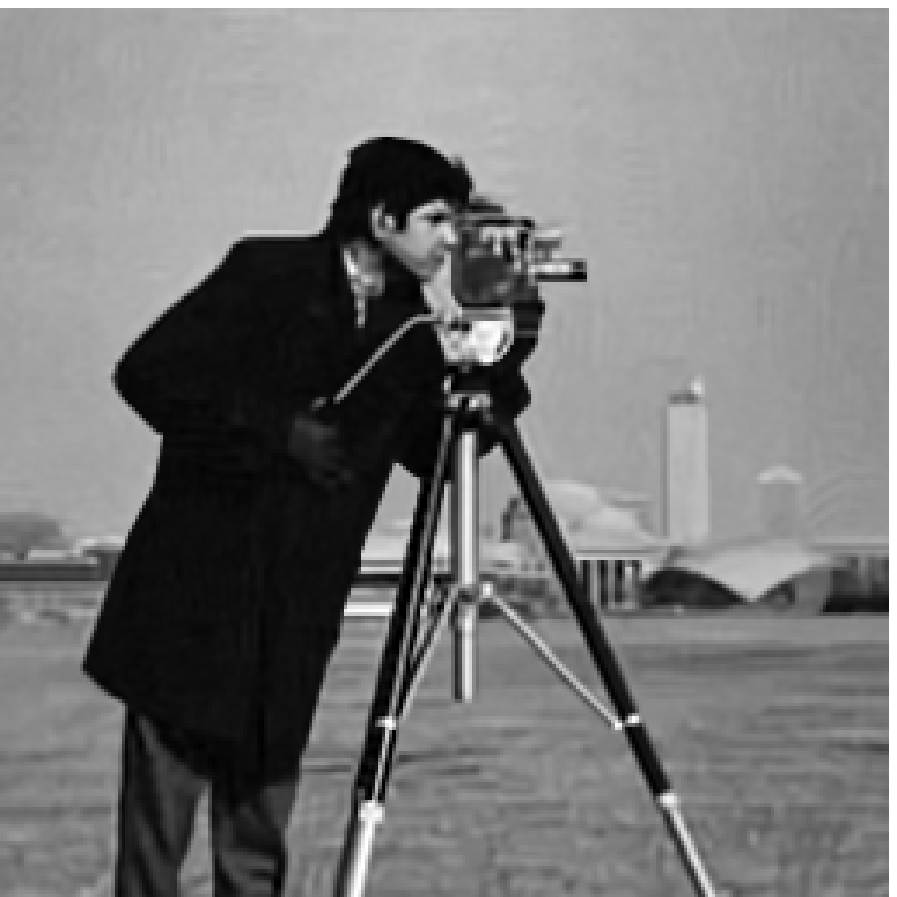}
				\caption{\centering The initial image \newline (IDD-BM3D) \newline PSNR = 28.63dB}
				\label{fig:db_Lin-Kernighan:idd_bm3d}
				\vspace*{6pt}
			\end{subfigure}
			\begin{subfigure}{0.32\textwidth}
				\includegraphics[width=\textwidth]{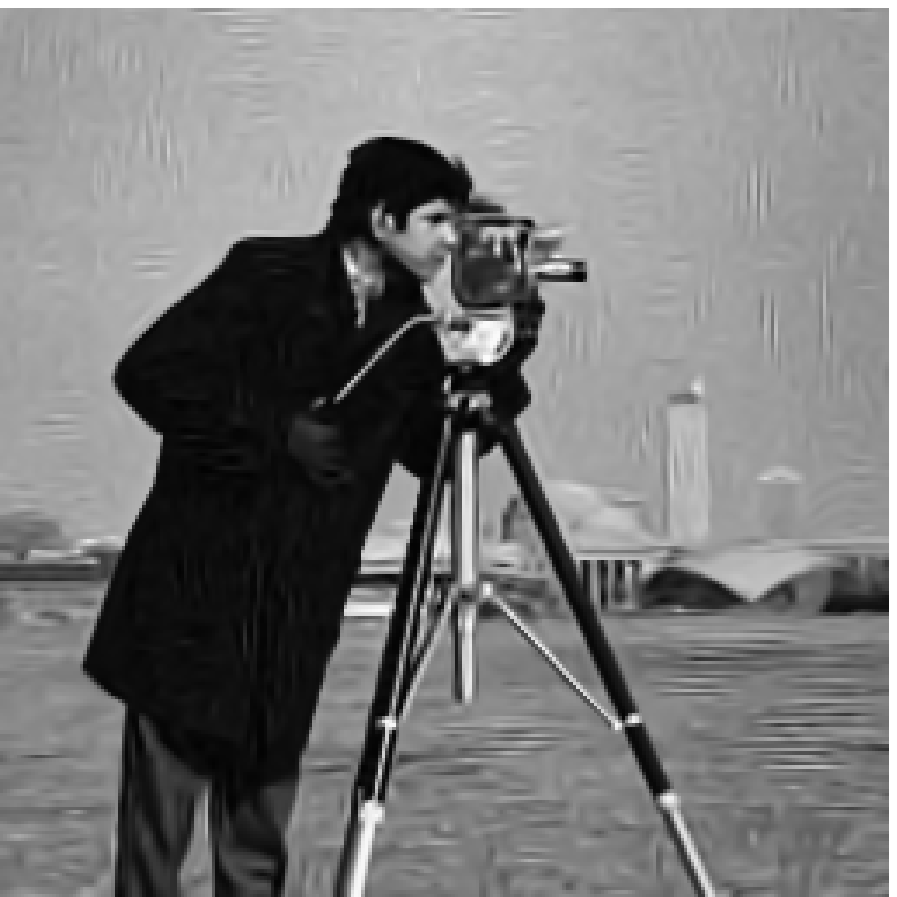}
				\caption{\centering Output with Lin-Kernighan \newline PSNR = 28.01dB \newline}
				\label{fig:db_Lin-Kernighan:lk_out}
				\vspace*{6pt}
			\end{subfigure}
			\begin{subfigure}{0.32\textwidth}
				\includegraphics[width=\textwidth]{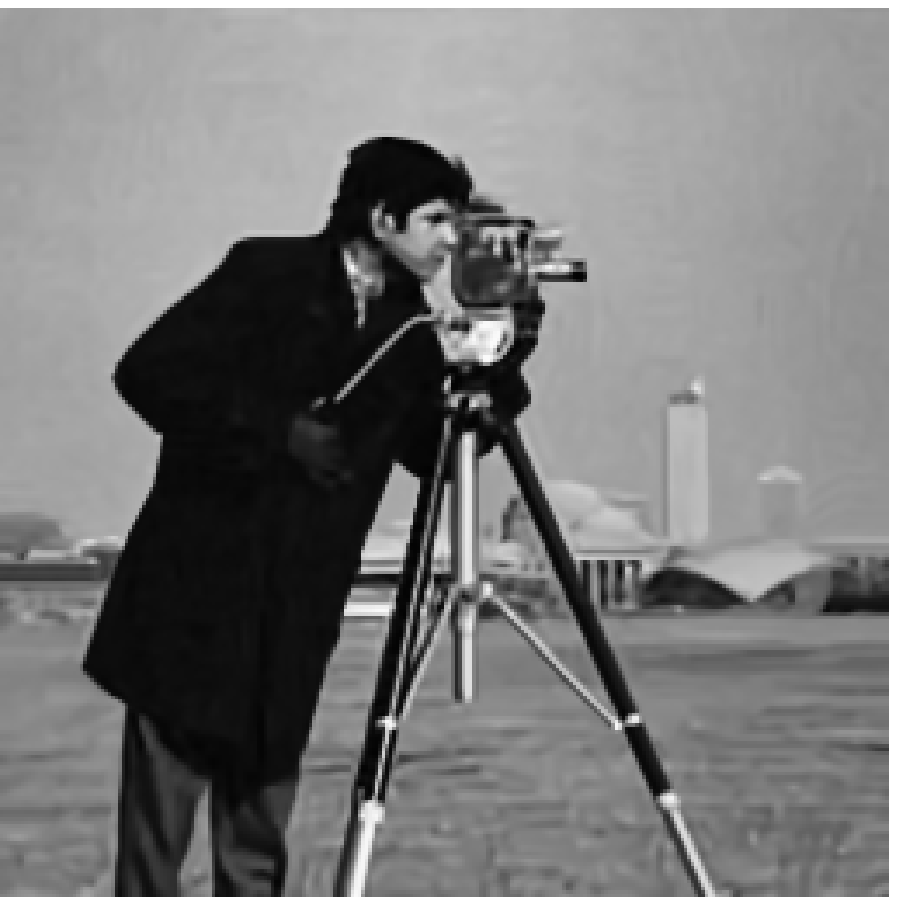}
				\caption{\centering Output with Algorithm~\ref{alg:reordering} \newline PSNR = 29.02dB \newline}
				\label{fig:db_Lin-Kernighan:NN_out}
				\vspace*{6pt}
			\end{subfigure}
			\caption{The output of our scheme with the Lin-Kernighan heuristics or Algorithm~\ref{alg:reordering} for approximating the TSP problem. The reconstruction scheme is applied for test 4 of the image debluring task, and initialized with the IDD-BM3D result.}
			\label{fig:db_Lin-Kernighan}
		\end{figure}

\begin{figure}
			\centering
			\begin{subfigure}{0.32\textwidth}
				\includegraphics[width=\textwidth]{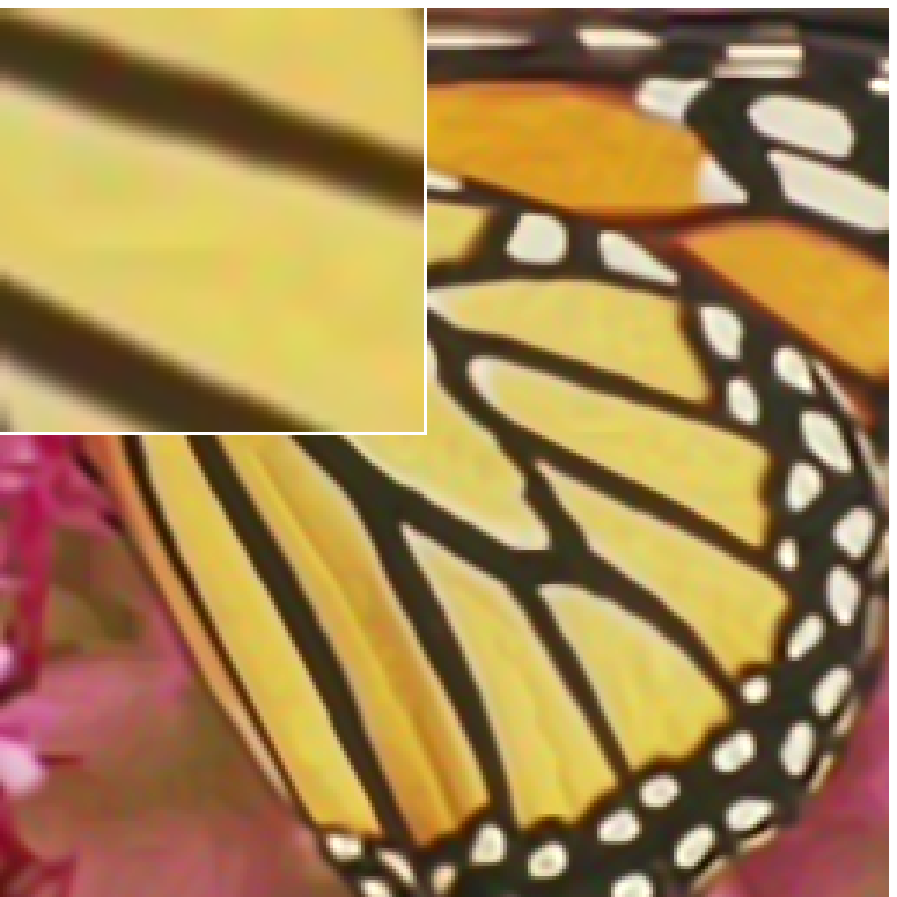}
				\caption{\centering The initial image (NSCR) \newline PSNR = 26.87dB}
				\label{fig:sr_Lin-Kernighan:nscr}
				\vspace*{6pt}
			\end{subfigure}
			\begin{subfigure}{0.32\textwidth}
				\includegraphics[width=\textwidth]{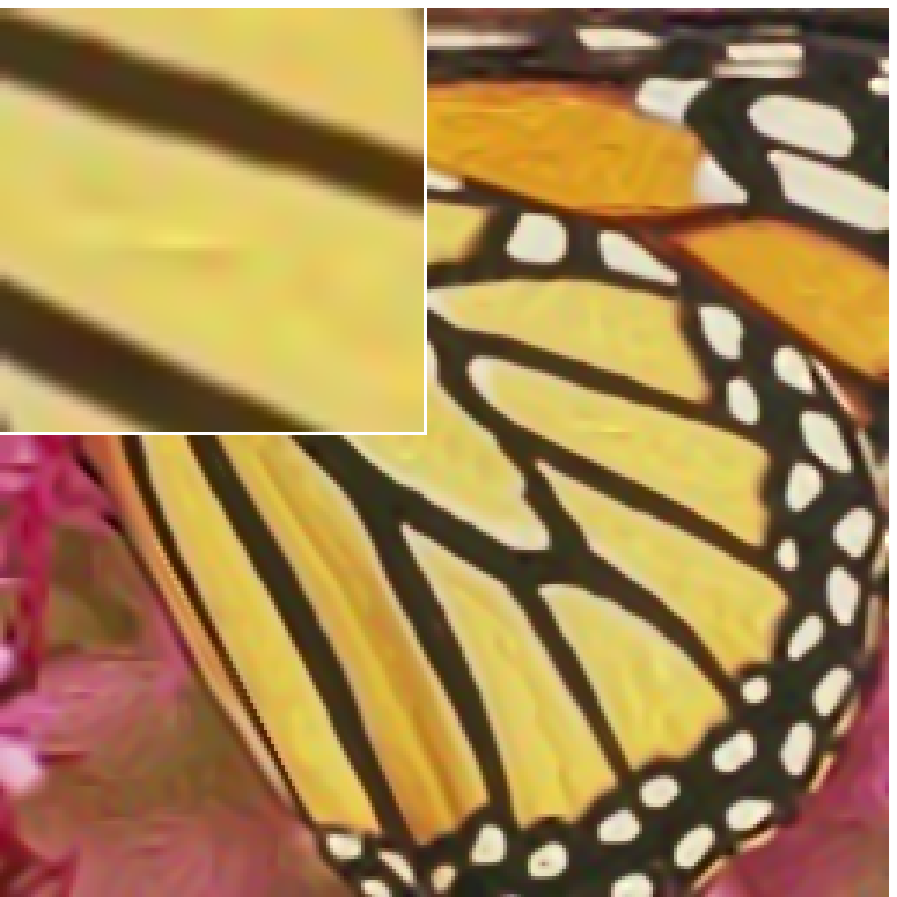}
				\caption{\centering Output with Lin-Kernighan \newline PSNR = 27.86dB}
				\label{fig:sr_Lin-Kernighan:lk_out}
				\vspace*{6pt}
			\end{subfigure}
			\begin{subfigure}{0.32\textwidth}
				\includegraphics[width=\textwidth]{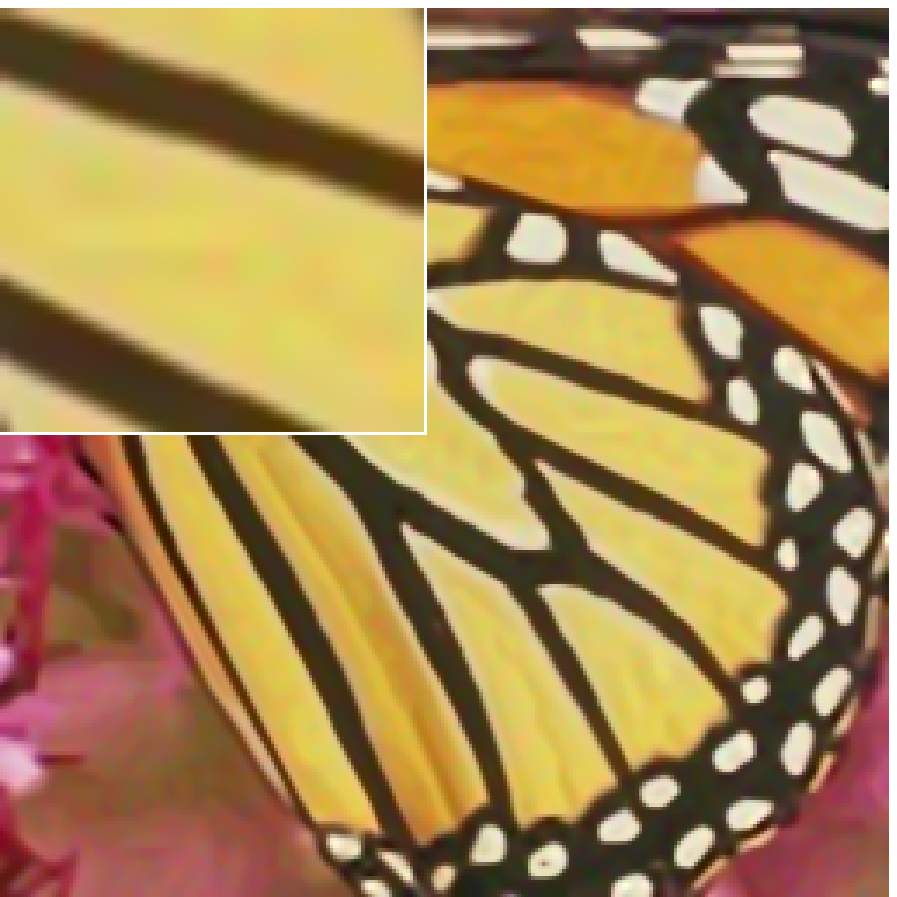}
				\caption{\centering Output with Algorithm~\ref{alg:reordering} \newline PSNR = 28.00dB}
				\label{fig:sr_Lin-Kernighan:NN_out}
				\vspace*{6pt}
			\end{subfigure}
			\caption{The output of our scheme with the Lin-Kernighan heuristics or Algorithm~\ref{alg:reordering} for approximating the TSP problem. The reconstruction scheme is applied for super-resolution task (with noise $\sigma = 5$), and initialized with the NSCR result.}
			\label{fig:sr_Lin-Kernighan}
		\end{figure}

\begin{figure}
			\centering
			\begin{subfigure}{0.32\textwidth}
				\includegraphics[width=\textwidth]{house4_poiss_spda}
				\caption{\centering The initial image (SPDA) \newline PSNR = 25.96dB}
				\label{fig:poiss_Lin-Kernighan:spda}
				\vspace*{6pt}
			\end{subfigure}
			\begin{subfigure}{0.32\textwidth}
				\includegraphics[width=\textwidth]{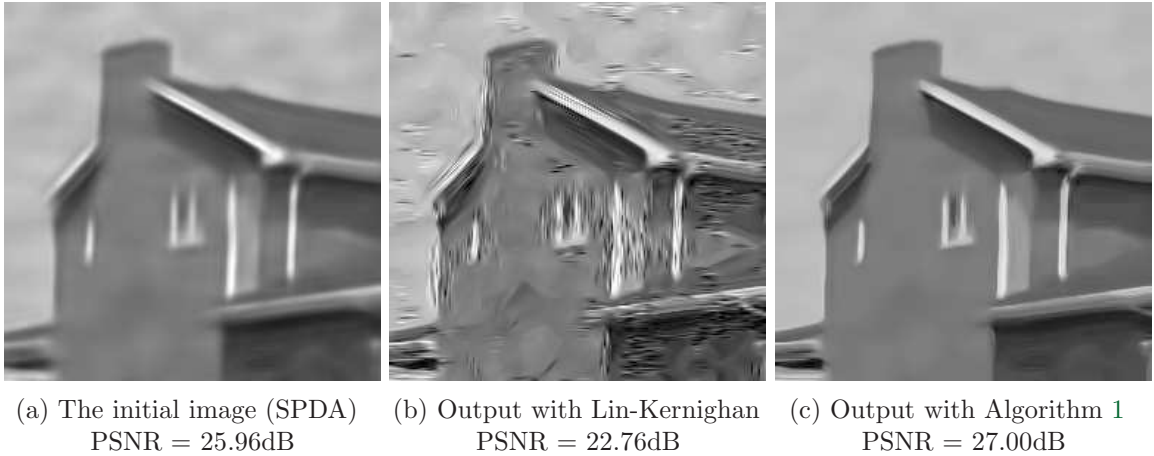}
				\caption{\centering Output with Lin-Kernighan \newline PSNR = 22.76dB}
				\label{fig:poiss_Lin-Kernighan:lk_out}
				\vspace*{6pt}
			\end{subfigure}
			\begin{subfigure}{0.32\textwidth}
				\includegraphics[width=\textwidth]{house4_s81_poiss_res}
				\caption{\centering Output with Algorithm~\ref{alg:reordering} \newline PSNR = 27.00dB}
				\label{fig:poiss_Lin-Kernighan:NN_out}
				\vspace*{6pt}
			\end{subfigure}
			\caption{The output of our scheme with the Lin-Kernighan heuristics or Algorithm~\ref{alg:reordering} for approximating the TSP problem. The reconstruction scheme is applied for Poisson denoising task with peak = 4, and initialized with the SPDA result.}
			\label{fig:poiss_Lin-Kernighan}
		\end{figure}

 In section~\ref{sec:permutations} we have shown artifact magnification effect when using a deterministic TSP solver. In this section we revisit this phenomena in order to better explain and demonstrate it. Recall that our algorithm starts with some reconstruction result, which is naturally not perfect and thus may contain artifacts. Our regularization relies on a permutation of patches from this image. Thus, a permutation obtained by a more exact TSP solver is in fact creating and magnifying artifacts, since it will assign patches with the same artifacts as neighbors in our ordering. If, on the other hand, one uses the more crude (and somewhat random) ordering that we propose, such similar patches will be separated in the ordering, thus leading to a smoothing out effect of some of these artifacts. In section~\ref{sec:permutations} Figure~\ref{fig:Lin-Kernighan} is showing an example of artifact magnification in the Gaussian denoising test. To complete this presentation, we bring here examples of this effect in other applications. In Figure~\ref{fig:db_Lin-Kernighan} we compare between the outputs of the proposed method with Lin-Kernighan heuristics and Algorithm~\ref{alg:reordering} for the debluring task. In Figures~\ref{fig:sr_Lin-Kernighan} and~\ref{fig:poiss_Lin-Kernighan} we do the same comparison for the super-resolution and Poisson denoising tasks respectively. Our conclusion from these experiments is that a more exact TSP solver magnifies artifacts in all applications.
		
	\subsection{Patch Reordering for Better Sparsification}
\begin{figure}
			\centering
			\includegraphics[width=0.9\textwidth]{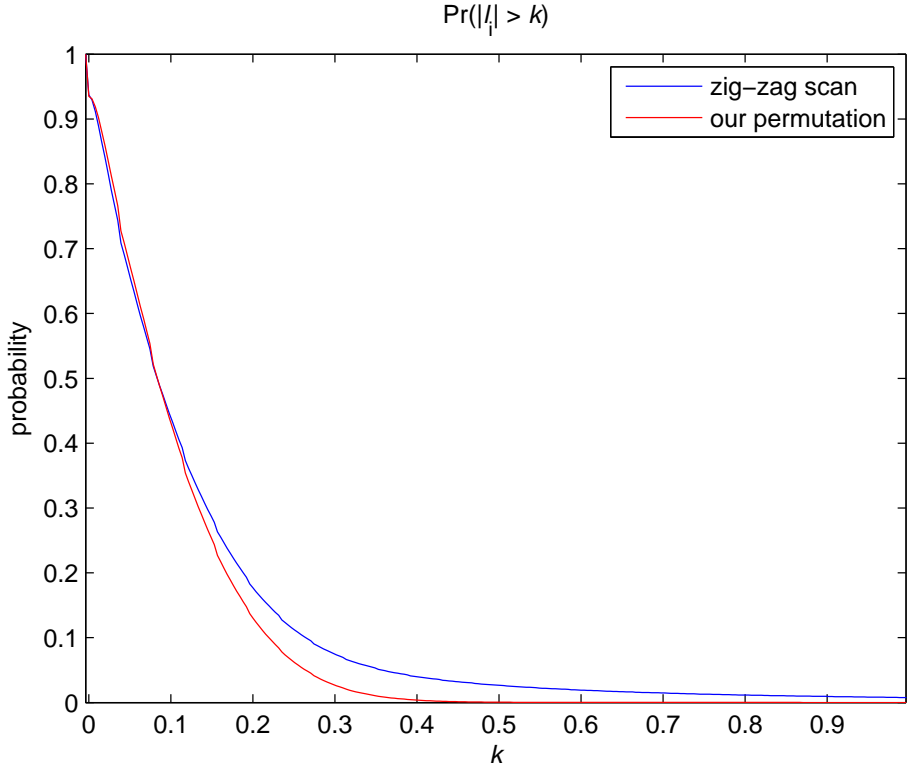}
			\caption{A comparison of the tail distribution functions of the Laplacian result with our ordering versus the zig-zag scan} 
			\label{fig:probability_graph}
		\end{figure}

		In order to better understand the role of the permutation in the proposed algorithm, we compare the tail distribution functions of the Laplacian result with our ordering versus a zig-zag scan. The Laplacian result are calculated by computing the vector ${\mathbf{l} = MLP\mathbf{x}}$, which is the essence of the regularization term we propose. Using all images listed in Table~\ref{tab:gauss_res} (referring to the Gaussian denoising test) we compute the cumulative distribution function, ${CDF(k) = Pr(|l_i| \le k)}$ by applying a cumulative sum on the histogram of the components of $|\mathbf{l}|$. For comparison we calculate this $CDF(k)$ while replacing our $P$ with the one obtained by a horizontal zig-zag scan ordering. A comparison of the tail distribution function, ${Pr(|l_i| > k) = 1 - CDF(k)}$ for zig-zag scan and our permutation is shown in Figure~\ref{fig:probability_graph}. As can be seen, our ordering leads to better sparsification of the Laplacian result, when applied to clean images. More specifically, the probability ${Pr(|l_i| > k)}$ for the zig-zag scan exhibits a tendency to have more non-zeros, and heavier tail, i.e. bigger values. 
	
	\subsection{Poorly Oredered Pixels}
		\par As already mentioned in Section~\ref{sec:permutations}, due to greedy nature of the NN heuristics in Algorithm~\ref{alg:reordering}, the last part of the reordered image is not as smooth as the rest of the ordering. Nevertheless, the reconstruction results of our algorithm, as shown in the previous section, seem to be of good quality, which may be puzzling. We therefore performed several experiments to better understand this behavior, and here we propose an explanation for this phenomenon.
		
		\par
\begin{figure}
			\centering
			\begin{subfigure}{0.73\textwidth}
				\includegraphics[width=\textwidth]{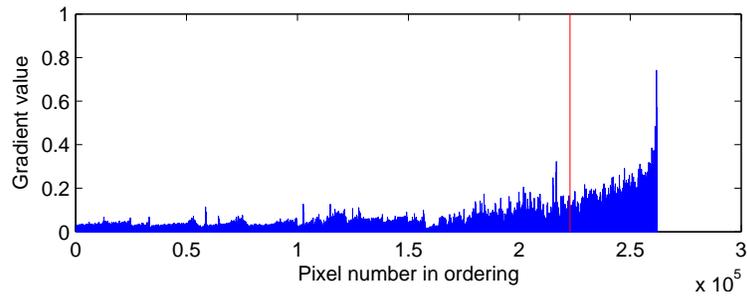}
				\caption{\centering Gradient's absolute value of the reordered pixels.}
				\vspace*{6pt}
				\label{fig:perm_grad}
			\end{subfigure}
			\begin{subfigure}{0.73\textwidth}
				\includegraphics[width=\textwidth]{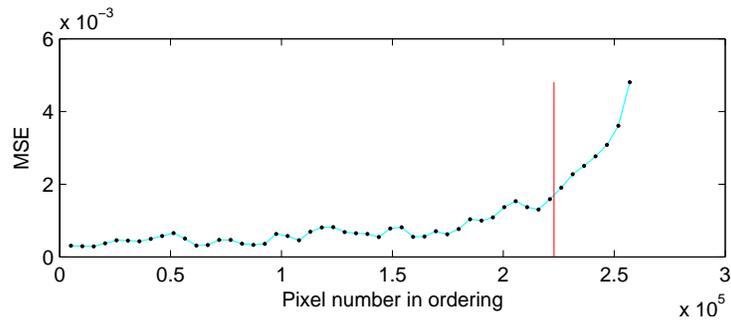}
				\caption{MSE of the output image.}
				\vspace*{6pt}
				\label{fig:mse}
			\end{subfigure}
			\caption{(a) The absolute value of the 1D gradient of the reordered \emph{Lena} image; (b) The MSE of Gaussian denoising of \emph{Lena} with $\sigma = 75$. Last 15\% pixels marked with red line.}
		\end{figure}

\begin{figure}
			\centering
			\begin{subfigure}{0.32\textwidth}
				\includegraphics[width=\textwidth]{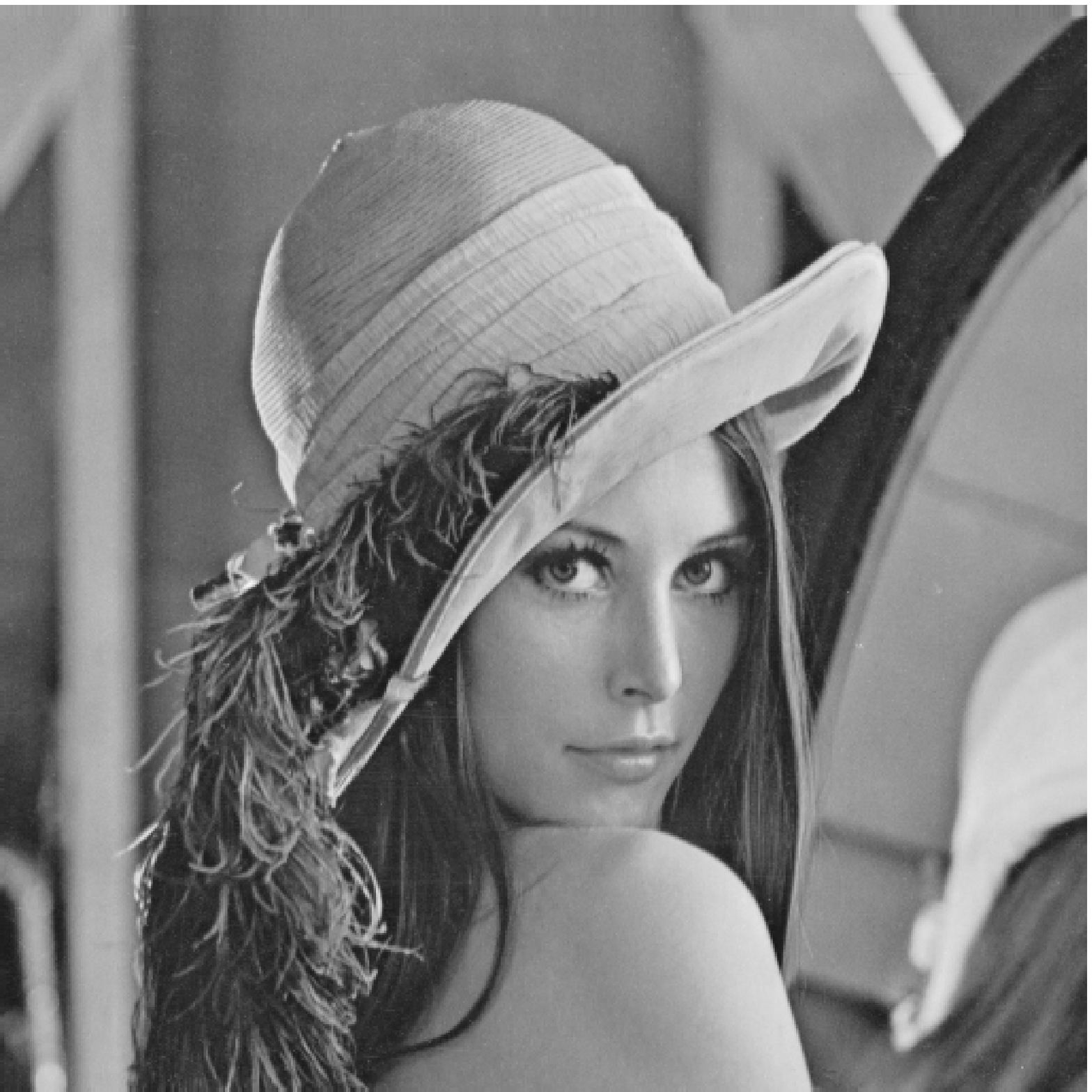}  
				\caption{\centering The original image \newline}
				\vspace*{6pt}
			\end{subfigure}
			\begin{subfigure}{0.32\textwidth}
				\includegraphics[width=\textwidth]{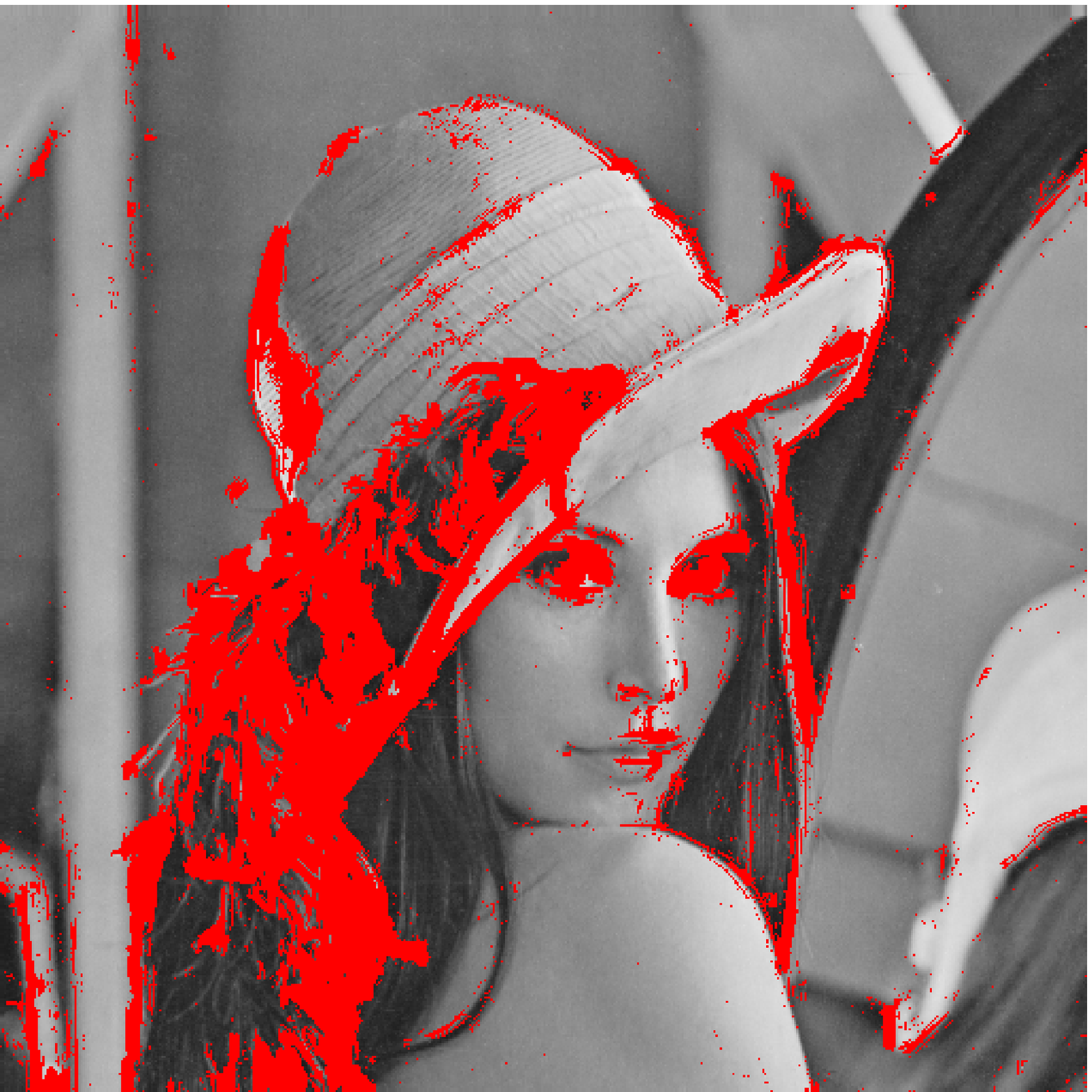}  
				\caption{\centering Last 15\% of the ordering marked Red}
				\label{fig:lena_15_last}
				\vspace*{6pt}
			\end{subfigure}
			\begin{subfigure}{0.32\textwidth}
				\includegraphics[width=\textwidth]{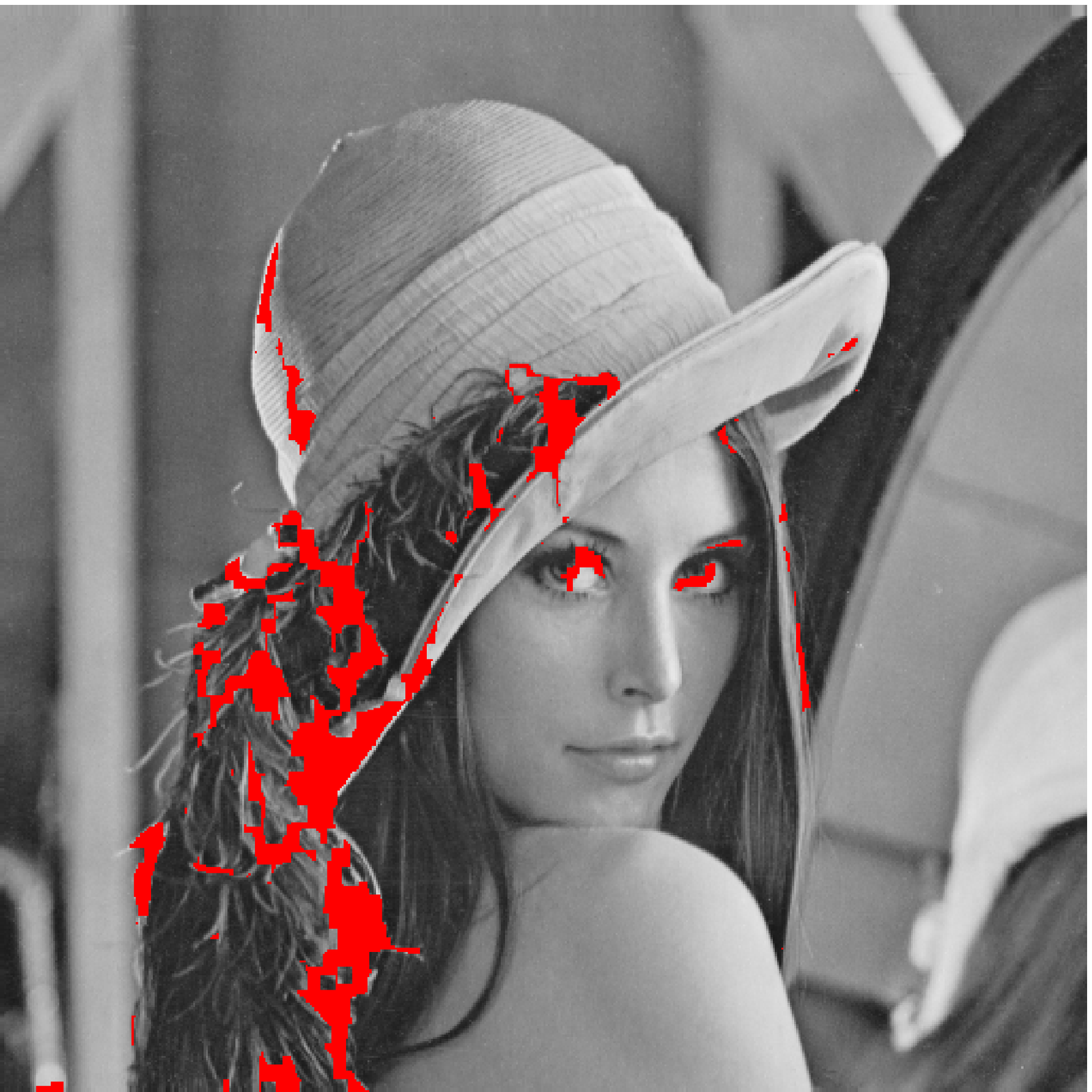}  
				\caption{\tabular[t]{@{}l@{}} Pixels that fall in last 15\% \\ of \emph{all} orderings marked Red \endtabular}
				\label{fig:lena_15_last_all}
				\vspace*{6pt}
			\end{subfigure}
			\caption{(a) The original image \emph{Lena}; (b) The last 15\% of ordering marked Red; (c) Pixels that fall in the last 15\% of all orderings (3.6\% in this experiment) marked Red.}
		\end{figure}

 In order to avoid side effects related to the initialization, we start our experiment by computing an oracle-permutation, i.e., applying Algorithm~\ref{alg:reordering} on the original clean image (\emph{Lena})\footnote{We repeated the following line of experiments for several images and the conclusions remain the same.}. First we plot a graph of the absolute value of the 1D gradient of the reordered image. As can be seen in Figure~\ref{fig:perm_grad}, approximately 15\% of the ordering towards the end is not smooth, just as expected. The location of these pixels in the image is shown in Figure~\ref{fig:lena_15_last}, and as can be seen, these pixels are mostly in edge and texture areas.
		
		\par We now perform a Gaussian denoising experiment with $\sigma = 75$ ($\mu = 0.13$), and our interest is in seeing how the MSE of the output pixels depends on their location in the ordering. We divide the permuted image into 50 groups of pixels with 50\% overlap between them, and plot a graph of the average MSE versus location in the ordering, where the x-axis corresponds to the center of each group and the y-axis is the MSE for the groups. This graph is shown in Figure~\ref{fig:mse}. Indeed, it is clear that the MSE grows as we tend towards the end of the ordering, implying that the last pixels is ill-treated. However, even for these seemly poorly-served pixels, the obtained MSE is significantly lower than the noise \mbox{level -- in} this test, the MSE of the last group is $4.8e{-}3$, where the noise level is $\left(\sigma / 255\right)^2 = 86.5e{-}3$. This suggests that even though the last ordered pixels seem to have poor choice of neighbors, their treatment is still rather effective. So, how come this happens?
		
		\par The answer resides in the sub-image accumulation proposed in our regularization scheme. If a pixel falls in the last part of the ordering, as shown above, it does not imply that this pixel may not be sufficiently close to its neighbors in the patch-ordering, because the poor neighbor assignments are true only in terms of the ordering applied to the central pixels of the patches. The very same poorly-positioned pixels are highly likely to be assigned with effective neighbors in other orderings, as each pixel participates in $n$ such optional permutations. In order to demonstrate this, we check how many pixels fall in the last 15\% of \emph{all} $n$ orderings. For the image \emph{Lena} this count drops to 3.6\% of the pixels, and the location of these pixels is shown in Figure~\ref{fig:lena_15_last_all}. These pixels are characterized by the fact that they have no relevant neighbors in the image, regardless of the permutation strategy adopted. As such, they are expected to be ill-treated in the restoration procedure.
		
%		\begin{figure}[Tbp]
		
	\subsection{Initialization Strategy}
		\par
\begin{figure}
			\centering
			\begin{subfigure}{0.32\textwidth}
				\includegraphics[width=\textwidth]{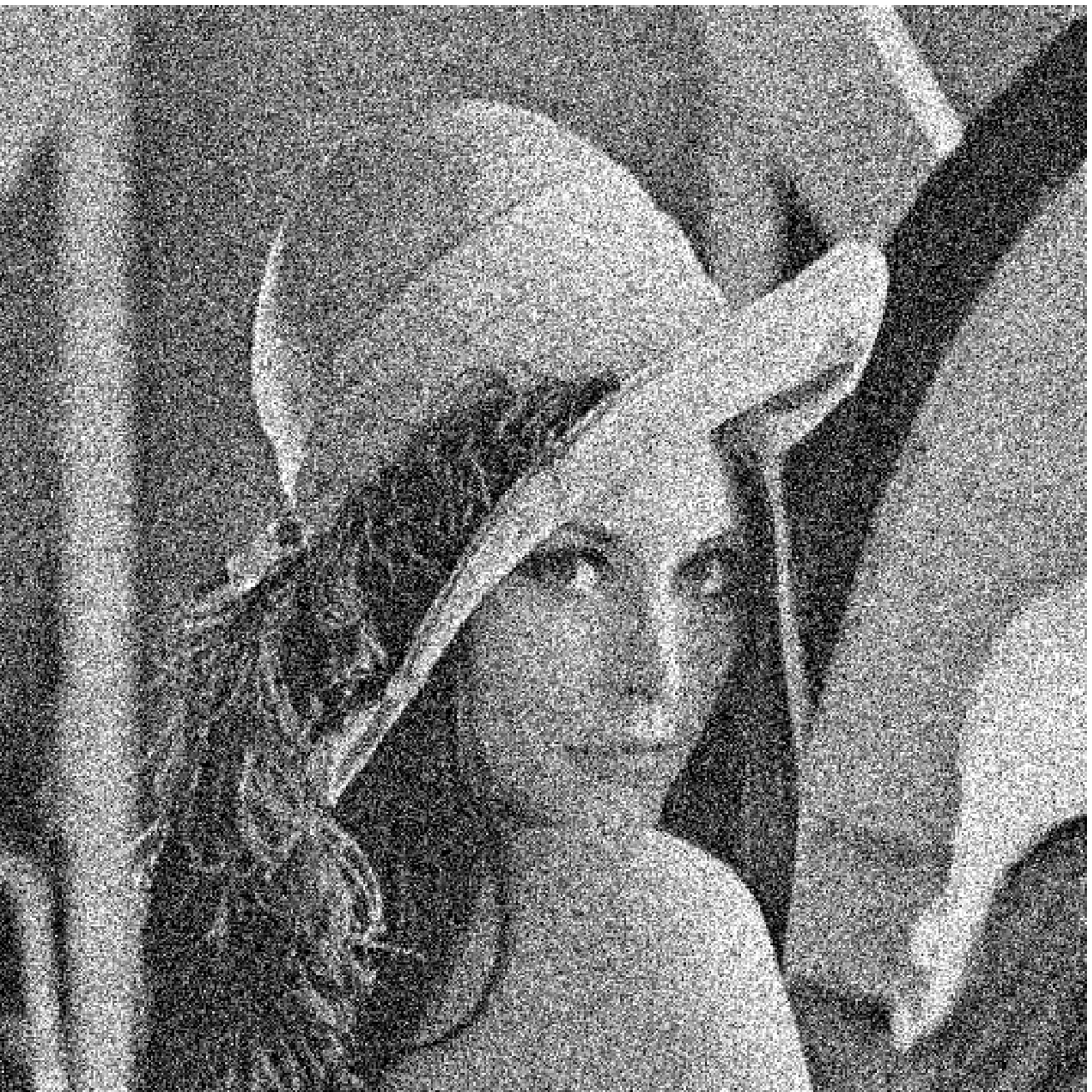}
				\caption{\centering Noisy \emph{Lena} \newline PSNR = 14.16dB}
				\vspace*{6pt}
			\end{subfigure}
			\begin{subfigure}{0.32\textwidth}
				\includegraphics[width=\textwidth]{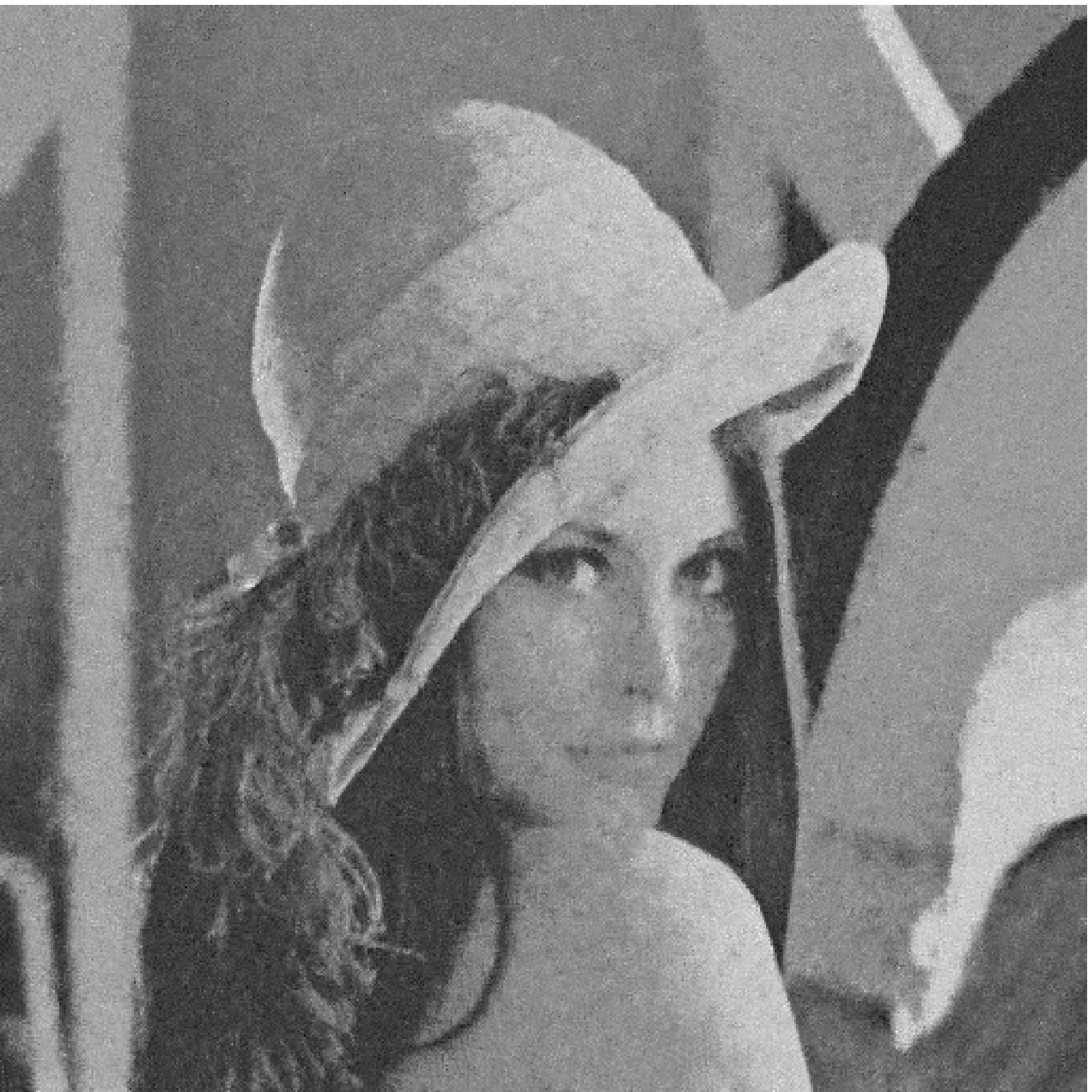}
				\caption{\centering Reconstructed: 1'st iteration \newline PSNR = 24.55dB}
				\vspace*{6pt}
			\end{subfigure} \\
			\begin{subfigure}{0.32\textwidth}
				\includegraphics[width=\textwidth]{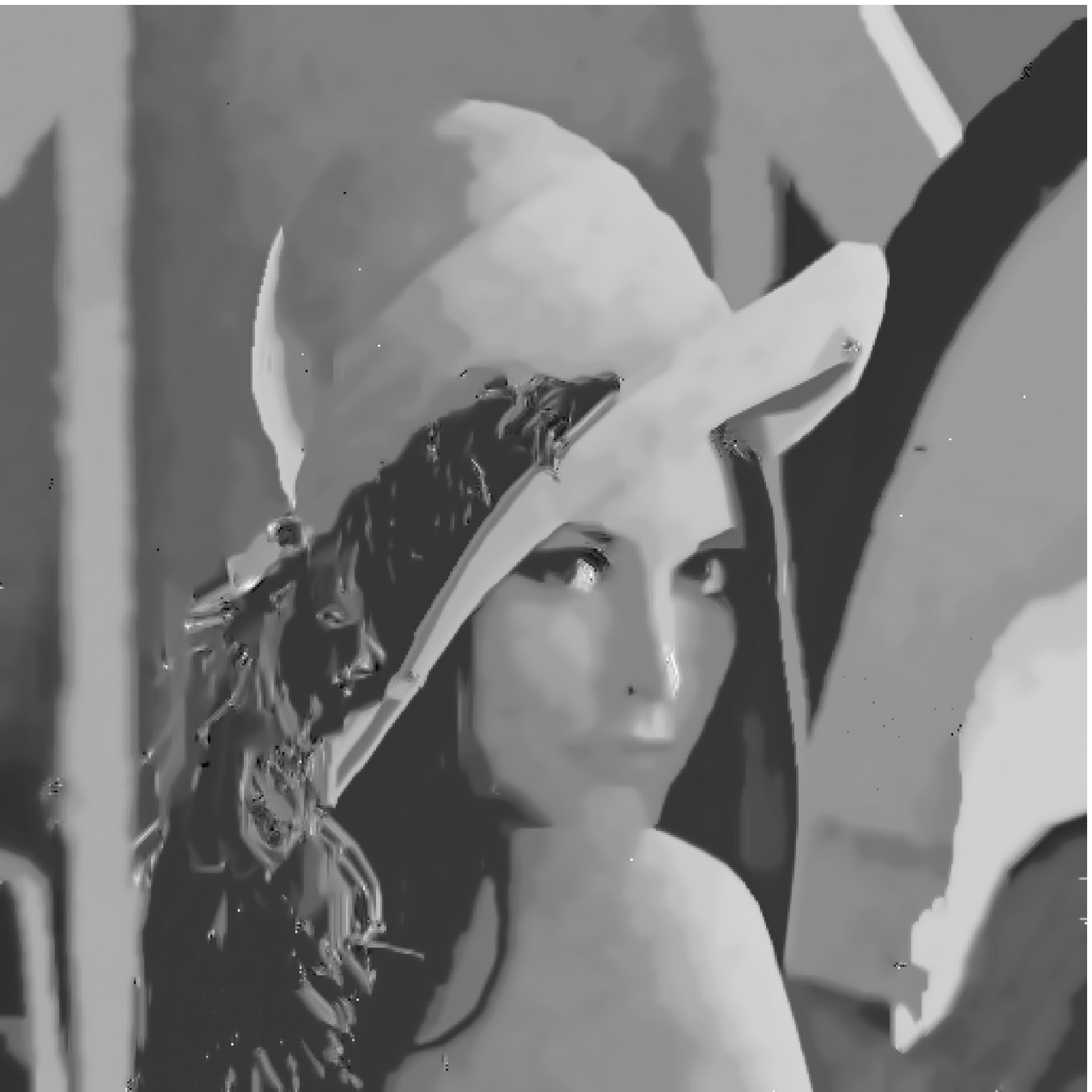}
				\caption{\centering Reconstructed: 7'th iteration \newline PSNR = 27.62dB}
				\vspace*{6pt}
			\end{subfigure}
			\begin{subfigure}{0.32\textwidth}
				\includegraphics[width=\textwidth]{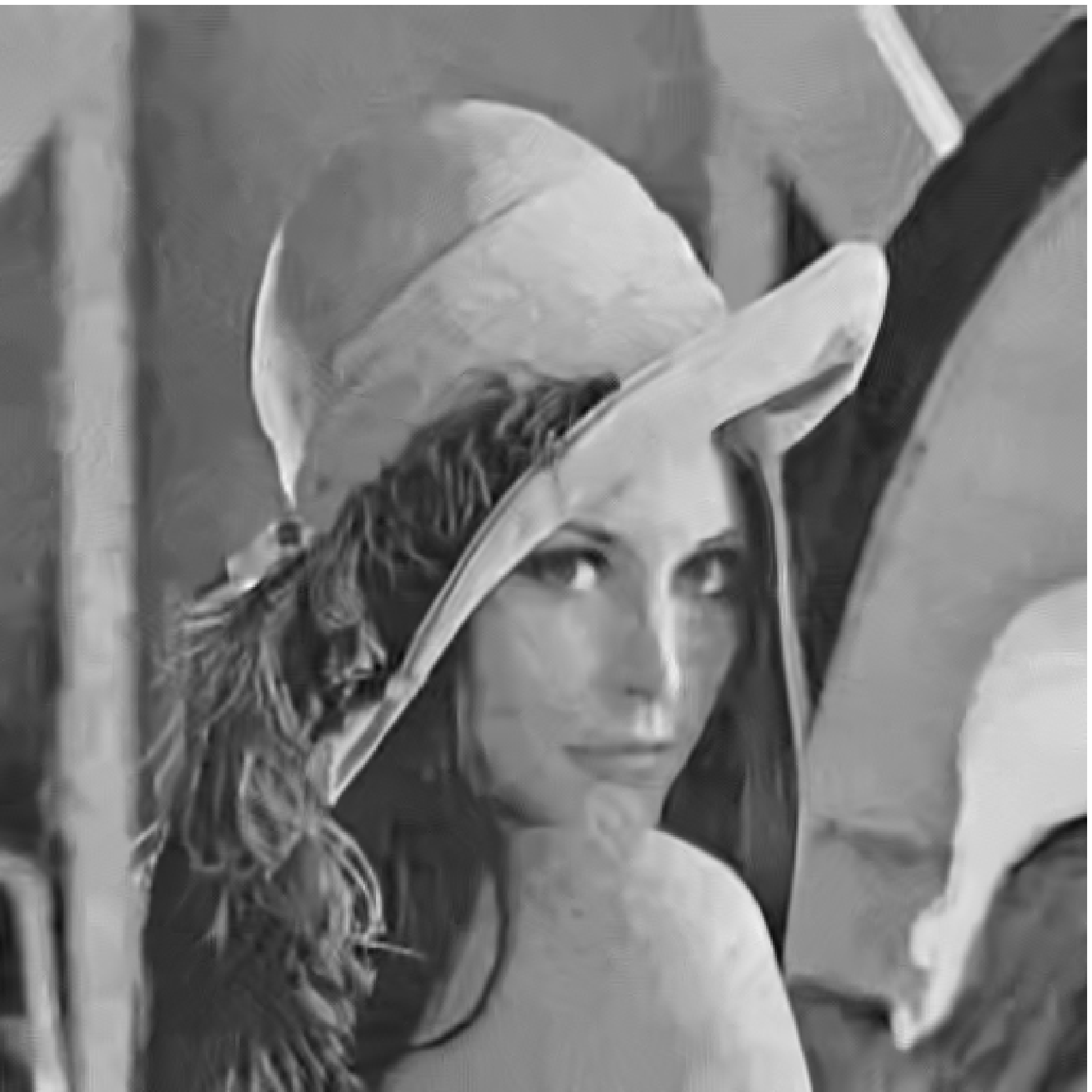}
				\caption{\centering BM3D result \newline PSNR = 29.05dB}
				\vspace*{6pt}
			\end{subfigure}
			\caption{Example of Gaussian denoising results with noisy initialization for the \emph{Lena} image with ${\sigma = 50}$.}
			%		\vspace*{12pt}
			\label{fig:gauss_corrupted_res}
		\end{figure}

 The prime drawback of the proposed method is that it relies on a good initialization. In order to demonstrate this we run a Gaussian denoising experiment on \emph{Lena} with ${\sigma = 50}$ and without BM3D initialization. In this experiment we apply seven rounds of minimization of the penalty function, each followed by a construction of the permutation. The first iteration uses the noisy image itself for building the permutation, while the next iterations rely on the temporary output. We use ${\mu = 0.45}$ for the first iteration, ${\mu = 0.12}$ for the second, and ${\mu = 0.08}$ for the rest. As for the remaining parameters, we used the values listed in~\cref{tab:gauss_params_common}. Qualitative and quantitative results of this experiment are shown in Figure~\ref{fig:gauss_corrupted_res}. Clearly, the result falls short compared to BM3D and other state-of-the-art denoising methods. Thus, our algorithm relies heavily on a good-quality initialization, and further work is required for seeking alternatives to this initialization strategy.
		
\section{Conclusion}
	\label{sec:conclusion}
	\par In this paper we have extended the work in~\cite{Ram_Patch_Ordering_2013}, which introduced the concept of patch ordering for handling image denoising and inpainting. Our work exploits the existing interrelation between image patches by building a MAP estimator with permutation-based smoothness-promoting prior on the objective image. The presented scheme is applicable to diverse set of inverse problems in image processing, and in this work we demonstrate its effectiveness for Poisson image denoising, Gaussian image denoising, image deblurring and single image super resolution. In most these tests, our method improved the image quality comparing to the initial image, in some experiments showing an improvement that goes beyong 1dB.
	
	\par Note that throughout the experiments reported above, we have set the patch ordering only once, based on the initialization image, and then restored a better outcome using our algorithm. In principle, one should iterate this process, by updating the ordering based on the improved image, and then minimizing the MAP functional again. Our initial tests suggest that this is not effective if we are using the same ordering algorithm as described in Section~\ref{sec:permutations}. In our future work we plan to investigate alternative ordering that would lead to an overall improvement. In a wider context, we are also seeking ways to depart from the greedy ordering method depicted in \textbf{Algorithm~\ref{alg:reordering}} in various ways, such that the final outcome is further improved.

\Appendix
\section{L-BFGS}
	\par This appendix provides brief description of the \mbox{L-BFGS} algorithm taken from~\cite{Nocedal_1999_optimization}. For this section we denote by~${f(\mathbf{x})}$ a smooth function of~${\mathbf{x}}$, where ${\mathbf{x} \in \mathbb{R}^n}$ is a real vector of length~${n}$.
	
	\par \mbox{L-BFGS} is a limited memory quasi-Newton method, which approximates the inverse of the Hessian matrix instead of calculating the exact one. In additional, it stores only ${m}$ vectors of length ${n}$ (${m \ll n}$) that capture curvature information from recent iterations, instead of saving the full ${n \times n}$ approximation of the Hessian.
	Each step of L-BFGS is given by
	\begin{IEEEeqnarray*}{lll}
		\mathbf{p}_k & \ = \ & -H_k \nabla f(\mathbf{x}_k) \\
		\mathbf{x}_{k+1} & \ = \ & \mathbf{x}_k + \alpha_k \mathbf{p}_k, \qquad k = 0, 1, 2 \ldots \;, \IEEEyesnumber 
	\end{IEEEeqnarray*}
	where ${\mathbf{x}_k}$~is~${\mathbf{x}}$ at iteration~${k}$, ${\alpha_k}$~is the step length, and ${H_k}$~is the approximation of the inverse of the Hessian of ${f(\mathbf{x})}$ at~${\mathbf{x}_k}$ (i.e. ${H^{-1}_k}$ is the approximated Hessian). The ${H_k}$~matrix is updated at every iteration using the formula
	\begin{equation}
	H_{k+1} = V^T_k H_k V_k + \rho_k \mathbf{s}_k \mathbf{s}^T_k \;,
	\label{eq:H_k_1}
	\end{equation}
	where 
	\begin{equation}
	\rho_k = \frac{1}{\mathbf{y}^T_k \mathbf{s}_x}, \qquad V_k = I - \rho_k \mathbf{y}_k \mathbf{s}^T_k \;,
	\end{equation}
	and
	\begin{equation}
	\mathbf{s}_x = \mathbf{x}_{k+1} - \mathbf{x}_x, \qquad \mathbf{y}_k = \nabla f(\mathbf{x}_{k+1}) - \nabla f(\mathbf{x}_k) \;.
	\label{eq:s_k_y_k}
	\end{equation}
	
	\par Instead of maintaining a full ${H_k}$ matrix of size ${n \times n}$, \mbox{L-BFGS} stores it implicitly, by storing ${m}$ vector pairs~${\{\mathbf{s}_i, \mathbf{y}_i\}}$. Given an initial inverse Hessian approximation~${H^0_k}$, and ${m}$~vector pairs~${\{\mathbf{s}_i, \mathbf{y}_i\}, \ i = k-m, ..., k-1}$, the~${H_k}$ matrix can be found by repeated application of Equation~(\ref{eq:H_k_1})
	\begin{IEEEeqnarray*}{l}
		\IEEElabel{eq:H_k_repeat}
		H_k = \left(V^T_{k-1} \ldots V^T_{k-m} \right) H^0_k \left(V_{k-m} \ldots V_{k-1} \right) \\ [4pt]
		+\: \rho_{k-m}\left(V^T_{k-1} \ldots V^T_{k-m+1} \right) \mathbf{s}_{k-m} \mathbf{s}^T_{k-m} \left(V_{k-m+1} \ldots V_{k-1} \right) \\ [4pt]
		+\: \rho_{k-m}\left(V^T_{k-1} \ldots V^T_{k-m+2} \right) \mathbf{s}_{k-m+1} \mathbf{s}^T_{k-m+1} \left(V_{k-m+2} \ldots V_{k-1} \right) \\ [4pt]
		+\: \ldots \\ [4pt]
		+\: \rho_{k-1} \mathbf{s}_{k-1} \mathbf{s}^T_{k-1} \;. \IEEEyesnumber
	\end{IEEEeqnarray*}
	Therefore, the product ${H_k \nabla f(\mathbf{x}_k)}$ can be obtained by a recursive algorithm that involves vector multiplications and summations (Algorithm~\ref{alg:l_bfgs_two_loops}). 
	
	\SetAlCapSkip{1em}
	\IncMargin{1em}
	\RestyleAlgo{boxed}
	\SetAlgoInsideSkip{medskip}
	
	\begin{algorithm}
		\SetKwInput{parameters}{Parameters}
		\SetKwInput{initialization}{Initialization}
		\flushleft
		\TitleOfAlgo{L-BFGS two loop recursion}
		\parameters{\mbox{${m}$ - number of stored vectors}, \mbox{${H^0_k}$ - initial} inverse Hessian approximation, \mbox{${\nabla f(\mathbf{x}_k)}$ - gradient of ${f(\mathbf{x})}$ at point ${\mathbf{x}_k}$}.}
		\initialization{${q = \nabla f(\mathbf{x}_k)}$} 
		\For{${i = k - 1, k - 2, \ldots, k - m}$} {
			${\alpha_i = \rho_i \mathbf{s}^T_i \mathbf{q}}$\;
			${q = q - \alpha_i \mathbf{y}_i}$\;
		}
		\For{${i = k - m, k - m + 1, \ldots, k - 1}$} {
			${\beta = \rho_i \mathbf{y}^T_i \mathbf{r}}$\;
			${\mathbf{r} = \mathbf{r} + \mathbf{s}_i (\alpha_i - \beta)}$\;
		}
		\KwOut{${H_k \nabla f(\mathbf{x}_k) = \mathbf{r}}$.}
		\caption{L-BFGS two-loop recursion}
		\label{alg:l_bfgs_two_loops}
	\end{algorithm}
	\DecMargin{1em}
	
	\par
\begin{figure}
		\centering
		%		\begin{subfigure}{0.48\textwidth}
		\includegraphics[width=0.9\textwidth]{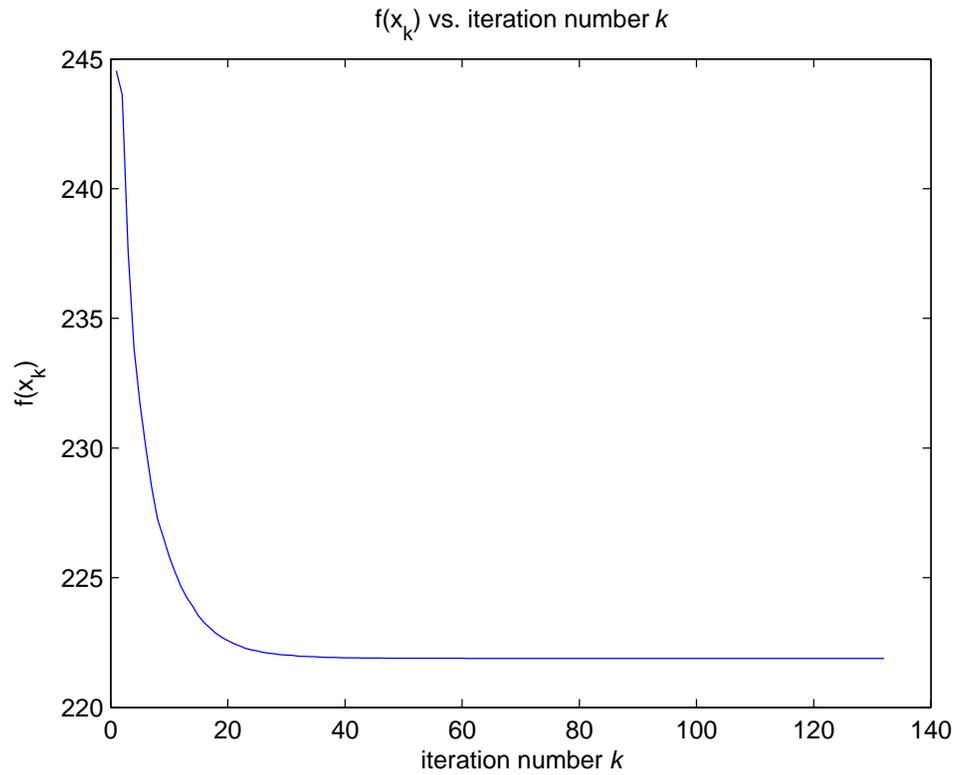}
		%			\caption{\centering Noisy \emph{Lena} \newline PSNR = 14.16}
		\caption{Example of L-BFGS convergence: ${f(\mathbf{x}_k)}$ vs. iteration number during Gaussian denoising of \emph{Peppers} image with~${\sigma = 25}$.} 
		%			\vspace*{6pt}
		%		\end{subfigure}
		\label{fig:convergence_graph}
	\end{figure}

 After each step of L-BFGS, the oldest vector pair~${\{\mathbf{s}_{k - m}, \mathbf{y}_{k - m}\}}$ is deleted from memory and replaced by the new pair ${\{\mathbf{s}_k, \mathbf{y}_k\}}$ obtained using Equation~(\ref{eq:s_k_y_k}). At each step of the algorithm, ${\alpha_k}$ is chosen to satisfy the Wolfe conditions
	\begin{IEEEeqnarray*}{rcl}
		\IEEElabel{eq:wolfe_cond}
		f(\mathbf{x}_k + \alpha_k \mathbf{p}_k)& \ \le \ & f(\mathbf{x}_k) + c_1 \alpha_k \nabla f(\mathbf{x}_k)^T \mathbf{p}_k \IEEEyesnumber \\ [4pt]
		\nabla f(\mathbf{x}_k + \alpha_k \mathbf{p}_k)^T \mathbf{p}_k & \ \ge \ & c_2 \nabla f(\mathbf{x}_k)^T \mathbf{p}_k \;.
	\end{IEEEeqnarray*}
	L-BFGS is summarized in Algorithm~\ref{alg:l_bfgs}. In our simulations we used minFunc implementation \cite{Schmidt_LBFGS_2005} with ${m = 8}$. Example of a typical L-BFGS convergence graph is shown in Figure~\ref{fig:convergence_graph}.
	
	\SetAlCapSkip{1em}
	\IncMargin{1em}
	\RestyleAlgo{boxed}
	\SetAlgoInsideSkip{medskip}
	
	\begin{algorithm}
		\SetKwInput{parameters}{Parameters}
		\SetKwInput{initialization}{Initialization}
		\flushleft
		\TitleOfAlgo{L-BFGS}
		\parameters{${m}$ - number of stored vectors.}
		\initialization{Choose starting point~${\mathbf{x}_0}$, and set ${k=0}$.} 
		\Repeat{convergense} {
			\begin{minipage}{0.95\hsize}
				\begin{itemize}[leftmargin=*] 
					\item[-] Choose ${H_k^0}$ (for example ${H_k^0 = I}$)\;
					\item[-] Compute ${\mathbf{p}_k = -H_k \nabla f(\mathbf{x}_k)}$ using Algorithm~\ref{alg:l_bfgs_two_loops}\;
					\item[-] Compute ${\mathbf{x}_{k+1} = \mathbf{x}_k + \alpha_k \mathbf{p}_k}$, where ${\alpha_k}$ satisfies the Wolfe conditions in Equation~(\ref{eq:wolfe_cond})\;
				\end{itemize} 
			\end{minipage}
			\flushleft
			\If{${k > m}$} {
				\begin{minipage}{0.95\hsize}
					\begin{itemize}[leftmargin=*] 
						\item[-] Discard the vector pair ${\{\mathbf{s}_{k-m}, \mathbf{y}_{k-m}\}}$ from storage\;
					\end{itemize} 
				\end{minipage} 
			}
			\begin{minipage}{0.95\hsize}
				\begin{itemize}[leftmargin=*] 
					\item[-] Update: ${\mathbf{y}_k = \nabla f(\mathbf{x}_{k+1}) - \nabla f(\mathbf{x}_k)}$, ${k = k + 1}$, and ${\mathbf{s}_k = \mathbf{x}_{k+1} - \mathbf{x}_k}$\;
				\end{itemize} 
			\end{minipage}
		}
		\KwOut{${\mathbf{x}_k}$ is a solution of the minimization problem.}
		\caption{L-BFGS}
		\label{alg:l_bfgs}
	\end{algorithm}
	\DecMargin{1em}
	
\FloatBarrier

\bibliography{PoissonBib}{}
\bibliographystyle{IEEEtran}

\end{document}